\documentclass[journal]{IEEEtran}
\usepackage{amsfonts}
\usepackage[noadjust]{cite}
\usepackage{amssymb}
\usepackage{algorithm}
\usepackage{bm}
\usepackage{algorithmic}
\usepackage{enumitem}
\usepackage{multirow,graphicx}
\usepackage{mathrsfs}
\usepackage{pdfpages}
\usepackage{lipsum}
\usepackage[breaklinks]{hyperref}
\usepackage{cuted}
\usepackage{booktabs,subfigure,caption,array}
\usepackage{tabularx,booktabs,xcolor}
\usepackage{amsmath}
\usepackage[normalem]{ulem}
\def\myurl#1{\setbox0\vbox{\hsize.5\maxdimen
\url{#1}\par
\setbox0\lastbox
\global\setbox1\hbox{\unhbox0\unskip\unskip\unpenalty}}\unhbox1 }
\usepackage{scalerel,stackengine}
\stackMath
\newcommand\reallywidehat[1]{%
\savestack{\tmpbox}{\stretchto{%
  \scaleto{%
    \scalerel*[\widthof{\ensuremath{#1}}]{\kern.1pt\mathchar"0362\kern.1pt}%
    {\rule{0ex}{\textheight}}
  }{\textheight}%
}{2.4ex}}%
\stackon[-6.9pt]{#1}{\tmpbox}%
}
\newcommand{\diag}{\mathop{\rm diag}}
\newcommand{\conj}{\mathop{\rm conj}}

\makeatletter
\NewDocumentCommand{\sotwo}{O{red}O{black}+m}
    {%
        \begingroup
        \setulcolor{#1}%
        \setul{-.5ex}{.4pt}%
        \def\SOUL@uleverysyllable{%
            \rlap{%
                \color{#2}\the\SOUL@syllable
                \SOUL@setkern\SOUL@charkern}%
            \SOUL@ulunderline{%
                \phantom{\the\SOUL@syllable}}%
        }%
        \ul{#3}%
        \endgroup
    }
\makeatother
\ifCLASSINFOpdf

\else

\fi

\begin{document}
\title{Fast and High-Quality Blind Multi-Spectral \\ Image Pansharpening}
\author{Lantao~Yu,~\IEEEmembership{Student Member,~IEEE,}
        Dehong~Liu,~\IEEEmembership{Senior Member,~IEEE,}
        Hassan~Mansour,~\IEEEmembership{Senior Member,~IEEE,}
        and~Petros~T.~Boufounos,~\IEEEmembership{Senior Member,~IEEE}
        
\thanks{Lantao Yu is with the Department of Electrical and Computer Engineering,
Rice University, Houston, TX 77005 USA.}

\thanks{Dehong Liu, Hassan Mansour, and Petros T. Boufounos are with the Signal
Processing Group, Mitsubishi Electric Research Laboratories, Cambridge,
MA 02139 USA (e-mail: liudh@merl.com).}
        
}
\markboth{IEEE TRANSACTIONS ON GEOSCIENCE AND REMOTE SENSING}%
{Shell \MakeLowercase{\textit{et al.}}: Bare Demo of IEEEtran.cls for IEEE Journals}

\maketitle

\begin{abstract}

Blind pansharpening addresses the problem of generating a high spatial-resolution multi-spectral (HRMS) image given a low spatial-resolution multi-spectral (LRMS) image with the guidance of its associated spatially misaligned high spatial-resolution panchromatic (PAN) image without parametric side information. In this article, we propose a fast approach to blind pansharpening and achieve the state-of-the-art image reconstruction quality. Typical blind pansharpening algorithms are often computationally intensive since the blur kernel and the target HRMS image are often computed using iterative solvers and in an alternating fashion. To achieve fast blind pansharpening, we decouple the solution of the blur kernel and of the HRMS image. First, we estimate the blur kernel by computing the kernel coefficients with minimum total generalized variation that blur a downsampled version of the PAN image to approximate a linear combination of the LRMS image channels. Then, we estimate each channel of the HRMS image using local Laplacian prior(LLP) to regularize the relationship between each HRMS channel and the PAN image. Solving the HRMS image is accelerated by both parallelizing across the channels and by fast numerical algorithms for each channel. Due to the fast scheme and the powerful priors we used on the blur kernel coefficients (total generalized variation) and on the cross-channel relationship (LLP), numerical experiments demonstrate that our algorithm outperforms the state-of-the-art model-based counterparts in terms of both computational time and reconstruction quality of the HRMS images.

\end{abstract}

\begin{IEEEkeywords}
blind image fusion, pansharpening, local Laplacian prior, total generalized variation
\end{IEEEkeywords}

\IEEEpeerreviewmaketitle

\section{Introduction}

\IEEEPARstart{P}{ansharpening} is a data fusion technique that aims to generate well-aligned multi-spectral (MS) and panchromatic (PAN) images preserving both the spectral resolution of the MS image and the spatial resolution of the PAN image. In remote sensing, the physical constraints of an onboard imaging system limit the ability to simultaneously capture high spectral and spatial resolution from the raw data. Instead, two types of images with complementary spectral and spatial characteristics are generated. Specifically, the images involve a multi-spectral image with multiple channels covering a wide electromagnetic spectrum, each of which is of low spatial resolution, and a panchromatic image which records a wide electromagnetic spectrum in a single channel but with higher spatial resolution. Practically, the raw MS and PAN images are often misaligned relative to each other, since they are typically captured by imaging sensors at different spatial positions. The gaps in spectral and spatial resolution, as well as the spatial misalignment, necessitate a pansharpening algorithm to address the fusion of the MS and PAN images.

The fusion objective of a pansharpening algorithm ensures that each of the generated MS channel shall 1): be well-aligned to the PAN image; 2) carry the same spatial resolution as the PAN image; and 3) carry similar sharpness to the PAN image. For notational convenience, we refer to the target MS image as the high spatial-resolution multi-spectral (HRMS) image and the input MS image as the low spatial-resolution multi-spectral (LRMS) image. 

Various approaches to the pansharpening problem can be found in the literature~\cite{loncan2015hyperspectral}. Those approaches vary with different extents of information on the imaging platform, e.g., the spatial displacement between the HRMS and PAN images, the point spread functions corresponding to the LRMS and PAN channels, and the spectral responses of the multi-spectral sensors for simulating the PAN image. In this article, we focus on addressing the blind pansharpening problem. Namely, we assume other than the LRMS and PAN images and their associated electromagnetic spectra, none of the following information is known: 1) the displacement between the LRMS image and the PAN image; 2) the point spread functions corresponding to the LRMS and the PAN images; 3) the spectral responses of the multi-spectral sensors. The assumptions are rooted in the fact that the available information about the sensors is often scarce. Even though the information is available, the actual responses can be inconsistent with the responses provided by the manufacturers due to the diversity of the physical scenes and atmospheric turbulence. Enforcing such inconsistency when solving the pansharpening problem can lead to performance degradation.

Blind pansharpening is an ill-posed inverse problem. One challenging aspect is how to deal with the spatial misalignment. Traditional approaches perform registration before pansharpening. However, registration cannot guarantee precise alignment and tiny misaligned registration can lead to a significant performance drop on pansharpening stage. Later approaches overcome this drawback by jointly modeling the spatial alignment and the relative blur between the LRMS image and the HRMS/PAN images as center-shifted blur kernel coefficients. The blur kernel coefficients are solved in a unified optimization framework where the objective function regularizes the blur kernel coefficients, the relationship between the HRMS image and the PAN image, as well as the relationship between the LRMS image and the HRMS image. Within the optimization framework, the blur kernel coefficients and the target HRMS images are often solved by iterative numerical algorithms. These approaches that employ regularization and are solved via optimization algorithms are often called model-based approaches.  

The success of model-based blind pansharpening approaches heavily depends upon the quality of modeling the blur kernel coefficients. The major reason is that the measured LRMS images are modeled as downsampled version of the convolution of the blur kernel with the associated HRMS images. Therefore, the blur kernel coefficients significantly influences the low frequency components of the target HRMS images, which constitute the vast majority of the signal energy. Sim\~{o}es {\it{et al.}}~\cite{simoes2014convex} proposed HySure which regularizes the blur kernel by penalizing the $\ell_2$ norm of its gradients. Such approach often generate non-zero coefficients far from the peak and violates the short-support property of blur kernels. Later approaches~\cite{bungert2018blind,bajaj2019blind} utilized Total Variation (TV) to regularize the blur kernel's gradients. TV-based regularizers are beneficial to preserving the kernel's compact spatial support. However, they force small gradients to be $0$ and form a sharp boundary surrounding the peak of the coefficients. These characteristics of the blur kernels regularized by TV violate the overall smoothness of blur kernel coefficients. 

The success of model-based blind pansharpening approaches is also rooted in the quality achieved in modeling the cross-channel relationship between the HRMS image and the PAN image. Current state-of-the-art blind pansharpening approaches often utilize variational approaches to model the cross-channel relationship. Typical models exploit the spectral correlation between the HRMS and the PAN channels~\cite{simoes2014convex, wei2016blind}, low-dimensionality of the cube consisting of the HRMS and the PAN channels~\cite{simoes2014convex}, the overall piece-wise smoothness of the target HRMS image~\cite{wei2016blind} and the local linear model between a MS channel and a PAN image~\cite{bajaj2019blind, yu2020blind}. Unfortunately, the spectral correlation-based model is flawed since the PAN image is not necessarily approximated by a linear combination of the MS channels and forcing such constraint in the pansharpening stage often lead to spectral distortion. In addition, the local linear model in~\cite{bajaj2019blind} imposes constraints on the pixel domain (including low and high frequency components) of the HRMS image and conflicts with the data fidelity term that regularizes the low-frequency components of the HRMS image. This compromises the regularization on the low-frequency subbands and leads to performance drop.

In addition to the limitations on regularizers, current model-based blind pansharpening approaches often demand high computational cost. This is because they~\cite{bajaj2019blind, yu2020blind} iteratively solve for the blur kernel and the HRMS image in an alternating fashion. HySure largely saved the computational cost by decoupling the solution of the blur kernel and the HRMS image. However, the pansharpening stage requires hundreds of iterations. Wei {\it{et al.}}~\cite{wei2016blind} proposed a fast algorithm rooted in R-FUSE~\cite{wei2016r}. This algorithm offered a close-form solution accelerated by the Fast Fourier Transform (FFT). Yet, it failed to handle spatial misalignment. It is worth mentioning that an unrolled projected gradient descent convolutional neural network (PGDCNN) proposed by Lohit {\it{et al.}}~\cite{lohit2019unrolled} exploits the convolutional neural networks (CNN) to replace the projection operator within the optimization framework. This approach can be considered as the hybrid approach combining the ingredients of both model-based and learning-based approaches. PGDCNN managed to efficiently implement blind pansharpening on GPU. However, due to the limited remote-sensing dataset to train the network, its capability to deal with a large variety of blur and spatial misalignment is limited. 

In this article, we propose a novel model-based approach that efficiently addresses the blind pansharpening problem with state-of-the-art image quality. This approach utilizes novel and powerful regularizers on both the blur kernel coefficients and on the cross-channel relationship (between the HRMS image and the PAN image). To save the overall computational cost, our approach not only decouples the solution of the blur kernel and the HRMS image but also accelerates solving each HRMS channel via a fast algorithm. In addition, our approach allows parallelizing the solution of the HRMS images in a channel-wise fashion to save runtime. In summary, our article contributes to the following:

\begin{enumerate}[label=\arabic*)]
	\item  We use second-order total generalized variation (TGV$^2$)~\cite{bredies2010total} to regularize the blur kernel coefficients, which offers a more robust and more accurate estimation of the blur kernel than existing TV-based and $\ell_2$-based priors.
	\item  We propose a novel local Laplacian prior (LLP) to regularize the relationship between each HRMS channel and the PAN image, which offers better performance than related regularizers on the cross-channel relationship.
    \item We propose a fast numerical algorithm that accelerates the estimate of the HRMS images via efficient solutions in a parallel, channel-wise fashion.
 \end{enumerate}

The remaining sections of this article are organized as follows. In Section II, we formulate the problems in an optimization framework. In Section III, we provide the numerical algorithm to address the optimization problems. In Section IV, we first validate the advantages of our regularizers in controlled experiments. Then, we assess the performances of the proposed approach and compare it with state-of-the-art blind pansharpening approaches. Section V concludes the article and draws future developments.

\section{Problem Formulation}

\subsection{Notation}
In this article, we use $\mathbf{X} \in \mathbb{R}^{hw \times N}$ to denote the measured LRMS image with $N$ spectral channels, where $h$ and $w$ are the height and width of each channel, respectively. We denote the measured PAN image as $\mathbf{Y} \in \mathbb{R}^{HW \times 1}$, where $H$ and $W$ are its height and width, respectively. $H$ and $W$ are multiples of $h$ and $w$ with the same ratio, i.e. $({H}/{h})=({W}/{w})=c$, $c \geqslant 2$. The target HRMS image is denoted as $\mathbf{Z} \in \mathbb{R}^{HW \times N}$. The blur kernel is denoted as $\mathbf{u} \in \mathbb{R}^{n^2 \times 1}$, where $n$ is significantly smaller than $h$ and $w$. We typically denote a single channel of an image by stacking their columns into a vector. For example, the $i$~th channel ($1\leqslant i \leqslant N$) of the HRMS image $\mathbf{Z}$ is denoted as $\mathbf{Z}_i$. Also, we use $\mathbf{A}^{\top}$ to denote the Hermitian transpose of a matrix $\mathbf{A}$ and $\|\mathbf{A}\|_\mathrm{F}$ to denote the Frobenius norm of a matrix $\mathbf{A}$, i.e., the squared root of the sum of squared entries in $\mathbf{A}$. The 2-D circular convolution operator is denoted as $\circledast$ so that the result of circularly convolving an image (e.g., $\mathbf{Y}$) with $\mathbf{u}$ is also an image of the same dimension of the input (e.g., $HW \times 1$).

\subsection{General Formulation}

We formulate an objective function in an optimization framework to appropriately regularize the relationship between the target HRMS image and the measured LRMS image, the blur kernel coefficients, and the cross-channel relationship between the PAN channel and the HRMS channels.  The problem is formulated as
\begin{equation}\label{eqn:objective0}
\underset{\mathbf{Z},\mathbf{u}}{\operatorname{min}}\frac{1}{2}\|\mathbf{D}\mathbf{B}(\mathbf{u})\mathbf{Z}-\mathbf{X}\|^2_\mathrm{F}+\mathrm{R}_1(\mathbf{u})+\mathrm{R}_2(\mathbf{Z},\mathbf{Y}),
\end{equation}
where the first term is the data fidelity term that forces the blurred and downsampled version of the HRMS image to be close to the LRMS image. In the first term, the blur kernel $\mathbf{u}$ incorporates the spatial displacement of $\mathbf{Y}$ and $\mathbf{X}$, and the relative blur between $\mathbf{Z}$ and a version of $\mathbf{X}$ prior to downsampling. $\mathbf{B}(\mathbf{u}) \in\mathbb{R}^{HW \times HW}$ is the Toeplitz matrix implementing the convolution of the HRMS image $\mathbf{Z}$ with the blur kernel $\mathbf{u}$, and $\mathbf{D}\in\mathbb{R}^{hw \times HW}$ is the downsampling operator. The second term $\mathrm{R}_1(\mathbf{u})$ is the regularizer on the blur kernel coefficients. The third term $\mathrm{R}_2(\mathbf{Z},\mathbf{Y})$ is the regularizer on the relationship between the HRMS image and the PAN image.

\subsection{Efficient Surrogate}

Solving~\eqref{eqn:objective0} is typically computationally-intensive, since it necessitates many iterations to solve the blur kernel and the HRMS image in an alternating fashion. To pursue an efficient solution, we decouple the estimation of the blur kernel and the estimation of the HRMS image in~\eqref{eqn:objective0}. Specifically, we first estimate the blur kernel and then perform pansharpening. When solving the blur kernel $\mathbf{u}$, we choose not to regularize the downsampled and blurred version of the HRMS image to approximate the LRMS image as our previous work~\cite{yu2020blind} did. Instead, we exploit a well-known observation that the PAN image can be well approximated by a linear combination of its corresponding well-aligned MS channels that share the electromagnetic spectrum with the PAN image~\cite{ballester2006variational}. The observation was exploited by Sim\~{o}es {\it{et al.}} in~\cite{simoes2014convex} to enforce an unknown linear combination of a known subset of LRMS channels that are spectrally overlapped with the PAN image to approximate the downsampled version of the PAN image blurred by $\mathbf{u}$. We denote these LRMS channels as $\mathbf{X^{\prime}} \in \mathbb{R}^{hw \times N^\prime}$ and the weights to linearly combine these LRMS channels as $\bm{\omega} \in \mathbb{R}^{N^\prime \times 1}$, where $N^\prime(N^\prime \leqslant N)$ denotes the number of MS bands overlapped with the PAN image in the electromagnetic spectrum. We aim to solve the following optimization problem for estimating the blur kernel $\mathbf{u}$
\begin{equation}\label{eqn:objective1}
\underset{\mathbf{u},{\bm{\omega}}}{\operatorname{min}}\frac{1}{2}\|\mathbf{X}^{\prime}\bm{\omega}-\mathbf{D}\mathbf{B}(\mathbf{u})\mathbf{Y}\|^2_\mathrm{F}+ \mathrm{R}_1(\mathbf{u}) + \mathrm{R}_3(\bm{\omega}),
\end{equation}
in which the first term is the data fidelity term to enforce the blurred and then downsampled version of PAN image be close to an image synthesized by the linear combination of the PAN image's corresponding LRMS channels. The second term $\mathrm{R}_1(\mathbf{u})$ is our regularizer on the blur kernel and the third term $\mathrm{R}_2(\bm{\omega})$ is a regularizer on the weights. We will discuss them in details in Section~\ref{subsec:blur_kernel} and~\ref{subsec:blur_kernel_solve}.

Given the estimated $\mathbf{u}$ from solving~\eqref{eqn:objective1}, we solve the HRMS image in a channel-wise, parallel fashion to save the runtime. Namely, for the $i$~th HRMS channel, we solve
\begin{equation}\label{eqn:objective2}
\underset{\mathbf{Z}_i}{\operatorname{min}}\frac{1}{2}\|\mathbf{D}\mathbf{B}(\mathbf{u})\mathbf{Z}_i-\mathbf{X}_i\|^2_\mathrm{F}+\mathrm{R}_2(\mathbf{Z}_i,\mathbf{Y}),
\end{equation}
where the first term is the data fidelity term and the second term is the regularizer on the cross-channel relationship between the $i$~th HRMS channel and the PAN image. The regularizer on the cross-channel relationship will be detailed in Section~\ref{subsec:cross_channel}.

\subsection{The Blur Kernel Prior}
\label{subsec:blur_kernel}
In this article, we assume the spatial misalignment is strictly a translational shift between the HRMS/PAN images and the LRMS images, and no rotational distortion is involved. The spatial misalignment and the blur between the HRMS images and the LRMS images are determined by center-shifted blur kernel coefficients. Instead of using a parametric representation, e.g. 2-D Gaussain kernels, to model the blur kernel, we design a regularizer that targets generic properties for the blur kernel. Namely, the blur kernel coefficients shall be smooth, band-limited, and have compact spatial support with a non-vanishing tail. Recognizing the aforementioned drawbacks of $\ell_2$-based and TV-based regularizers, we propose to use a higher-order total variation based regularizer to preserve higher-order smoothness of the blur kernel and to simultaneously reject non-trivial coefficients far from the peak. Therefore, we adopt the $\mathrm{TGV}^2$~\cite{bredies2010total} as the regularizer, given by 
\begin{equation}\label{eqn:reg1}
\mathrm{R}_{1}(\mathbf{u})= \min_{\mathbf{p}} \left\{ \alpha_{1}\|\mathbf{\nabla} \mathbf{u}-\mathbf{p}\|_{2,1}+\alpha_{2}\|\mathcal{E}(\mathbf{p})\|_{2,1} \right\}+\mathbf{I}_{\mathbb{S}}(\mathbf{u}), 
\end{equation}
where $\mathbf{\nabla} \mathbf{u}=\left[\mathbf{\nabla}_h \mathbf{u} ~\mathbf{\nabla}_v \mathbf{u}\right] \in \mathbb{R}^{n^2\times2}$ are the horizontal and vertical gradients of $\mathbf{u}$; $\mathbf{p}=[\mathbf{p}_1 ~ \mathbf{p}_2]$ is an auxiliary variable to approximate $\mathbf{\nabla} \mathbf{u}$; 
$\mathcal{E}(\mathbf{p})=\left[{\mathbf{\nabla}_{h} \mathbf{p}_{1}}  ~
((\mathbf{\nabla}_{v}  \mathbf{p}_{1}+\mathbf{\nabla}_{h} \mathbf{p}_{2})/{2}) ~
((\mathbf{\nabla}_{v}  \mathbf{p}_{1}+\mathbf{\nabla}_{h} \mathbf{p}_{2})/{2}) ~
{\mathbf{\nabla}_{v} \mathbf{p}_{2}}\right]$ represents the first-order partial derivatives of $\mathbf{p}$; $\|\mathbf{X}\|_{2,1}=\sum_{i=1}^{n} (\sum_{j=1}^{m} x_{i,j}^{2})^{1/2}$ represents the sum of the vector norms of the partial derivatives of $\mathbf{p}$, $\alpha_1$, $\alpha_2$ are both scalars that control the regularization strength of $\mathbf{p}$'s approximation to $\mathbf{\nabla}\mathbf{u}$ and of the partial derivatives of $\mathbf{p}$; \begin{equation}
\label{eqn:simplex} 
\mathbb{S}=\{\mathbf{S} \in \mathbb{R}^{n^2 \times 1}| s_{i} \geqslant 0, \sum_{i} s_{i}=1\}	
\end{equation}
Equation~\eqref{eqn:simplex} is a simplex, and $\mathbf{I}_\mathbb{S}(\cdot)$ is its indicator function, which ensures that the computed blur kernel's coefficients are non-negative and has sum equal to $1$ to preserves the energy of the image. The intuition behind using $\mathrm{TGV}^2$ is that the desired smooth blur kernel shall also have small second-order derivatives, which favors the minimization of $\alpha_{2}\|\mathcal{E}(\mathbf{p})\|_{2,1}$. 

\subsection{The Cross-Channel Prior}
\label{subsec:cross_channel}

The data-fidelity term in~\eqref{eqn:objective2} reliably constrains the low-frequency components of the HRMS image to match the LRMS image. To recover the high-frequency components of the HRMS image, we regularize the high-frequency components of the HRMS image to be similar to those of the PAN image. The similarity is rooted in the well-admitted local linear model~\cite{he2012guided}, i.e., for an image recording multiple channels of the electromagnetic signals from the same scene, the pixel values in a spatial block from a channel can be well approximated by a linear affine function of the pixel values in the same spatial block from another channel. Thus, from the variational perspective if we were to regularize the cross-channel relationship on high-frequency domain, we can minimize the overall loss of approximating each block of high-frequency components of a target channel by the closest linear affine function of each block of high-frequency components from another channel. In light of the variational viewpoint inspired by the local linear model, we express the regularization function as
\begin{equation}\label{eqn:reg2}
\begin{aligned}
& \mathrm{R}_2(\mathbf{Z},\mathbf{Y})= \\
& \frac{\lambda}{2}\sum_{i,j}\sum_{k\in w_j} \bigg[	\big([{\cal{L}}(\mathbf{Z}_i)]_{j,k}- a_{i,j}[{\cal{L}}(\mathbf{Y})]_{j,k}-c_{i,j}\big)^2  + \epsilon a^2_{i,j}\bigg], 
\end{aligned}
\end{equation}
and name it as the local Laplacian prior ($\mathrm{LLP}$) since we use a 2-D Laplacian, expressed as ${\cal{L}}(\cdot)$ to extract the high-frequency component. In~\eqref{eqn:reg2}, the parameters are defined as follows: $\lambda$ is a scalar to control the regularization strength; $w_j$ is the $j$~th square window of size $(2r+1)\times(2r+1)$ in a $H\times W$ image, with $r$ an integer that is significantly smaller than $H$ and $W$; $k$ refers to the $k$~th element within the window, $k=1,2,\ldots,(2r+1)^2$;  ${\cal{L}}$$({\mathbf{Z}}_i)={\mathbf{Z}}_i\circledast \mathbf{S}$, with $\mathbf{S}=\begin{bmatrix}
0 & -1 & 0 \\
-1 & 4 & -1\\
0 & -1 & 0 \\
\end{bmatrix}$; $a_{i,j}$ and $c_{i,j}$ are scalar coefficients of the linear affine function in window $\omega_j$, corresponding to the Laplacian of the $i$~th band; $\mathbf{Z}_i$ is the $i$~th band of $\mathbf{Z}$; $\epsilon$ is a constant to avoid having a large $a_{i,j}$. The motivation behind using the $\ell_2$ loss here is to pursue a closed-form solution to save the runtime.

\begin{figure}[tb]
\centering
\begin{minipage}{0.6 \linewidth}
\centerline{\includegraphics[width=8 cm]{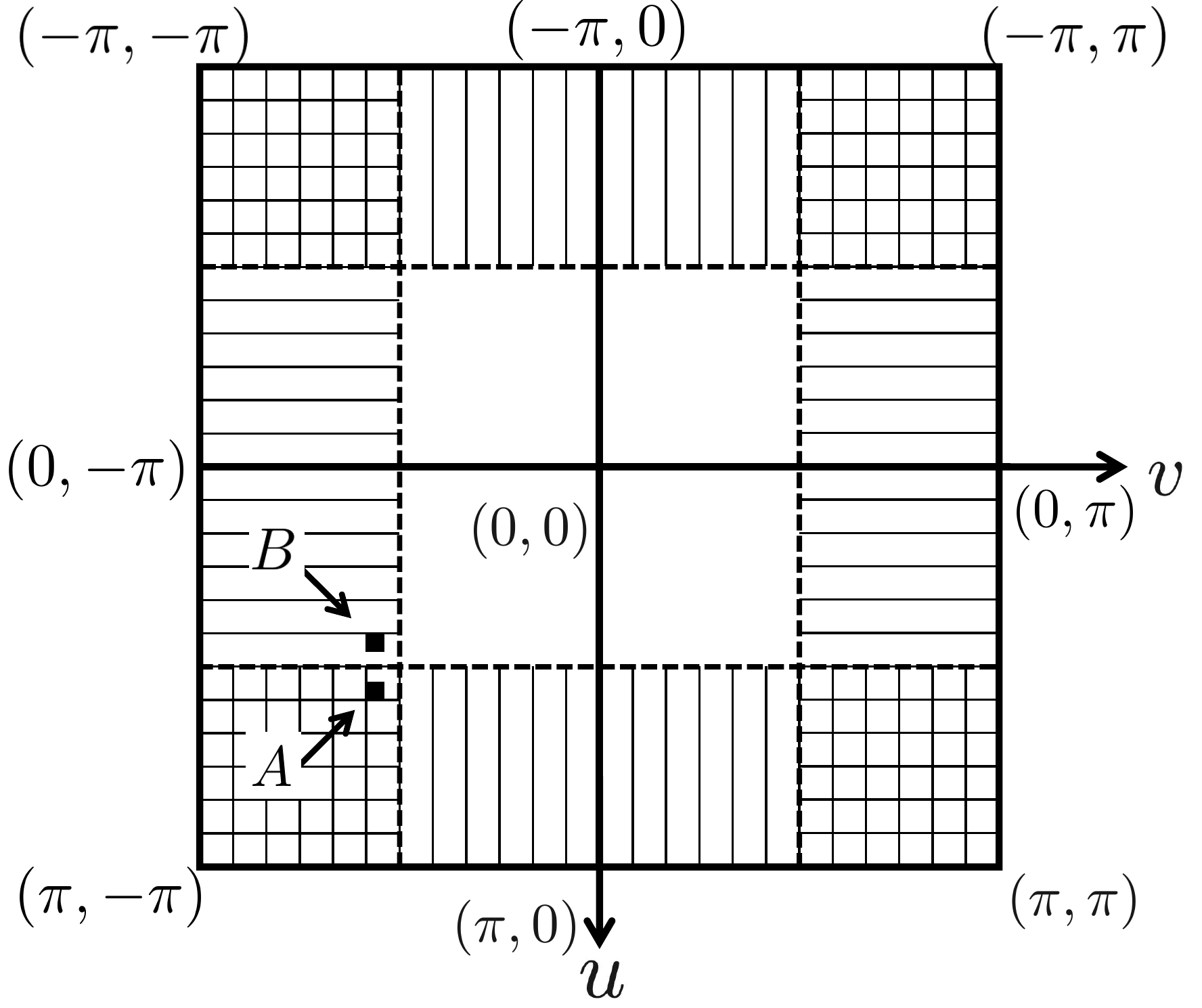}}
\end{minipage}
\caption{$\mathrm{LGC}$'s regularization on 2-D frequency. The horizontally and vertically shaded regions denote the schematic pass band of the 2-D frequency response of the horizontal and vertical gradients, respectively. A and B are two closely located 2-D frequency components, in the intersected region and the non-intersected region, respectively.}
\label{fig:2d_frequency_aware}
\end{figure}

It is worth noticing that Local Gradient Constraint ($\mathrm{LGC}$) from the literature~\cite{fu2019variational} had the same idea as we do in terms of regularizing the high-frequency components and demonstrated impressive performance when regularizing the cross-channel relationship. $\mathrm{LGC}$ is expressed as
\begin{equation}\label{eqn:reg2_lgc}
\begin{aligned}
& \mathrm{R}^{\prime}_2(\mathbf{Z},\mathbf{Y})= \frac{\lambda}{2}\sum_{i,j}\sum_{k\in \omega_j}\\
& \bigg\{\bigg[\big([{\cal{G}}^h(\mathbf{Z}_i)]_{j,k}- a^{h}_{i,j}[{\cal{G}}^h(\mathbf{Y})]_{j,k}-c^{h}_{i,j}\big)^2  + \epsilon (a^{h}_{i,j})^2\bigg]  \\	
& + \bigg[\big([{\cal{G}}^v(\mathbf{Z}_i)]_{j,k}- a^{v}_{i,j}[{\cal{G}}^v(\mathbf{Y})]_{j,k}-c^{v}_{i,j}\big)^2  + \epsilon (a^v_{i,j})^2\bigg]\bigg\},
\end{aligned}
\end{equation}
where ${\cal{G}}^h(\cdot)$ and ${\cal{G}}^v(\cdot)$ are functions that compute the horizontal and vertical gradient of the input image, respectively, i.e., ${\cal{G}}^{h}$$({\mathbf{Z}}_i)={\mathbf{Z}}_i\circledast \mathbf{S}^{h}$, with $\mathbf{S}^{h}=[-1~~1]$ and ${\cal{G}}^{v}$$({\mathbf{Z}}_i)={\mathbf{Z}}_i\circledast \mathbf{S}^{v}$, with $\mathbf{S}^v= [-1~~1]^{\top}$; $a^{h}_{i,j}$ and $c^{h}_{i,j}$, $a^{v}_{i,j}$ and $c^{v}_{i,j}$ are both constant coefficients of the linear affine function in window $\omega_j$, corresponding to the horizontal and vertical gradient of the $i^{th}$ band. The major drawback of $\mathrm{LGC}$ lies in that its flawed regularization on 2-D high-frequency components. Since the 2-D high-frequency components whose 1-D absolute frequencies are beyond $\frac{\pi}{2}$ (shown as the intersection of the horizontally and vertically shaded regions in Fig.~\ref{fig:2d_frequency_aware}) will be more regularized than the non-intersected shaded region, two high-frequency components close to each other on the 2-D spectrum (e.g., A and B) are distinctly regularized. Therefore, $\mathrm{LGC}$'s 2-D high-frequency regularization is unreasonably non-smooth.

The power of $\mathrm{LLP}$ lies in multiple aspects:
\begin{enumerate}[label=\arabic*)]

\item Complimentary Regularization: the regularization on the high-frequency domain avoids the HRMS image to bear a resemblance to the PAN image in terms of low-frequency components, which can compromise the effectiveness of the data fidelity term in~\eqref{eqn:objective2}.
\item Adaption: our regularizer will favor the high-frequency details in the HRMS image that are aligned with the PAN image in terms of location and direction, and will not force the unaligned details in the HRMS image to be close to the details in the PAN image.
\item Smoothly-varying Regularization Strength: $\mathrm{LLP}$ will enforce 2-D high-frequency components that are close on the 2-D spectrum to have close regularization strictness. 
\item Short Support: $\mathrm{LLP}$'s kernel is small enough to allow flexible regularization on fine details.
\end{enumerate}

We will demonstrate the superiority of $\mathrm{LLP}$ over the related regularizers in Section IV.

\section{Pansharpening Algorithm} 
\subsection{Solving the Blur Kernel}
\label{subsec:blur_kernel_solve}

Variables $\mathbf{u}$ and $\bm{\omega}$ in~\eqref{eqn:objective1} can be simultaneously solved by the alternating direction method of multipliers (ADMM)~\cite{MinTao_ADMM} with respect to both variables at the same time. However, we choose to separately solve $\mathbf{u}$ and $\bm{\omega}$. The major reason is that our solver for $\mathbf{u}$ is accelerated by the FFT, which necessitates the coefficient matrix of $\mathbf{u}$, (i.e. corresponding to $\mathbf{\nabla}\mathbf{u}$ in~\eqref{eqn:reg1}) be Toeplitz. When combining $\mathbf{u}$ and $\bm{\omega}$ as an individual vector, the corresponding coefficient matrix will not be Toeplitz and the FFT acceleration is no longer applicable. 

Therefore, we decouple~\eqref{eqn:objective1} into two sequential sub-problems: solving $\bm{\omega}$ and solving $\mathbf{u}$. To estimate $\bm{\omega}$, we take the same approach as~\cite{simoes2014convex}. To mitigate the effect of not knowing $\mathbf{u}$, we blur $\mathbf{X}^{\prime}\bm{\omega}$ and $\mathbf{D}\mathbf{B}(\mathbf{u})\mathbf{Y}$ with strong blur kernels. Specifically, we solve the following convex problem for estimating $\bm{\omega}$:
\begin{equation}\label{eqn:objective3}
\underset{\bm{\omega}}{\operatorname{min}}\frac{1}{2}\|\mathbf{B}(\mathbf{u}_1)\mathbf{X}^{\prime}\bm{\omega}-\mathbf{D}\mathbf{B}(\mathbf{u}_2)\mathbf{Y}\|^2_\mathrm{2}+\mathrm{R}_{3}(\bm{\omega}),
\end{equation}
where $\mathbf{u}_1\in\mathbb{R}^{(2l+1)^2 \times 1}$ and $\mathbf{u}_2\in\mathbb{R}^{(cl+1)^2 \times 1}$ are two vectorized unit gain 2-D box filters with the window size of $(l+1)\times(l+1)$ and $(cl+1)\times(cl+1)$, respectively. $\mathbf{u}_1$ and $\mathbf{u}_2$ are low-pass filters with non-trivial $l$ to make sure $\mathbf{u}$ will trivially influence the estimation of $\bm{\omega}$. $\mathrm{R}_{3}(\bm{\omega}) = (\lambda_\omega)/(2)\|\mathbf{\nabla}\bm{\omega}\|_2^2$ penalizes the difference between consecutive entries of $\bm{\omega}$ to enforce neighboring MS bands to have similar weights. $\mathbf{\nabla}$ is an operator that computes the differences and $\lambda_\omega$, a scalar, is the regularization parameter. $\nabla$ operator is effectively pre-multiplying $\bm{\omega}$ by $\mathbf{G}\in \mathbb{R}^{{(N^{\prime}-1) \times N^{\prime}}}$, a matrix extracting the gradients between the weights of adjacent bands. \eqref{eqn:objective3} can be solved with a closed-form solution. After solving $\bm{\omega}$, we estimate $\mathbf{u}$ via solve the following problem:
\begin{equation}
\label{eqn:u_solve}
\begin{aligned}
\underset{\mathbf{u}, \mathbf{p}}{\operatorname{min}} &  \frac{1}{2}\|\mathbf{D}\mathscr{C}({\mathbf{Y}})\mathbf{u}-\mathbf{X}^{\prime}\bm{\omega}\|_{2}^{2} + \\ & \alpha_1\|\mathbf{\nabla}\mathbf{u}-\mathbf{p}\|_{2,1} +  \alpha_{2}\|\mathcal{E}(\mathbf{p})\|_{2,1}+ \mathbf{I}_\mathbb{S}(\mathbf{u}),
\end{aligned}
\end{equation}
where we use $\mathscr{C}(\mathbf{Y})\mathbf{u}$ to express $\mathbf{B}(\mathbf{u})\mathbf{Y}$ since $\mathbf{B}(\mathbf{u})\mathbf{Y}=\mathbf{u}\circledast\mathbf{Y}=\mathbf{Y}\circledast\mathbf{u}=\mathscr{C}(\mathbf{Y})\mathbf{u}$. $\mathscr{C}(\mathbf{Y})$ is a $HW\times n^2$ matrix, each row of which stores $\mathbf{Y}$'s pixels that convolve with the blur kernel $\mathbf{u}$ to generate a pixel in the blurred $\mathbf{Y}$.

To solve~\eqref{eqn:u_solve}, we introduce constraints $\mathbf{x}=\mathbf{\nabla}\mathbf{u}-\mathbf{p}$, $\mathbf{y}=\mathcal{E}(\mathbf{p})$, and $\mathbf{z}=\mathbf{u}$, and apply the classical augmented Lagrangian method by minimizing
\begin{equation}
\label{eqn:Lagrangian}
\begin{aligned}
 &~~~{\Phi}(\mathbf{u},\mathbf{p},\mathbf{x},\mathbf{y},\mathbf{z},\bm{\Lambda_1},\bm{\Lambda_2},\bm{\Lambda_3}) = 
\frac{1}{2}\|\mathbf{D}\mathscr{C}({\mathbf{Y}})\mathbf{z}-\mathbf{X}^{\prime}\bm{\omega}\|_{2}^{2}   \\ &
 +\alpha_{1}\|\mathbf{x}\|_{2,1} +\frac{\alpha_{1}\mu_{1}}{2}\left\|\mathbf{x}-(\mathbf{\nabla}\mathbf{u}-\mathbf{p})-{\bm{\Lambda}_1}\right\|_{\mathrm{F}}^{2} \\ & +\alpha_{2}\|\mathbf{y}\|_{2,1}+\frac{\alpha_{2}\mu_{2}}{2}\left\|\mathbf{y}-\mathcal{E}(\mathbf{p})-{\bm{\Lambda}_2}\right\|_{\mathrm{F}}^{2} \\ &
 +\mathbf{I}_{\mathbb{S}}(\mathbf{z})+\frac{{\mu}_{3}}{2}\left\|\mathbf{z}-\mathbf{u}-{\bm{\Lambda}_3}\right\|_{2}^{2},
\end{aligned}
\end{equation}
where $\mu_1, \mu_2, \mu_3>0$ are scalars. We solve the problem using the generalized ADMM~\cite{deng2016global} by alternating between a succession of minimization steps and update steps. After applying the generalized ADMM, we arrive at the following algorithms:
\begin{equation}
\label{eqn:subproblems}
\begin{cases}
\begin{aligned}
& \mathbf{x}^{t+1}  = \underset{\mathbf{x}}{\operatorname{argmin}}\|\mathbf{x}\|_{2,1} + \frac{\mu_{1}}{2}\left\|\mathbf{x}-(\mathbf{\nabla}\mathbf{u}^t-\mathbf{p}^t)-{\bm{\Lambda}^t_1}\right\|_{\mathrm{F}}^{2}		 \\
& \mathbf{y}^{t+1} =  \underset{\mathbf{y}}{\operatorname{argmin}}\|\mathbf{y}\|_{2,1}+\frac{\mu_{2}}{2}\left\|\mathbf{y}-\mathcal{E}(\mathbf{p}^t)-{\bm{\Lambda}^t_2}\right\|_{\mathrm{F}}^{2} \\
& \mathbf{z}^{t+1}  =\underset{\mathbf{z}}{\operatorname{argmin}}  \frac{1}{2}\left\|\mathbf{D} \mathscr{C}(\mathbf{Y})\mathbf{z}-\mathbf{X}^{\prime}\bm{\omega}\right\|_{2}^{2} + \\ & ~~~~~~~~~~\frac{\mu_{3}}{2}\left\|\mathbf{z}-\mathbf{u}^{t}-\mathbf{\Lambda}_{3}^{t}\right\|_{2}^{2}+\mathbf{I}_{\mathbf{S}}(\mathbf{z}) \\
& (\mathbf{u}^{t+1}, \mathbf{p}^{t+1})  = \underset{\mathbf{u},\mathbf{p}}{\operatorname{argmin}}\frac{\alpha_1\mu_1}{2}\left\|\mathbf{x}^t-(\mathbf{\nabla}\mathbf{u}-\mathbf{p})-{\bm{\Lambda}^t_1}\right\|_{\mathrm{F}}^{2} + \\ & ~~~~~~~~~~~~~~~~~~~\frac{\alpha_2\mu_{2}}{2}\left\|\mathbf{y}^t-\mathcal{E}(\mathbf{p})-{\bm{\Lambda}^t_2}\right\|_{\mathrm{F}}^{2} \\
& {\bm{\Lambda}}^{t+1}_1  =  {\bm{\Lambda}}^{t}_1 + \rho(\mathbf{\nabla}\mathbf{u}^{t+1}-\mathbf{p}^{t+1}-\mathbf{x}^{t+1}) \\
& {\bm{\Lambda}}^{t+1}_2  =  {\bm{\Lambda}}^{t}_2 + \rho(\mathcal{E}(\mathbf{p}^{t+1})-\mathbf{y}^{t+1}) \\
& {\bm{\Lambda}}^{t+1}_3  =  {\bm{\Lambda}}^{t}_3 + \rho(\mathbf{u}^{t+1}-\mathbf{z}^{t+1}), \\
\end{aligned}
\end{cases}
\end{equation}	
where $\rho$ is a constant that controls the convergence. For fixed positive $\mu_1$, $\mu_2$, $\mu_3$, we typically choose $0 < \rho < (1 + \sqrt{5})/(2)$ to guarantee convergence based on the convergence analysis in~\cite{deng2016global} and~\cite{glowinski1984numerical}.
	
The $\mathbf{x}$- and $\mathbf{y}$- subproblems are similar to each other and the solutions are given by component-wise soft-thresholding. The $l$~th row of $\mathbf{x}^{t+1}$ and $\mathbf{y}^{t+1}$ are updated using
\begin{align}
\mathbf{x}^{t+1}(l) = & \operatorname{shrink}_{2}(\mathbf{\nabla}\mathbf{u}^{t}(l)-\mathbf{p}^{t}(l)+\bm{\Lambda}^{t}_1(l),\frac{1}{{\mu}_1}), \label{eqn:x_update}  \\ 
\mathbf{y}^{t+1}(l) = & \operatorname{shrink}_{2}(\mathcal{E}(\mathbf{p}^t)(l)+\bm{\Lambda}^{t}_2(l),\frac{1}{\mu_2}),  \label{eqn:y_update} 
\end{align}
where \(\operatorname{shrink}_{2}(\mathbf{e}, t)= \max{\left(\|\mathbf{e}\|_{2}-t,0\right) \frac{\mathbf{e}}{\|\mathbf{e}\|_{2}}}\).

To solve the $\mathbf{z}$-subproblem, we first solve
\begin{equation} \label{eqn:z_update}
\underset{\mathbf{z}}{\operatorname{min}} \frac{1}{2}\left\|\mathbf{D} \mathscr{C}(\mathbf{Y})\mathbf{z}-\mathbf{X}^{\prime}\bm{\omega}\right\|_{2}^{2} +  \frac{\mu_{3}}{2}\left\|\mathbf{z}-\mathbf{u}^{t}-\mathbf{\Lambda}_{3}^{t}\right\|_{2}^{2}.
\end{equation} Specifically, we first generate a $hw\times n^2$ matrix $\mathbf{E}=\mathbf{D}\mathscr{C}({\mathbf{Y}})$ and a $hw\times 1$ vector $\mathbf{f}=\mathbf{X}^{\prime}\bm{\omega}$ in the first regularization term. Then,~\eqref{eqn:z_update} can be solved through
\begin{equation} \label{eqn:z_solution}
\mathbf{z}=(\mathbf{E}^{\top}\mathbf{E}+\mu_3 \mathbf{I})^{-1}[\mathbf{E}^{\top}\mathbf{f}+\mu_3(\mathbf{u}^t +\mathbf{\Lambda}^t_3)].
\end{equation}

To enforce $\mathbf{z}$ in simplex $\mathbb{S}$, we projects $\mathbf{z}$ onto $\mathbb{S}$ with the algorithm's details in~\cite{wang2013projection}.

The $\{\mathbf{u},\mathbf{p}\}$-subproblem, that is, the fourth $\operatorname{argmin}$ problem in~\eqref{eqn:subproblems}, can be solved using the first-order necessary conditions~\cite{guo2014new}. The detailed derivation can be found in Appendix A. After introducing the solutions for all the subproblems in~\eqref{eqn:subproblems}, we conclude the algorithm for solving the blur kernel in Algorithm~\ref{alg:blur_kernel}.
\begin{algorithm}[h] 
\caption{Blur Kernel Estimation}
\label{alg:blur_kernel}
\begin{algorithmic}[1]
\REQUIRE ~~\\
1. Input MS image $\mathbf{X}^{\prime}$ and PAN image $\mathbf{Y}$. \\
2. Choose $\lambda_\omega$, $R$, $l$, $\alpha_1$, $\alpha_2$, $\mu_1$, $\mu_2$, $\mu_3$, $\rho$, $\epsilon$, $t_{max}$, $\mathrm{th}$.\\
3. Initialize $\bm{\Lambda}^0_1$, $\bm{\Lambda}^0_2$, $\bm{\Lambda}^0_3$, $\mathbf{x}^0$, $\mathbf{y}^0$, $\mathbf{z}^0$, $\mathbf{u}^0$, $\mathbf{p}^0$.  \\
 \hspace{-20 pt} $\bm{\omega}$ is given by solving~\eqref{eqn:objective3}. \\
 \hspace{-20 pt} \textbf{For $t=0,1,2,\ldots, t_{max}$}, run the following computations: \\
  1.  $\mathbf{x}^{t+1}$ is given by~\eqref{eqn:x_update}.\\
  2.  $\mathbf{y}^{t+1}$ is given by~\eqref{eqn:y_update}.\\
  3.  $\mathbf{z}^{t+1}$ is given by~\eqref{eqn:z_solution} and the algorithm in~\cite{wang2013projection}.\\
  4.  $\mathbf{u}^{t+1}$ and $\mathbf{p}^{t+1}$ are given by~\eqref{eqn:u_p_solution}.\\
  5. ${\bm{\Lambda}}^{t+1}_1  =  {\bm{\Lambda}}^{t}_1 + \rho(\mathbf{\nabla}\mathbf{u}^{t+1}-\mathbf{p}^{t+1}-\mathbf{x}^{t+1})$. \\
  6. ${\bm{\Lambda}}^{t+1}_2  =  {\bm{\Lambda}}^{t}_2 + \rho(\mathcal{E}(\mathbf{p}^{t+1})-\mathbf{y}^{t+1})$. \\
  7. ${\bm{\Lambda}}^{t+1}_3  =  {\bm{\Lambda}}^{t}_3 + \rho(\mathbf{u}^{t+1}-\mathbf{z}^{t+1})$. \\
 $\mathbf{Until}:$ $\|\mathbf{u}^{t}-\mathbf{u}^{t+1}\|_{\mathrm{2}} /\|\mathbf{u}^{t+1}\|_{\mathrm{2}} < \mathrm{th}$. \\
 \ENSURE~~\\
 blur kernel $\mathbf{u}$.
 \end{algorithmic}
 \end{algorithm}
\subsection{Solving the HRMS Image}
Given the estimated blur kernel $\mathbf{u}$, solving the HRMS image $\mathbf{Z}$ can be implemented by solving each channel in a parallel fashion. Therefore, solving~\eqref{eqn:objective2} effectively solves the following problem for each channel
\begin{equation}
\label{eqn:pansharp}
\begin{aligned}
&\underset{\mathbf{Z}_i,\mathbf{A}_i,\mathbf{C}_i}{\operatorname{min}}  \frac{1}{2}\|\mathbf{D}\mathbf{B}(\mathbf{u})\mathbf{Z}_i-\mathbf{X}_i\|^2_2+ \\ & \frac{\lambda}{2}\sum_{j}\sum_{k\in \omega_j} \bigg[\big([{\cal{L}}(\mathbf{Z}_i)]_{j,k}- a_{i, j}[{\cal{L}}(\mathbf{Y})]_{j,k}-c_{i, j}\big)^2  + \epsilon (a_{i, j})^2\bigg],
\end{aligned}
\end{equation}
where $\mathbf{A}_i$, $\mathbf{C}_i$ are the vectors storing all $a_{i,j}$, $c_{i,j}$ corresponding to $\mathbf{Z}_i$.~\eqref{eqn:pansharp} is a multi-convex, non-convex problem. Fu et al.~\cite{fu2019variational} proposed to use a Fast Iterative Shrinkage-Thresholding Algorithm (FISTA)~\cite{beck2009fast} to address this problem. This solver can be computationally intensive since it often requires a large number of iterations. To shorten the runtime, we aim to compute a better initial estimate of $\mathbf{Z}_i$ (i.e. $\mathbf{Z}^0_i$) to save the number of iterations. Generally speaking, the closer $\mathbf{Z}^0_i$ to the solution of~\eqref{eqn:pansharp}, the fewer iterations are needed.

To efficiently compute $\mathbf{Z}^0_i$, we aim to estimate $\mathbf{Z}^0_i$ by solving a problem with a closed-form expression as a function only of $\mathbf{u}$, $\mathbf{X}_i$ and $\mathbf{Y}$, without involving $\mathbf{A}_i$, $\mathbf{C}_i$. Fortunately, this closed-form expression exists based on Theorem~$\mathrm{1}$ from~\cite{levin2007closed} which states that the minimum of the expression
\begin{equation}
\label{eqn:pansharp_penalty}
J(\bm{\alpha}, \mathbf{a}, \mathbf{c})=\sum_{j\in \bm{I}}\sum_{k\in \omega_j} \bigg[	\big(\bm{\alpha}_{j,k}- \mathbf{a}_{j}\bm{I}_{j,k}-\mathbf{c}_{j}\big)^2  + \epsilon (\mathbf{a}_{j})^2\bigg]
\end{equation}
among all the choices of $\mathbf{a}$, $\mathbf{c}$, is $\bm{\alpha}^{\top} \mathbf{M} \bm{\alpha}$, where $k$ refers to the $k^{th}$ element within a window, $\mathbf{a}_{j}$ and $\mathbf{c}_{j}$ are scalars coefficients of the linear affine transform in window $\omega_j$ that maps the pixels in image $\bm{I}$ to another image $\bm{\alpha}$: $\bm{\alpha}_{i} \approx a \bm{I}_{i}+c, \forall i \in \omega$, $\mathbf{M}$ is a $HW \times HW$sparse matrix, where its $(m,n)$-th entry is ($1 \leqslant m \leqslant HW, 1 \leqslant n \leqslant HW$)
\begin{equation}\label{eqn:matting_matrix}
\sum_{k \mid(m, n) \in w_{k}}\left[\delta_{mn}-\frac{1}{\left|w_{k}\right|}\left(1+\frac{\left(\bm{I}_{m}-\mu_{k}\right)\left(\bm{I}_{n}-\mu_{k}\right)}{\frac{\epsilon}{\left|w_{k}\right|}+\sigma_{k}^{2}}\right)\right].
\end{equation}
Here, $\delta_{mn}$ is the Kronecker delta, $\mu_k$ and $\delta^2_k$ are the mean and variance of the intensities in the window $w_k$ around $k$, and $|w_k|$ is the number of pixels in this window. Because matrix $\mathbf{M}$ was originally proposed to solve the natural image matting problem, we call $\mathbf{M}$ the matting matrix of image $\bm{I}$. 

Equation~\eqref{eqn:pansharp_penalty} is the same as the second term in~\eqref{eqn:pansharp} except with different notations. Similar to~\eqref{eqn:pansharp},~\eqref{eqn:pansharp_penalty} minimizes the overall loss of approximating each block of image $\bm{\alpha}$ by the closest linear affine function of the co-located block in $\bm{I}$. The $\bm{\alpha}$ that minimizes $J(\bm{\alpha}, \mathbf{a}, \mathbf{c})$ is analogous to the ${\cal{L}}(\mathbf{Z}_i)$ that minimizes the second term in~\eqref{eqn:pansharp}. $\bm{\alpha}$ is constrained to be within the null space of matrix $\mathbf{M}$ and should be regularized via an extra constraint to have a unique solution. Therefore, we involve the data fidelity term in~\eqref{eqn:pansharp} as the constraint and solve $\mathbf{Z}^0_i$ by solving
\begin{equation}
\label{eqn:initialize_z_fast}
\underset{\mathbf{Z}^0_i}{\operatorname{min}} \frac{1}{2} \|\mathbf{D}\mathbf{B}(\mathbf{u}_0)\mathbf{Z}^0_i-\mathbf{X}_i\|^2_\mathrm{2} + 
\frac{\lambda}{2}(\mathbf{L}\mathbf{Z}^0_i)^{\top} \mathbf{M}_{\mathbf{L}\mathbf{Y}}(\mathbf{L}\mathbf{Z}^0_i),
\end{equation}
where ${\mathbf{L}} \in \mathbb{R}^{HW \times HW}$ denotes the Toeplitz matrix of the Laplacian operator and $\mathbf{M}_{\mathbf{L}\mathbf{Y}}$ is the matting matrix of the Laplacian of the PAN image. Equation~\eqref{eqn:initialize_z_fast} can be solved with a closed form expression: 
\begin{equation}\label{eqn:Z_ini_matting}
\mathbf{Z}^0_i= (\mathbf{B}^\top\mathbf{D}^\top\mathbf{D}\mathbf{B}+\lambda\mathbf{L}^\top\mathbf{M}_{\mathbf{L}\mathbf{Y}}\mathbf{L})^{-1}(\mathbf{B}^\top\mathbf{D}^\top\mathbf{X}_i).
\end{equation}
Here $\mathbf{I}_{HW}$ is a $HW\times HW$ identity matrix. $\mathbf{Z}^0_i$ in~\eqref{eqn:Z_ini_matting} is numerically solved by the conjugate gradient method. Since $\mathbf{M}_{\mathbf{L}\mathbf{Y}}$ is a sparse matrix, solving~\eqref{eqn:Z_ini_matting} is fast. We will verify its usefulness in initializing a closer approximation to the solution of $\mathbf{Z}_i$ in Section~\ref{subsec:speed_verify}.

Due to $\mathbf{Z}^0_i$'s close distance to the converged result, we use only one iteration involving solving $\mathbf{A}_i, \mathbf{C}_i$-subproblem and solving $\mathbf{Z}_i$-subproblem to generate the pansharpened $\mathbf{Z}$. 

\subsubsection{Solving $\mathbf{A}_i$, $\mathbf{C}_i$-subproblem}
The $\mathbf{A}_i, \mathbf{C}_i$-subproblem can be solved by guided image filtering. $a_{i,j}$, $c_{i,j}$ can be stably computed using ${\cal{L}}(\mathbf{Z}_i)$'s local window as the input image and ${\cal{L}}(\mathbf{Y})$'s local window as the guide image
\begin{equation}\label{eqn:ac_h_compute}
\begin{cases}
\begin{aligned}
a_{i,j} & =\frac{\frac{1}{(2r+1)^2} \sum_{k \in \omega_{j}} [{\cal{L}}(\mathbf{Y})]_{j,k}[{\cal{L}}(\mathbf{Z}^0_i)]_{j,k}-\mu_{i,j} \bar{p}_{i,j}}{(\sigma_{i,j})^{2}+\epsilon}  \\
c_{i,j} & =\bar{p}_{i,j}-a_{i,j} \mu_{i,j}, \\
\end{aligned}
\end{cases}
\end{equation}
where $\mu_{i,j}$ and $\sigma_{i,j}$ are the mean and stand deviation of ${\cal{L}}(\mathbf{Y})$ in the $j$~{th} window $\omega_j$. $\bar{p}_{i,j}$ is the mean of ${\cal{L}}(\mathbf{Z}_i)$ in $\omega_j$.

\subsubsection{Solving $\mathbf{Z}_i$-subproblem}
Given the solved $\mathbf{A}_i$, $\mathbf{C}_i$ from~\eqref{eqn:ac_h_compute}, we denote $\hat{\mathbf{L}}^{\mathbf{z}}_i$ as the output of guided image filtering using ${\cal{L}}(\mathbf{Y})$ as the guide image and using ${\cal{L}}(\mathbf{Z}_i)$ as the filtering input. Solving the $\mathbf{Z}_i$-subproblem is effectively solving
\begin{equation}
\label{eqn:Z_solution}
\underset{\mathbf{Z}_i}{\operatorname{min}}  \frac{1}{2}\|\mathbf{D}\mathbf{B}(\mathbf{u})\mathbf{Z}_i-\mathbf{X}_i\|^2_2+\frac{\lambda}{2}\|\mathrm{\mathbf{L}}\mathbf{Z}_i-\hat{\mathbf{L}}^{\mathbf{z}}_i\|^2_2,
\end{equation}
$\mathbf{Z}_i$ can be solved by the conjugate gradient method. To accelerate the solution, we take the approach of~\cite{zhao2016fast} and replace the iterations in the conjugate gradient method with a single iteration with a few FFT operations. In the interest of space, we refer the readers to~\cite{wei2015bayesian, zhao2016fast} for details.

We conclude the details of our fast approach in solving the pansharpening problem given the blur kernel in Algorithm~\ref{alg:fast_pansharpening}.

\begin{algorithm}[h] 
\caption{Fast Pansharpening}
\label{alg:fast_pansharpening}
\begin{algorithmic}[1]
\REQUIRE ~~\\
1. Input MS image $\mathbf{X}$, PAN image $\mathbf{Y}$ and blur kernel $\mathbf{u}$. \\
2. Choose $\lambda$, $r$, $\epsilon$.\\
3. Compute Matting Matrix $\mathbf{M}_{\mathbf{L}\mathbf{Y}}$ by solving~\eqref{eqn:matting_matrix}. \\
\hspace{-20 pt} \textbf{For $i=1,2,\ldots, N$}, do in parallel: \\
Compute $\mathbf{Z}^0_i$ via~\eqref{eqn:Z_ini_matting}. \\
Solve $\mathbf{A}_i$, $\mathbf{C}_i$ via~\eqref{eqn:ac_h_compute} given $\mathbf{Z}^0_i$ and $\mathbf{Y}$. \\
Solve $\mathbf{Z}_i$ via~\eqref{eqn:Z_solution} (accelerated by FFT) given $\mathbf{A}_i$, $\mathbf{C}_i$, $\mathbf{u}$.
\ENSURE ~~ \\
HRMS image: $\mathbf{Z}$.
\end{algorithmic}
\end{algorithm}
 
\subsection{Summary of the Blind Pansharpening Scheme}

Since the spatial misalignment between the LRMS and PAN images in remote sensing data are typically bounded to a few pixels, the estimation of $\bm{\omega}$ is sufficient to have a good estimation of the blur kernel and the pansharpened images. Refinement of $\bm{\omega}$ via~\eqref{eqn:objective1} given $\mathbf{u}$ provides marginal improvement on image quality. Therefore, the blind pansharpening scheme involves Algorithm~\ref{alg:blur_kernel} and~\ref{alg:fast_pansharpening} for only once without iterative loops. We summarize our scheme as:
\begin{enumerate}[label=\arabic*)]
    \item Solve the blur kernel via Algorithm~\ref{alg:blur_kernel};
    \item Solve the HRMS image via Algorithm~\ref{alg:fast_pansharpening}. 
\end{enumerate}

\section{Numerical Experiments}

\subsection{Dataset}

To evaluate the proposed approach, we run it on extensive datasets,~\textit{Pavia University},~\textit{Moffett},~\textit{Los Angeles},~\textit{Cambria Fire},~\textit{Chikusei}~\cite{NYokoya2016}, and~\textit{Stockholm}, from various remote-sensing platforms.
 
\subsubsection{\textit{Pavia University}} a 4-channel MS image (blue, green, red and infra-red) and a PAN image, both of spatial resolution $610 \times 338$. These images are synthesized from the \textit{Pavia University} dataset using the spectral response of the IKONOS satellite: each MS channel or PAN is generated by a weighted linear combination of bands of hyper-spectral imagery spanning the $430\sim 860$ $n$m spectral range from the ROSIS.

\subsubsection{\textit{Moffet},~\textit{Los Angeles},~\textit{Cambria Fire}} the dataset is from AVIRIS NASA airborne hyper-spectral dataset with a MS imager with 16 channels and a PAN image of spatial resolution of $512 \times 512$. These images are synthesized using the AVIRIS hyper-spectral image database so that each MS image channel is a weighted linear combination of hyper-spectral channels. 

\subsubsection{\textit{Chikusei}} the original Chikusei dataset is an airborne hyper-spectral dataset spanning $363\sim 1018$ $n$m and mosaiced by multiple images from the Headwall Hyperspec-VNIR-C imaging sensor. It involves $128$ bands of images, each of which has the spatial resolution of $2517 \times 2335$. Using the same technique as we synthesized~\textit{Pavia University} dataset, we synthesize a 4-channel MS image and a PAN image and crop three typical regions: manmade, grass, farmland, and a region of the mixture of these three regions, each of whose channels has $512 \times 512$ pixels. We name them as~\textit{Manmade}, ~\textit{Grass}, ~\textit{Farmland}, ~\textit{Mix} for notational convenience. 

\subsubsection{\textit{Stockholm}} the original dataset is from World View-2 satellite, downloaded from \url{digitalglobe-marketing.s3.amazonaws.com/product_samples/Stockholm_zipped/Stockholm_View-Ready_8_Band_Bundle_40cm.zip}, with $8$ MS bands of images (Coastal, Blue, Green, Yellow, Red, Red Edge, Near Infra-Red 1 and Near Infra-Red 2) of $1882\times2335$ spatial resolution and a PAN image of $9340\times7528$ spatial resolution. We crop $10$ pairs of $128\times 128$ LRMS and $512 \times 512$ PAN images as the dataset. We draw the reader's attention that the ground-truth HRMS image is not available. The quantitative metric on the pansharpening quality without the reference will be detailed in Section~\ref{subsec:metric}.


\subsection{Simulating the LRMS images}

The MS bands of images from~\textit{Pavia~University},~\textit{Moffett},~\textit{Los Angeles},~\textit{Cambria Fire} and~\textit{Chikusei} will be blurred and downsampled without adding noise to simulate the observed MS images. The motivation behind not adding noise is because in many satellite images, the physical size of each pixel are large and the Time Delay Integration (TDI)~\cite{holdsworth1990time} technique allows effective noise cancellation. In addition, we model the blur kernel as the convolution of a 2-D Gaussian with the motion blur. The Gaussian blur is due to the difference of the motion-free point spread functions corresponding to the PAN and the LRMS images and is modeled as follows
\begin{equation}\label{eqn:gaussian_blur}
\mathbf{g}(x,y)=\frac{1}{{\Sigma}_{g}}e^{-\frac{(x-c_x)^2+(y-c_y)^2}{2\sigma^2}},
\end{equation} 
where $\sigma$ is the standard deviation, $(x, y)$ is an integer coordinate, $(c_x, c_y)$ is the coordinate of the peak of Gaussian and carries the spatial misalignment information, the denominator $\Sigma_G$ is the sum of all the coefficients to make sure the blur kernel has unit gain. The motion blur is due to the difference between the scan duration of each line of the PAN sensor and the LRMS sensor. For the satellite platform where there exists a $\times 4$ resolution resolution gap between the PAN and the LRMS images along horizontal and vertical dimensions, scanning each line of LRMS sensor takes four times the time as scanning each line of the PAN sensor. Therefore, in this case, the motion blur is three-pixel wide. Likewise, the motion blur is one-pixel wide when the resolution gap is $\times 2$. Considering the possible rotation of satellites~\cite{jacobsen2008satellite}, we assume the direction along the satellite orbit not perpendicular to the scan line of LRMS/PAN sensors, but with a $\theta$ offset. Therefore, we model the motion blur as
\begin{equation}\label{eqn:motion_blur}
\mathbf{h}(x,y)=\begin{cases}
\begin{aligned}
&\frac{\delta(-x\sin{\theta}+y\cos{\theta})}{d} ,~\|x\cos{\theta}+y\sin{\theta}\|\leqslant \frac{d}{2} \\
& 0, \text{~otherwise,}
\end{aligned}
\end{cases}
\end{equation}
where $\delta$ is the Dirac delta function, $d$ is the width of the motion blur. The blur kernel $\mathbf{U}$ is the convolution of $\mathbf{g}$ and $\mathbf{h}$ and is expressed in~\eqref{eqn:blur_kernel_final} in Appendix B. We simulate the LRMS images when the downsampling rate is $2$ and $4$ whose corresponding $\sigma$ of $\mathbf{g}$ are $1$ and $2$, respectively. (For~\textit{Pavia~University}, we crop the bottom and right borders of each band of the ground-truth HRMS image to make sure its spatial resolution is $608 \times 336$ before downsampling by a factor of $4$.) Each set of simulated LRMS and PAN images will be fed into the algorithm to generate the HRMS image. For~\textit{Stockholm} without the ground-truth HRMS image we directly feed the observed LRMS and PAN images into the algorithm to generate the HRMS image. 

\subsection{Quantitative Metrics for Pansharpening Quality}
\label{subsec:metric}
To quantify the quality of pansharpening results, four metrics taken from the literature are used when a ground-truth HRMS image is available, as in the case for datasets~\textit{Pavia University},~\textit{Moffett},~\textit{Los Angeles},~\textit{Cambria Fire} and~\textit{Chikusei}. Notice that the intensities in all channels will be scaled within the range of $0$ to $255$.
The first index is Average Peak Signal-to-Noise Ratio ($\overline{\operatorname{PSNR}}$), which is the mean of PSNRs of all reconstructed HRMS channels, namely
\begin{align}
& \overline{\operatorname{PSNR}}(\reallywidehat{\mathbf{MS}},{\mathbf{MS}}) = \frac{1}{N}\sum_i^{N} \operatorname{PSNR}({\reallywidehat{\mathbf{MS}}_i},{\mathbf{MS}_i}),  \label{eqn:psnr_avg} \\
& \operatorname{PSNR}({\reallywidehat{\mathbf{MS}}_i},{\mathbf{MS}_i})= 20 \log_{10} \left(255/\sqrt{\mathrm{MSE}_i}\right), \label{eqn:psnr_def}
\end{align}
and $N$ is the number of spectral bands of a MS image, $\mathrm{MSE}_i = \frac{1}{HW}\left\|{\reallywidehat{\mathbf{MS}}_i}-{\mathbf{MS}_i}\right\|^2_2$, $\reallywidehat{\mathbf{MS}}_i, \mathbf{MS}_i \in \mathbb{R}^{HW \times 1}$ refer the $i$-th channel of estimated HRMS image and the $i$-th channel of ground-truth HRMS image, respectively. We also use Average Regressed PSNR ($\overline{\operatorname{PSNR}}_{reg}$)~\cite{gupta2018cnn} as an index to make a fair comparison only with a deep learning-based approach~\cite{lohit2019unrolled}
\begin{equation}\label{eqn:psnr_avg_reg} 
\overline{\operatorname{PSNR}}_{reg}(\reallywidehat{\mathbf{MS}},{\mathbf{MS}}) = \frac{1}{N}\sum_i^{N} \operatorname{PSNR}_{reg}({\reallywidehat{\mathbf{MS}}_i},{\mathbf{MS}_i}),   
\end{equation}
\begin{equation}\label{eqn:psnr_reg}
\operatorname{PSNR}_{reg}(\reallywidehat{\mathbf{MS}}_i, \mathbf{MS}_i) = \arg \max_{a, b} \operatorname{PSNR}(a \reallywidehat{\mathbf{MS}}_i+b, \mathbf{MS}_i).
\end{equation}

The rest three indices are Erreur Relative Globale Adimensionnelle de Synthèse (ERGAS)~\cite{wald2000quality}, Spectral Angle Mapper (SAM)~\cite{yuhas1992discrimination}, and Relative Average Spectral Error (RASE)~\cite{vivone2014critical}. Unlike the other three metrics, SAM mainly accounts for spectral distortion. When a ground-truth HRMS image is not available, as in the case for dataset~\textit{Stockholm}, we use $\overline{\operatorname{SSIM}}$, the mean of each HRMS channel's structural similarity index measure (SSIM)~\cite{wang2004image} with respect to the PAN image, as the metric to rate the quality of pansharpened MS images.
\begin{equation}\label{eqn:ssim}
\overline{\operatorname{SSIM}}(\reallywidehat{\mathbf{MS}}) = \frac{1}{N}\sum_i^{N} \operatorname{SSIM}(\reallywidehat{\mathbf{MS}}_i, \mathbf{PAN}).
\end{equation}

The reason we choose $\overline{\operatorname{SSIM}}$ as the metric is because this metric favors the HRMS image that is structurally similar to the PAN image. Note that we choose not to use the spatial distortion index $D_s$ in Quality with No Reference~\cite{alparone2008multispectral}(QNR) due to the unknown spatial misalignment. In addition, we choose not to use the the spectral distortion index $D_{\lambda}$, since it favors blurry HRMS images and is incapable of characterizing the high-frequency statistics

\subsection{Verification of the Blur Kernel Prior}

To demonstrate the advantage of our blur kernel regularizer over state-of-the-art blur kernel regularizers, we design an experiment that seeks to estimate blur kernels by modeling the observed low-resolution image as a noisy version of the blurred and downsampled ground-truth image (given). The reason we add noise in the measured image is to simulate the possible mismatches between the priors where the blind pansharpening algorithms depend upon and the actual statistics. We quantify each regularizer's performance in each experiment by comparing the relative error of the estimated blur kernel to the actual blur kernel. Namely,
\begin{equation}
\label{eqn:relative_error_kernel}
\epsilon_r=\frac{100\|\mathbf{U}-\hat{\mathbf{U}}\|_{\mathrm{F}}}{\|\mathbf{U}\|_{\mathrm{F}}}\%
\end{equation}
is the metric to quantify the reconstruction performance, where $\hat{\mathbf{U}}$ is the estimated blur kernel and ${\mathbf{U}}$ is the ground-truth blur kernel parameterized by~\eqref{eqn:blur_kernel_final}. In the experiment, we investigate widely-used regularizers~\cite{simoes2014convex,vivone2014pansharpening,bungert2018blind,bajaj2019blind} in the literature that cover $\ell_2$-based and $\mathrm{TV}$-based regularizers.

For the convenience of enumerating the regularizers' optimization functions in both experiments, we first denote the vectorized version of $\mathbf{U}$ as $\mathbf{u} \in \mathbb{R}^{(2R+1)^2\times 1}$. We also denote $\mathbf{E}$ in the experiment as a matrix where each row stores the pixels in the PAN image that convolve with the blur kernel and generates a noise-free pixel in the measurement image. We also denote $\mathbf{f}$ as the vectorized form of the measurement image. Those regularizers can be expressed as follows
\begin{align}
 &\mathrm{\ell_2 + NN}: \nonumber \\
 &\min_{\mathbf{u}} \frac{1}{2}\|\mathbf{E}\mathbf{u}-\mathbf{f}\|_{2}^{2}+\alpha_1 \|\mathbf{\nabla}\mathbf{u}\|^2_{2} + \alpha_2\mathbf{u}^{\top}\mathbf{u} + \mathbf{I}_\mathbb{S}(\mathbf{u}), \label{eqn:kernel_l2_semiblind} \\
 &\mathrm{TV + NN}: \nonumber  \\
 & \min_{\mathbf{u}} \frac{1}{2}\|\mathbf{E}\mathbf{u}-\mathbf{f}\|_{2}^{2}+\alpha \|\mathbf{\nabla}\mathbf{u}\|_{2,1}+\mathbf{I}_\mathbb{S}(\mathbf{u}), \label{eqn:kernel_tv} \\
 &\mathrm{{TGV}^2 + NN}: \nonumber  \\
  & \min_{\mathbf{u},\mathbf{p}} \frac{1}{2}\|\mathbf{E}\mathbf{u}-\mathbf{f}\|_{2}^{2}+\alpha_{1}\|\mathbf{\nabla} \mathbf{u}-\mathbf{p}\|_{2,1}+\alpha_{2}\|\mathcal{E}(\mathbf{p})\|_{2,1}+\mathbf{I}_\mathbb{S}(\mathbf{u}). \label{eqn:kernel_tgv}
\end{align}
Equation~\eqref{eqn:kernel_l2_semiblind} is based on a widely used $\ell_2$-based regularizer in the literature~\cite{simoes2014convex,vivone2014pansharpening}. $\|\mathbf{\nabla}\mathbf{u}\|^2_{2}$ is effectively $\|\mathbf{D}_h\mathbf{u}\|^2_{2}+\|\mathbf{D}_v\mathbf{u}\|^2_{2}$ where $\mathbf{D}_h, \mathbf{D}_v \in \mathbb{R}^{(2R+1)^2 \times (2R+1)^2}$ are matrices extracting the horizontal and vertical gradients of the blur kernel. This regularizer operates on the sum of the squared horizontal and vertical gradients to ensure the smoothness of the blur kernel. The penalty term on the sum of the squared coefficients is for tuning the steepness of the blur kernel. We also add the simplex constraint~\eqref{eqn:simplex}, $\mathbf{I}_\mathbb{S}(\mathbf{u})$, in~\eqref{eqn:kernel_l2_semiblind} to make sure all the coefficients are non-negative. For notational convenience, we denote the regularizer in~\eqref{eqn:kernel_l2_semiblind} as $\ell_2+\mathrm{NN}$, where $\mathrm{NN}$ stands for non-negative. The $\mathrm{TV}$-based regularizer in~\eqref{eqn:kernel_tv} penalizes the $\ell_1$ norm of the modulus of gradients of the blur kernel to pursue a smooth blur kernel. We denote it as isotropic total variation ($\mathrm{TV}$). For notational convenience, we denote the regularizer~\eqref{eqn:kernel_tv} as $\mathrm{TV}+\mathrm{NN}$. Similarly, our regularizer~\eqref{eqn:kernel_tgv} is denoted as $\mathrm{TGV}^{2}+\mathrm{NN}$.

We use $R=9$, $x=1.392$, $y=0.093$, $\sigma=2$, $d=3$ and $\theta = -13.7^{\circ}$ to initialize the blur kernel $\mathbf{U}$ via~\eqref{eqn:blur_kernel_final} for blurring the $540 \times 600$ PAN channel of~\textit{West of Sichuan} from IKONOS satellite due to the rich variety of image structures, edges, textures, smooth regions, etc. The blurred image will be downsampled by a factor of $4$, both horizontally and vertically, followed by adding Additive White Gaussian Noise, thereby generating observed images at $10$ dB, $20$ dB, $30$ dB, $40$ dB and $50$ dB PSNRs (PSNR is defined in~\eqref{eqn:psnr_def} using the blurred image as the ground truth.).

\begin{table}[!tbp]
\caption{Relative Error of Typical Blur Kernel Regularizers, $\ell_2+\mathrm{NN}$, $\mathrm{TV}+\mathrm{NN}$, $\mathrm{TGV}^2+\mathrm{NN}$, at Different Noise Levels. The Smallest Error at each PSNR Level is Highlighted in Bold.}
	\centering
    \setlength\tabcolsep{4 pt}
	\begin{tabular}{||c|c|c|c|c|c||}
		\hline
		 PSNR(dB) & 10 & 20 & 30 & 40 & 50  \\
		\hline
		$\ell_2+\mathrm{NN}$  & $28.52\%$ & $16.03\%$  &  $9.04\%$  &  $5.05\%$  & $2.75\%$ \\				
		\hline
		$\mathrm{TV}+\mathrm{NN}$  & $32.91\%$ & $19.18\%$  &  $9.73\%$  &  $5.14\%$  & $3.09\%$ \\
		\hline
	    $\mathrm{TGV}^2+\mathrm{NN}$  & $\mathbf{17.39}\%$  & $\mathbf{9.55}\%$ &  $\mathbf{5.15}\%$  & $\mathbf{2.90}\%$ & $\mathbf{1.68}\%$ \\
		\hline       
	\end{tabular}
	\label{tab:results2}	
\end{table}

\begin{figure*}[htb]
\centering
\begin{minipage}{0.24\linewidth}
\centerline{\includegraphics[width=0.95\linewidth]{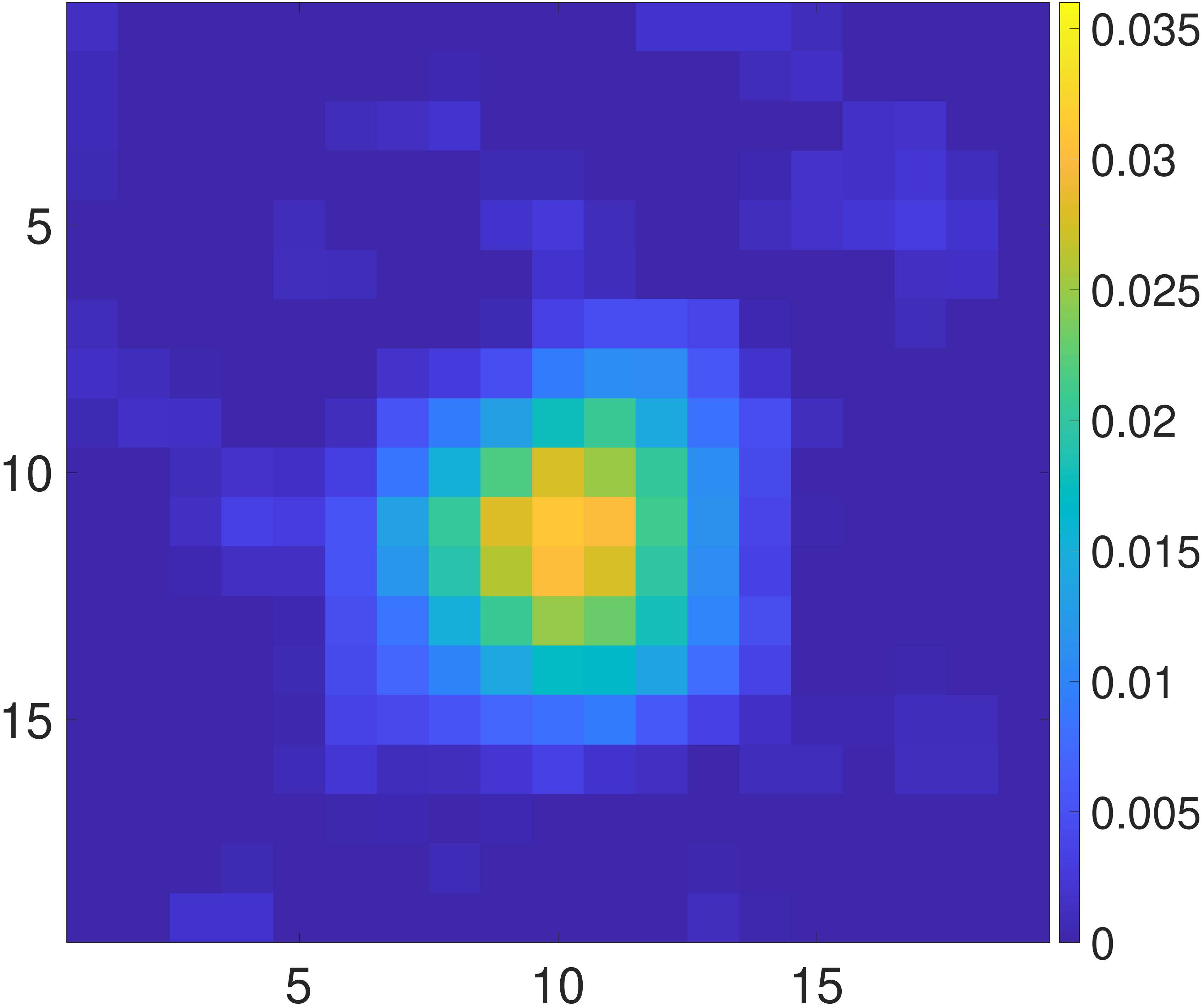}}
\centerline{(a) }\medskip
\end{minipage}
\centering
\begin{minipage}{0.24\linewidth}
\centerline{\includegraphics[width=0.95\linewidth]{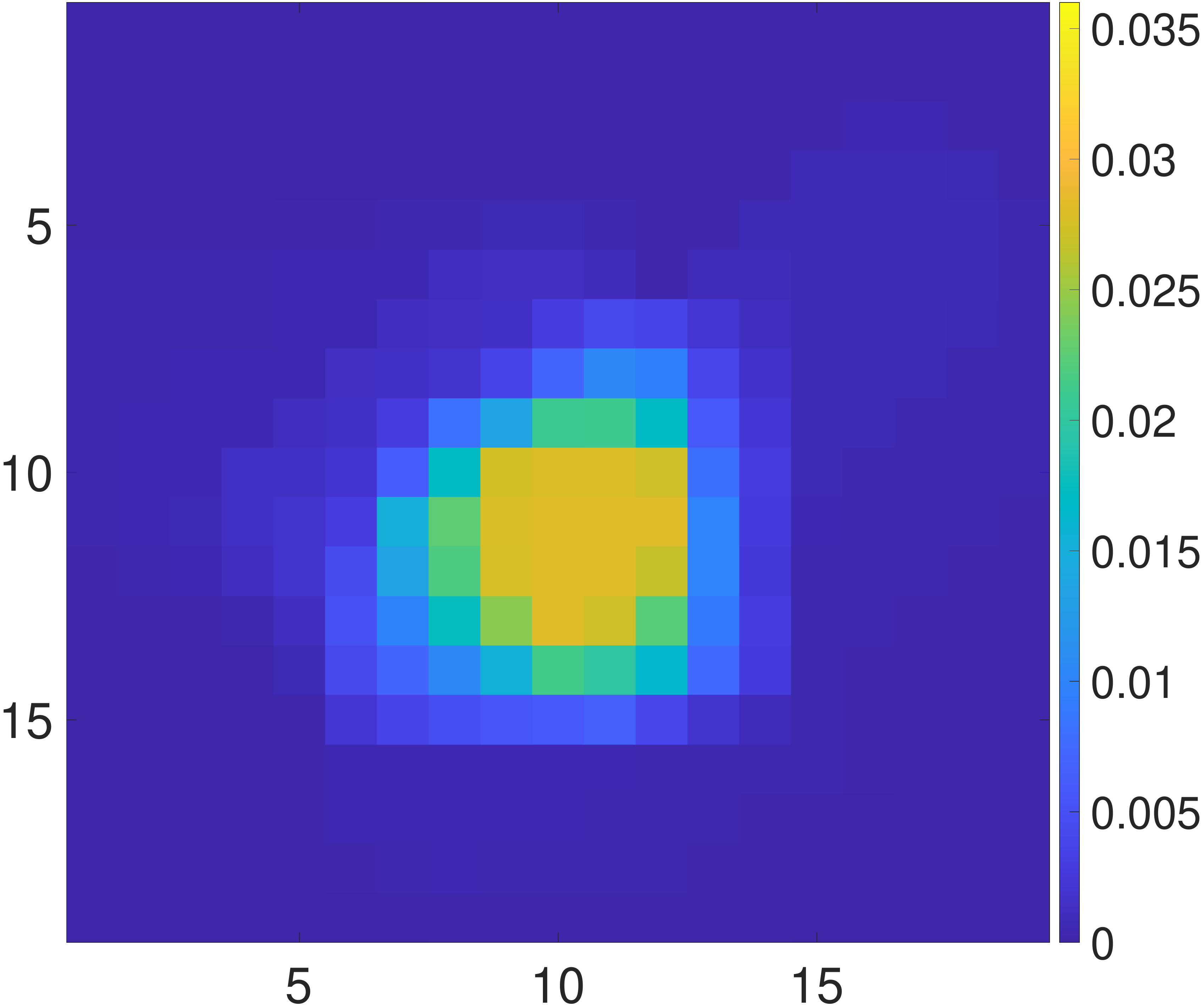}}
\centerline{(b) }\medskip
\end{minipage}
\begin{minipage}{0.24\linewidth}
\centerline{\includegraphics[width=0.95\linewidth]{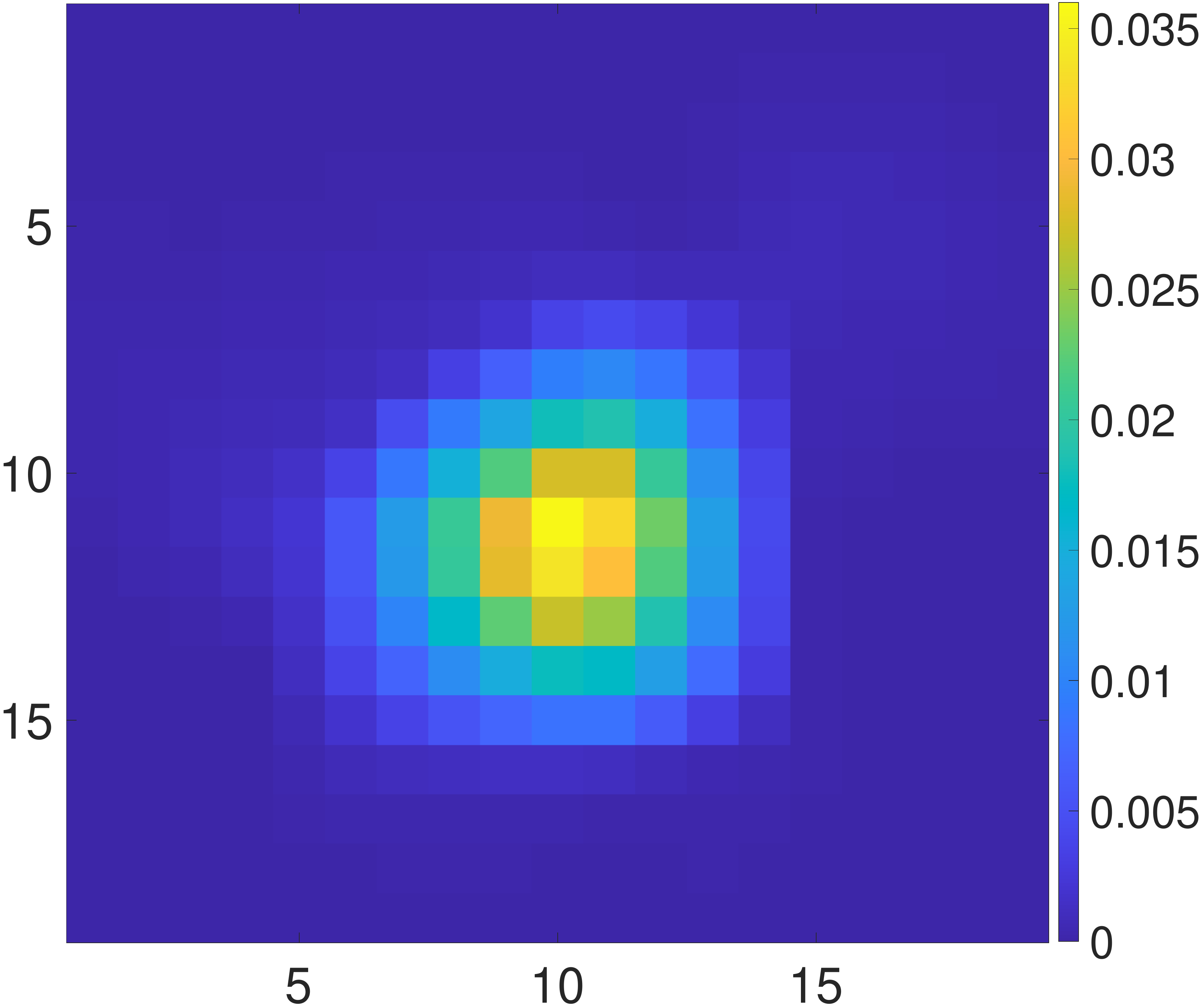}}
\centerline{(c) }\medskip
\end{minipage}
\begin{minipage}{0.24\linewidth}
\centerline{\includegraphics[width=0.95\linewidth]{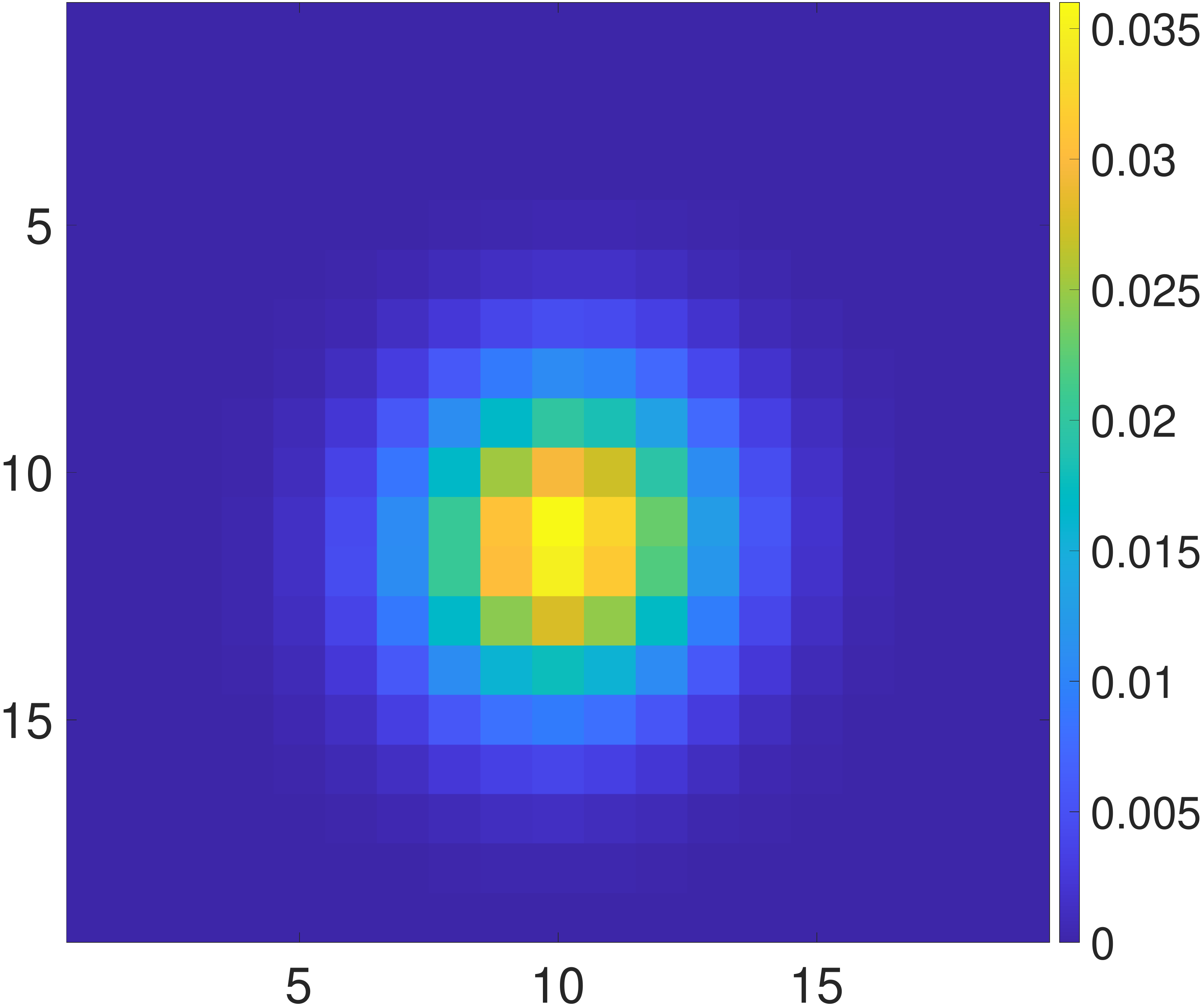}}
\centerline{(d) }\medskip
\end{minipage}
\caption{Comparisons of the estimated blur kernels from (a) $\ell_2+\mathrm{NN}$, (b) $\mathrm{TV}+\mathrm{NN}$,  (c) $\mathrm{TGV}^2+\mathrm{NN}$ regularizers with ground-truth blur kernel (d) when the $\mathrm{PSNR}$ of the blurred and downsampled~\textit{West~of~Sichuan} image is $20$ dB. Those three regularizers have the relative estimation error: $16.03\%$, $19.18\%$ and $9.55\%$, respectively.}
\label{fig:results_kernel}
\end{figure*}

Table~\ref{tab:results2} lists the relative errors for three regularizers in all noise levels, with the smallest error in each case highlighted in bold. Each experiment's parameters have been fine-tuned to ensure the smallest relative error. As evident, $\mathrm{TGV}^2$ recovers the best estimate of the ground truth kernel and maintains the highest robustness to noise among all the three regularizers.

We illustrate the estimated blur kernels from three regularizers at 20-dB PSNR level in Fig.~\ref{fig:results_kernel}. The $\ell_2$-based regularizer in (a) has non-zero coefficients beyond outskirts of the ground truth kernel, which is an undesired feature for a blur kernel. Due to the sparsity of $l_1$ norm, the $\mathrm{TV}$-based regularizer addressed the undesired feature of $\ell_2$-based regularizers. However, its estimated blur kernel [See Fig.~\ref{fig:results_kernel}(b)] has approximately $0$-gradients near the peak of the blur kernel and a sharp transition surrounding the peak, which is inconsistent to the smoothness of the ground-truth blur kernel in (d). Our $\mathrm{TGV}^2+\mathrm{NN}$ is more suitable to regularize the blur kernel than $\mathrm{TV}+\mathrm{NN}$ and $\ell_2+\mathrm{NN}$, since it preserves the overall smoothness of the blur kernel: it neither forces the small gradients of the blur kernel to be $0$ as $\mathrm{TV}+\mathrm{NN}$ does nor generates non-trivial coefficients far from the blur kernel's peak as $\ell_2+\mathrm{NN}$ does. In real blind pansharpening experiments, $\mathrm{TGV}^2+\mathrm{NN}$ also has its superiority over $\mathrm{TV}+\mathrm{NN}$ and $\ell_2+\mathrm{NN}$ in Table~\ref{tab:result_2x} and~\ref{tab:result_4x} with the smallest relative errors $\epsilon_r$.

\subsection{Verification of Avoiding Bad Local Minima}

To demonstrate our algorithm's effectiveness in fusing misaligned images, we conduct two experiments on~\textit{Pavia University}~dataset and set the kernel's center to be $(0.87,0.11)$ and $(5.87,4.11)$ to simulate small and large misalignments, respectively, with the same standard deviation $\sigma = 1$, motion displacement $d = 1$, and $\theta=36.1^{\circ}$. The LRMS image is generated from the ground-truth HRMS image by applying the blur kernel and downsampling by a factor of $2$, both horizontally and vertically. We compare our blur kernel with the blur kernel from Graph Laplacian Regularization (GLR)~\cite{bajaj2019blind}, a blind image fusion approach using a 2-D Dirac delta function as the initial estimate of the blur kernel.
\begin{table}[!tb]
\caption{Quantitative analysis of blind pansharpening results when spatial misalignments are $(0.87,0.11)$ and $(5.87,4.11)$, respectively, in the task of pansharpening by a factor of $2$ in~\textit{Pavia University} dataset.}
\centering
\begin{tabular}{||c|c|c||}	
\hline
Approach & GLR  & Proposed  \\
\hline
$\overline{\operatorname{PSNR}}/\mathrm{dB}$ & 36.77/21.44 &  39.34/39.32  \\
\hline
$\epsilon_r/\%$ & 84.32/135.13 &  0.86/2.61 \\
\hline
\end{tabular}
\label{tab:psnr_misalignment}	
\end{table}
 
\begin{figure*}[!tb]
\centering
\begin{minipage}{0.325 \linewidth}
  \centering
  \centerline{\includegraphics[width=0.9\linewidth]{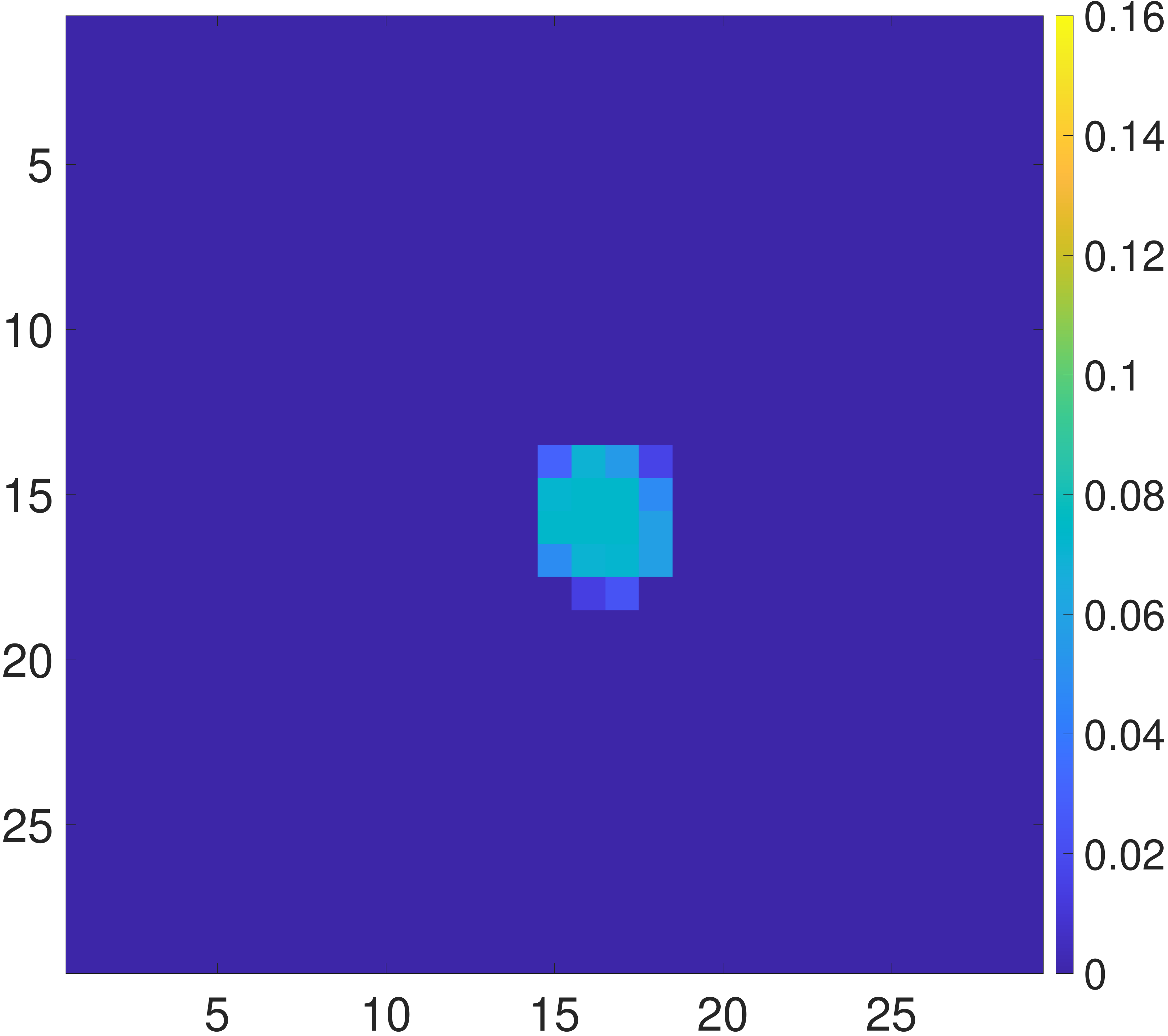}}
  \centerline{(a)} \medskip
  \end{minipage}
\begin{minipage}{0.325\linewidth}
  \centering
  \centerline{\includegraphics[width=0.9\linewidth]{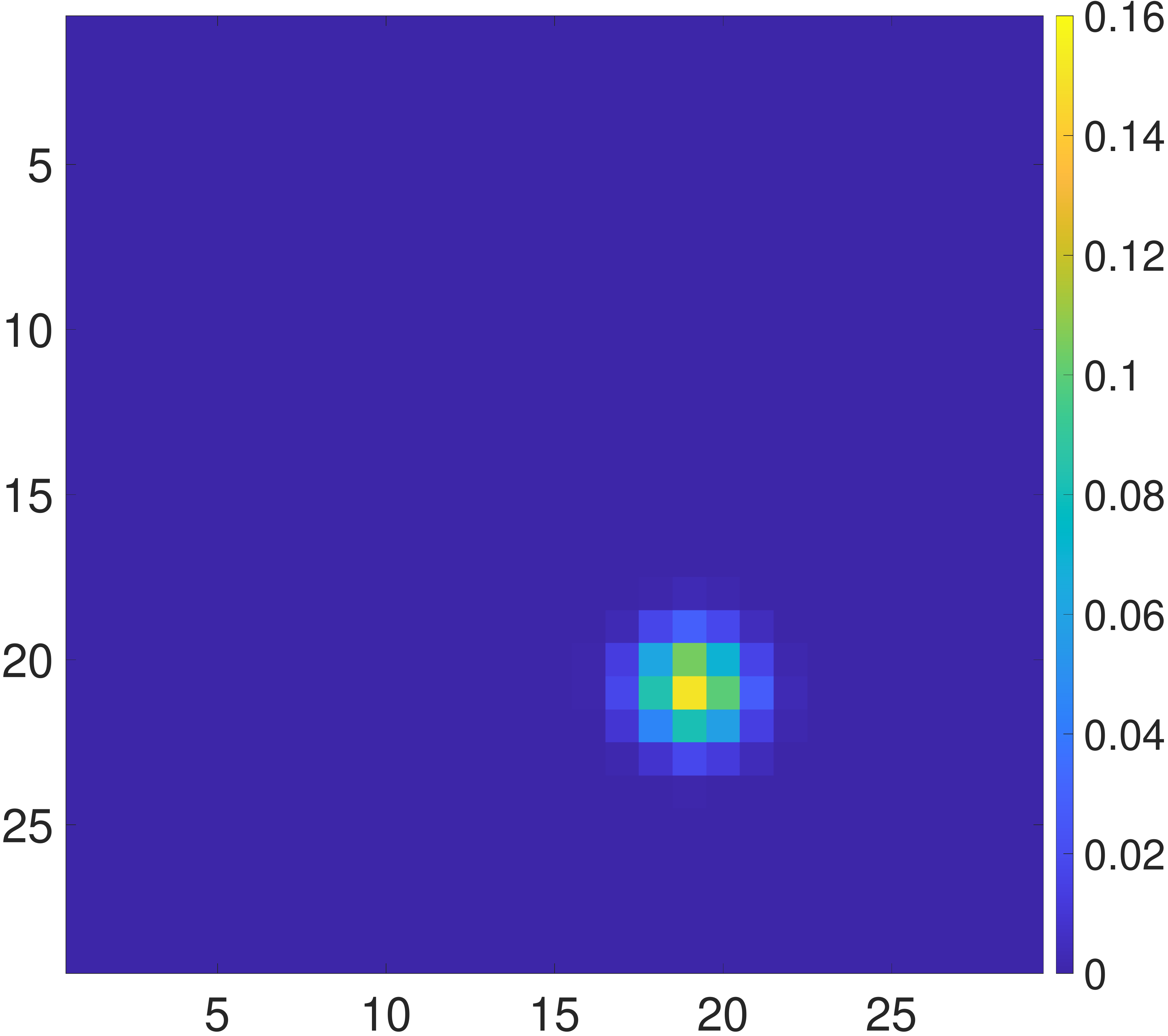}}
  \centerline{(b)}\medskip
\end{minipage}
\begin{minipage}{0.325\linewidth}
  \centering
  \centerline{\includegraphics[width=0.9\linewidth]{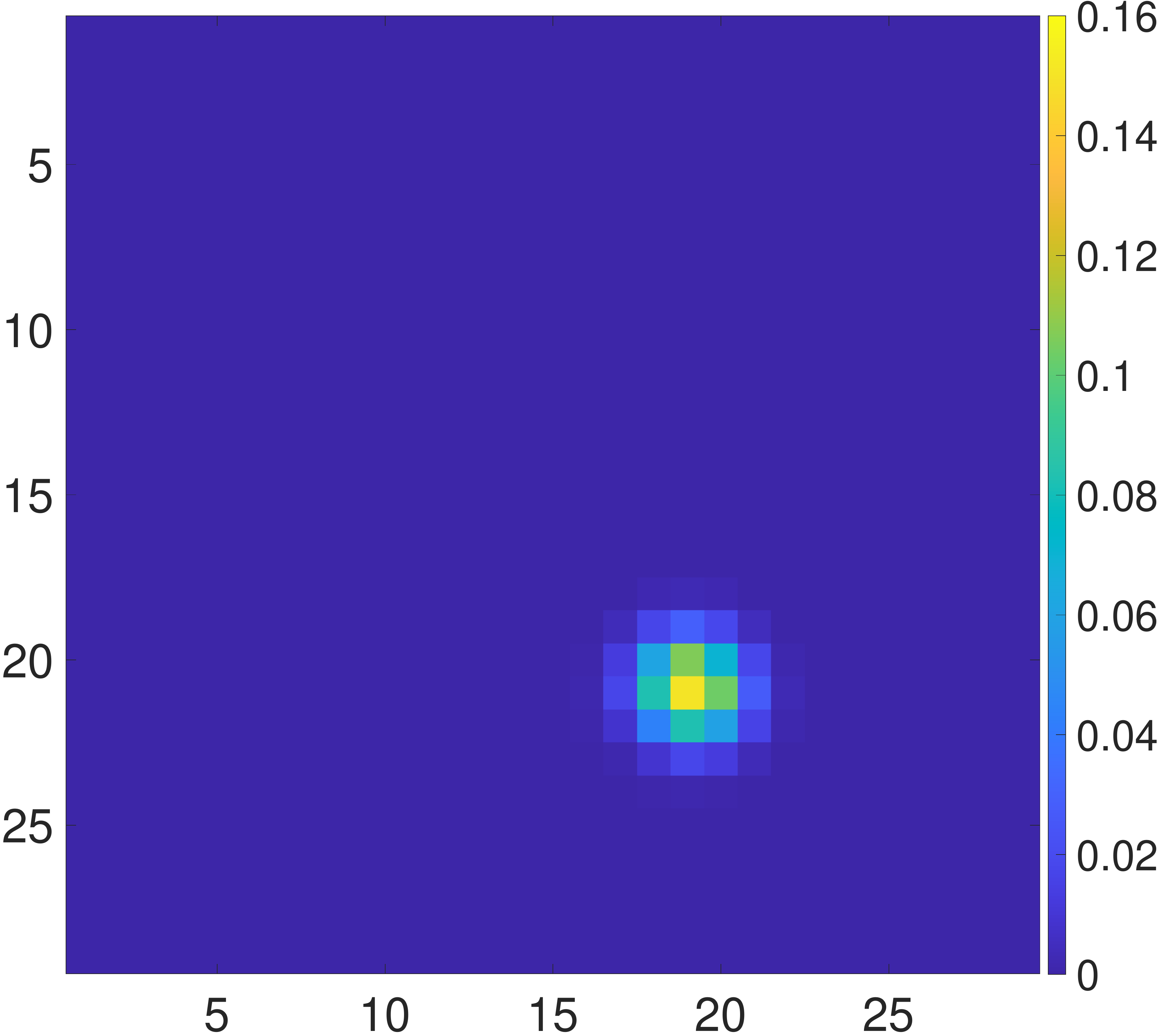}}
  \centerline{(c)}\medskip
\end{minipage}
\caption{Comparison of estimated blur kernels in the task of pansharpening~\textit{Pavia University} by a factor of $2$. (a) The estimated blur kernel from GLR. (b) The estimated kernel from the proposed approach. (c) The ground-truth blur kernel.}
\label{fig:kernel_results}
\end{figure*}

Quantitative performance comparisons of two algorithms are shown in Table~\ref{tab:psnr_misalignment}. Our algorithm generates consistent $\overline{\operatorname{PSNR}}$ both at small and large misalignments. In comparison, when the misalignment is large, GLR fails to generate a similar $\overline{\operatorname{PSNR}}$ as it generates when the misalignment is small. The blur kernel estimated when the offset is $(5.87,4.11)$, shown in Fig.~\ref{fig:kernel_results}(a), was trapped in a bad local minimum which is far away from the ground truth, shown in Fig.~\ref{fig:kernel_results}(c). This is due to the poor initial estimate of the blur kernel. Similar to~\cite{simoes2014convex}, our approach enforces the blurred and downsampled version of the PAN image to be close to the linear combination of its corresponding LRMS channels. Therefore, since the misalignment information has already been encoded in the blur kernel coefficients, the blur kernel can be directly estimated via solving~\eqref{eqn:u_solve}.
 
\subsection{Verification of the Cross-Channel Prior}

To avoid the conflict with the regularizer on the low-frequency components of the target HRMS that forces the blurred and downsampled HRMS to be the input LRMS, we choose the regularizer on the cross-channel relationship to only constrain the relationship between the high-frequency components of a HRMS channel and the high-frequency components of the PAN image. We extract the high-frequency components by a 2-D Laplacian and denoted our regularizer as local Laplacian prior (LLP). This subsection demonstrates the superiority of Laplacian over alternative operators by comparing the performance of solving problems similar to~\eqref{eqn:pansharp} when the ground-truth blur kernel ($\mathbf{u}_g$) is given. Within the alternative regularizers, we first introduce a cross-channel image prior, local low-pass filtering prior ($\mathrm{LPF}$), which regularizes the low-frequency components of both channels by replacing $\mathcal{L}$ in~\eqref{eqn:pansharp} with convolving with the blur kernel, i.e. the lowpass filter $\mathbf{u}_g$. We also involve another prior in the comparison as local pixel prior ($\mathrm{LPP}$) which regularizes the pixels. For the high-pass filters, we introduce a cross-channel image prior as local high-pass filtering prior ($\mathrm{HPF}$). HPF regularizes the high-frequency components that is complementary to  $\mathbf{u}_g$ where $\mathcal{L}$ in~\eqref{eqn:reg2} is replaced with convolving with a high-pass filter which equals to $\mathbf{\delta}-\mathbf{u}_g$, where $\mathbf{\delta}$ is a 2-D Dirac delta function. Furthermore, we also involve the aforementioned prior, $\mathrm{LGC}$, in Section~\ref{subsec:cross_channel} with the expression of~\eqref{eqn:reg2_lgc}.

Given all the priors, we solve~\eqref{eqn:pansharp} via Algorithm~\ref{alg:fast_pansharpening} in scenarios of pansharpening by a factor of $2$ and $4$. To make a fair comparison, we fine-tune the parameters to make sure the output has the highest $\overline{\operatorname{PSNR}}$. We show each cross-channel regularizer's $\overline{\operatorname{PSNR}}$ on Table~\ref{tab:cross_channel_prior}. From the $\mathrm{LPF}$ column, we can tell that using low-pass coefficients to regularize the cross-channel relationship has the lowest $\overline{\operatorname{PSNR}}$. This is due to the conflict between the data fidelity term and the cross-channel relationship that both operate on the lowpass frequencies of HRMS. When replacing $\mathrm{LPF}$ with $\mathrm{HPF}$, the $\overline{\operatorname{PSNR}}$ grows drastically. It is worth mentioning that compared with $\mathrm{HPF}$, $\mathrm{LPP}$'s $\overline{\operatorname{PSNR}}$ significantly drops by $0.91$ dB in the $\times 2$ case, and by $0.32$ dB in the $\times 4$ case. This demonstrates the advantage of involving only high-pass coefficients when regularizing the cross-channel relationship. It is worth mentioning that the graph Laplacian regularization in GLR~\cite{bajaj2019blind} is closely related to $\mathrm{LPP}$ by replacing $\mathrm{LPP}$ with the regularizer similar to the second term in~\eqref{eqn:initialize_z_fast}. The gap between $\mathrm{LPP}$ and $\mathrm{LLP}$ partly explains our algorithm's performance over GLR.

$\mathrm{HPF}$'s corresponding high-pass filter has a uniform regularization strength on the 2-D high-frequency components and has $0.07$ dB $\overline{\operatorname{PSNR}}$ gain over $\mathrm{LGC}$ in the $\times 2$ case. However, compared with $\mathrm{LGC}$, $\mathrm{HPF}$ has a longer support when the blur kernel is smooth, which compromises the flexibility to regularize very localized structures and results in a $0.21$ dB $\overline{\operatorname{PSNR}}$ drop in the $\times 4$ case. $\mathrm{LGC}$'s corresponding high-pass filter has a short support so that each high-frequency coefficient is only a function of two pixel values. This allows the approximation in blocks to be localized enough. However, from the 2-D perspective, the 2-D high-frequency components whose 1-D absolute frequency are beyond $\frac{\pi}{2}$ shown as the intersection of the horizontally and vertically shaded regions in Fig.~\ref{fig:2d_frequency_aware} are distinctly regularized compared with the non-intersected shaded regions. Thus, two high-frequency components close to each other on the 2-D spectrum can have unreasonably distinct regularization strictness.

Our $\mathrm{LLP}$ avoids the drawbacks of both $\mathrm{HPF}$ and $\mathrm{LGC}$. It not only has a short filter support to make block-wise cross-channel approximation localized enough but also has a smoothly varying regularization strictness for 2-D high-frequency components, thereby generating the highest $\overline{\operatorname{PSNR}}$ both in the $\times 2$ case and in the $\times 4$ case among all the related cross-channel image priors.

\begin{table}[tb]
\caption{The Average PSNR (in decibels) of different cross-channel regularizers on  different datasets in $\times 2$ and $\times 4$ scenario. The results with the highest Average PSNR averaged on all the datasets are highlighted in bold.}
\centering
\begin{tabular}{||c|c|c|c|c|c||}
\hline
\textbf{Blur Kernel} & $\mathrm{LPF}$ & $\mathrm{HPF}$  & $\mathrm{LPP}$  & $\mathrm{LLP}$  & $\mathrm{LGC}$\\
\hline
$\times 2~(\sigma = 1)$ & 4.21  & 43.10  & 42.19 & $\mathbf{43.35}$ & 43.03 \\
\hline
$\times 4~(\sigma =2)$  & 4.26 & 37.01 & 36.69  & $\mathbf{37.25}$ & 37.22 \\
\hline
\end{tabular}
\label{tab:cross_channel_prior}
\end{table}

\subsection{Verification of Convergence Speed}
\label{subsec:speed_verify}
\begin{figure}[tb]
\centering
\begin{minipage}{0.7 \linewidth}
\centering
\centerline{\includegraphics[height=1.0\linewidth]{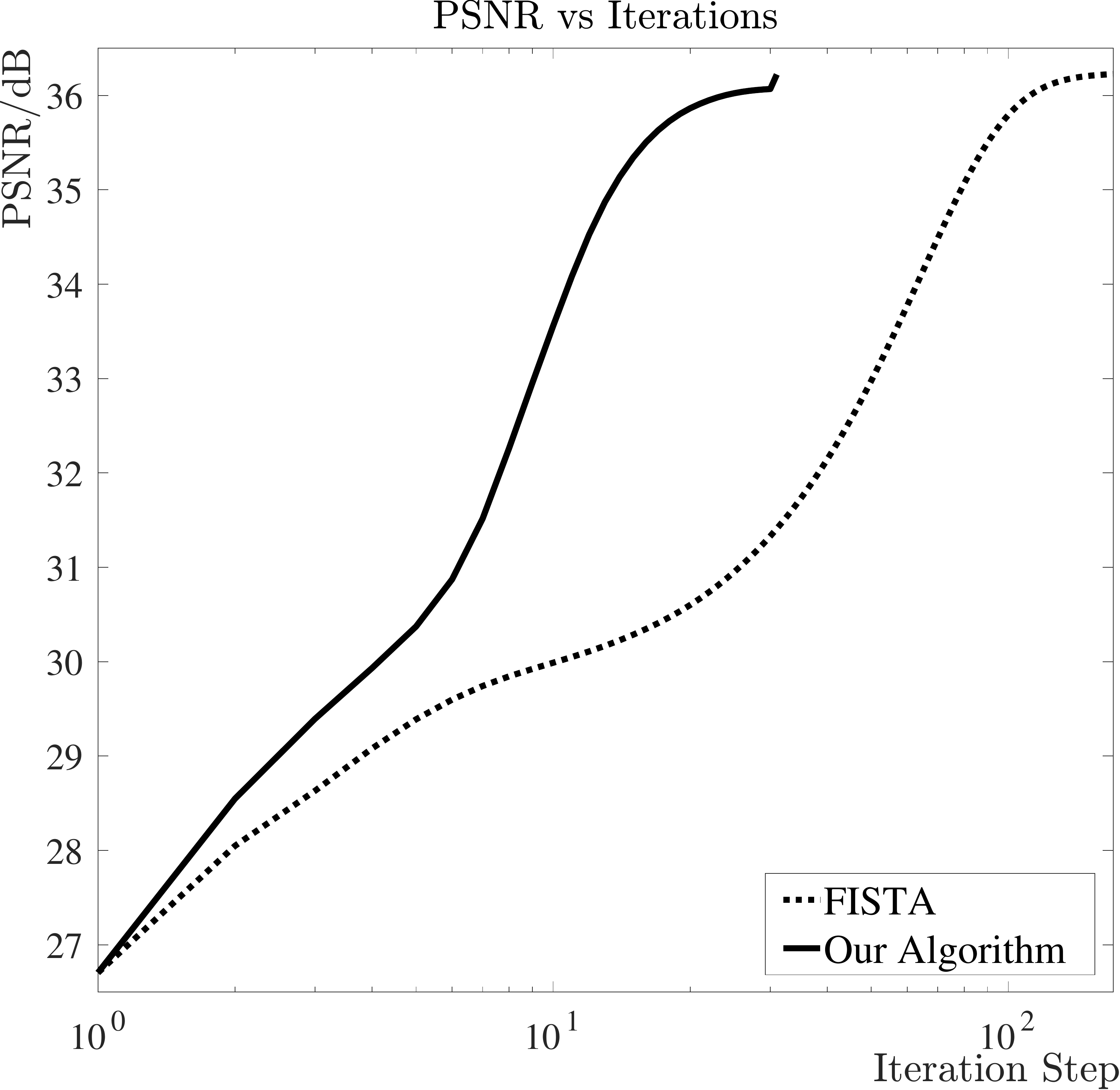}}
\end{minipage}
\caption{PSNR evolution in the process of sequentially solving~\eqref{eqn:Z_ini_matting} (solid curve) and solving~\eqref{eqn:Z_solution} using~\cite{zhao2016fast} (short straight solid segment in the last iteration) versus using FISTA (dashed curve) from~\cite{fu2019variational} in the task of pansharpening the seventh MS channel of~\textit{Los Angeles} by a factor of $2$.}
\label{fig:warmstart_convergence}
\end{figure}

To demonstrate the usefulness of our warmstart strategy, we record the evolution of the PSNR during the iterations when pansharpening the seventh MS channel of~\textit{Los Angeles} dataset by a factor of $2$, using the ground-truth kernel ($\sigma = 2, x = 0, y = 0, d = 1, \theta = -13.7^{\circ} $). In Fig.~\ref{fig:warmstart_convergence}, we compare these evolutions corresponding to Algorithm~\ref{alg:fast_pansharpening} and the FISTA algorithm for pansharpening in~\cite{fu2019variational}. The iterations of FISTA algorithm will stop when its PSNR is equal to or larger than ours after one iteration. 
Our algorithm generates a very fast convergence, with only a tiny fraction of the iterations of FISTA. What's more, the initial estimate of the pansharpened channel from our warmstart strategy (via solving~\eqref{eqn:Z_ini_matting}) is so close to the converged result (solution) that we only use a single iteration by solving each~\eqref{eqn:ac_h_compute} and~\eqref{eqn:Z_solution} in Algorithm~\ref{alg:fast_pansharpening} to get a closer approximation to the solution. Accelerating~\eqref{eqn:Z_solution} by a single iteration via FFT has a nontrivial PSNR increase as shown in Fig.~\ref{fig:warmstart_convergence} which justifies its usefulness.

\subsection{Implementation Details}

\subsubsection{Parameter Setting} 
We choose $\lambda_\omega = 10$ in~\eqref{eqn:objective3} to solve the spectral response $\bm{\omega}$, $\alpha_1 = 1$, $\alpha_2 = 0.006$, $\mu_1,\mu_2,\mu_3=100$, $\rho = 0.5$, $t_{max} = 10000$, $\mathrm{th}=0.00001$ in Algorithm~\ref{alg:blur_kernel} and $\lambda = 0.0002$, $r = 1$ for~\eqref{eqn:pansharp}. When solving~\eqref{eqn:Z_ini_matting} via the conjugated gradient method, we set the threshold as $5 \times 10^{-5}$ and when the relative change in the Frobenius norm is below the threshold we stop the iteration. In the task of pansharpening by a factor of $2$ and $4$, we set the blur kernel's size as $29\times 29$ to accommodate large spatial misalignments.

\subsubsection{Excluding Pixels Near the Boarder}

Due to circular convolution, the pixels near upper, lower, left and right boundaries will be involved in computing the Laplacian, which will capture spurious high-frequency components. Some anti-reflective operators~\cite{serra2004note, donatelli2006improved} manage to achieve better boundary approximations than circular convolutions via trigonometric matrix algebras. To exclude the influence of these high-frequency components in the metrics, we simply exclude the pixels within $10$ pixels distance to any of the four boundaries when computing $\overline{\operatorname{PSNR}}$, $\mathrm{ERGAS}$, $\mathrm{SAM}$ and $\mathrm{RASE}$.

\subsection{Experimental Results}

We dub our algorithm F-BMP (Fast and high-quality Blind Multi-spectral image Pansharpening). We compare F-BMP with current reproducible blind pansharpening algorithms in the literature: Hyperspectral Superresolution (HySure)~\cite{simoes2014convex}, Robust Fast Fusion of multi-band images based on solving a Sylvester equation (R-FUSE)~\cite{wei2016blind}, Graph Laplacian Regularization (GLR)~\cite{bajaj2019blind} and out previous work Blind pan-sharpening with local Laplacian prior and Total generalized variation prior (BLT)~\cite{yu2020blind}. The parameters are fine-tuned to achieve the highest $\overline{\operatorname{PSNR}}$. It is worth mentioning that recent pansharpening work, Joint Spatial-Spectral Smoothing in a Minimum-Volume Simplex for Hyperspectral Image Super-Resolution (JSMV-CNMF)~\cite{ma2020joint} is related to our approach. JSMV-CNMF demonstrates a longer runtime but a higher PSNR than HySure has. However, JSMV-CNMF does not take spatial misalignment into consideration. Due to the unavailable source code, we do not involve JSMV-CNMF in quantitative comparison. In addition, we draw the reader's conclusion that many alternative approaches~\cite{chen2015sirf,fu2019variational} have different degradation models which instead model a LRMS image as the bicubic interpolated version of its corresponding HRMS image. To pursue a fair comparison, we only make comparisons with~\cite{simoes2014convex,wei2016blind,bajaj2019blind,yu2020blind} since these approaches and our approach all model a LRMS image as the downsampled version of the blurred HRMS image through convolving with a blur kernel.

 \begin{table}[tb]
 \caption{Average Quantitative Metrics of Different Algorithms on Different Datasets in Pansharpening by a Factor of $2$ Task when the Misalignment is Small or Large. The Results Closest to the Ideal Value are Highlighted in Bold.}
 \setlength\tabcolsep{0.6 pt}
 \centering
 \begin{tabular}{||c|c|c|c|c|c|c|c|c|c|c||}
 \hline
 \multirow{2}{*}{\scriptsize{\textbf{Algorithm}}} &
 \multicolumn{10}{c||}{X2 (small misalignment\big|large misalignment)}  \\
 \cline{2-11}
 & \multicolumn{2}{c|}{$\overline{\operatorname{PSNR}}/\mathrm{dB}$} &\multicolumn{2}{c|}{$\mathrm{ERGAS}$} & \multicolumn{2}{c|}{$\mathrm{SAM}/^{\circ}$} & \multicolumn{2}{c|}{$\mathrm{RASE}$} & \multicolumn{2}{c||}{$\epsilon_r/\%$} \\
 \hline
  \scriptsize{\textit{HySure}}  & 33.30 & 33.22 & 10.89 & 10.97 & 4.96 & 4.99 & 15.17 & 15.26 &  7.76 &   9.16 \\
 \hline
   \scriptsize{\textit{R-FUSE}} & 33.77 & 25.67 &  9.70 & 23.96 & 3.19 & 8.49 & 14.99 & 36.77 & 11.97 &  15.89 \\
 \hline   
  \scriptsize{\textit{GLR}}     & 39.60 & 23.95 &  5.33 & 28.02 & 2.11 & 8.16 &  7.46 & 47.25 & 88.43 & 216.36 \\
  \hline
  \scriptsize{\textit{BLT}}     & 40.93 & 40.93 &  4.80 &  4.80 & 1.97 & 1.97 &  6.57 &  6.57 & 12.01 &  12.02 \\
  \hline
  \scriptsize{\textit{F-BMP}} & $\bm{43.29}$ & $\bm{43.18}$ & $\bm{3.85}$ & $\bm{3.87}$ & $\bm{1.59}$ &  $\bm{1.61}$ & $\bm{5.16}$ & $\bm{5.19}$ & $\bm{2.64}$ & $\bm{3.17}$\\  
 \hline
 \scriptsize{Ideal Value}  & \multicolumn{2}{c|}{$\infty$} & \multicolumn{2}{c|}{0}  &  \multicolumn{2}{c|}{0}  & \multicolumn{2}{c|}{0}  & \multicolumn{2}{c||}{0} \\    
 \hline
 \end{tabular}
 \label{tab:result_2x}
 \end{table}
 
  \begin{table}[tb]
  \caption{Average Quantitative Metrics of Different Algorithms on Different Datasets in Pansharpening by a Factor of $4$ Task when the Misalignment is Small or Large. The Results Closest to the Ideal Value are Highlighted in Bold.}
  \setlength\tabcolsep{0.6 pt}
  \centering
  \begin{tabular}{||c|c|c|c|c|c|c|c|c|c|c||}
  \hline
  \multirow{2}{*}{\textbf{Algorithm}} &
  \multicolumn{10}{c||}{X4 (small misalignment\big|large misalignment)}  \\
  \cline{2-11}
  & \multicolumn{2}{c|}{$\overline{\operatorname{PSNR}}/\mathrm{dB}$} &\multicolumn{2}{c|}{$\mathrm{ERGAS}$} & \multicolumn{2}{c|}{$\mathrm{SAM}/^{\circ}$} & \multicolumn{2}{c|}{$\mathrm{RASE}$} & \multicolumn{2}{c||}{$\epsilon_r/\%$} \\
  \hline
  \scriptsize{\textit{HySure}}  & 31.67 & 31.64 & 6.51 &  6.54 & 6.62 & 6.66 & 18.18 & 18.37 &  7.89 &   7.56 \\
  \hline
  \scriptsize{\textit{R-FUSE}}  & 33.73 & 27.68 & 5.16 & 10.29 & 4.43 & 8.59 & 14.51 & 28.66 &  7.90 &   7.59 \\
  \hline   
  \scriptsize{\textit{GLR}}     & 34.43 & 23.84 & 4.69 & 14.12 & 4.24 & 8.53 & 13.71 & 47.78 & 51.94 & 167.56 \\
  \hline
  \scriptsize{\textit{BLT}}     & 36.08 & 36.02 & 3.91 &  3.91 & 3.43 & 3.42 & 11.11 & 11.15 & 11.18 &  11.25 \\
  \hline
  \scriptsize{\textit{F-BMP}} & $\bm{37.19}$ & $\bm{37.17}$ & $\bm{3.58}$ & $\bm{3.59}$ &  $\bm{3.22}$ & $\bm{3.23}$ & $\bm{9.91}$ & $\bm{9.92}$ & $\bm{4.97}$ & $\bm{5.21}$ \\  
  \hline
  \scriptsize{Ideal Value}  & \multicolumn{2}{c|}{$\infty$} & \multicolumn{2}{c|}{0}  &  \multicolumn{2}{c|}{0}  & \multicolumn{2}{c|}{0}  & \multicolumn{2}{c||}{0} \\    
  \hline
  \end{tabular}
  \label{tab:result_4x}
  \end{table}

\begin{figure*}[!thb]
\centering
\begin{minipage}{0.19 \linewidth}
  \centering
  \centerline{\includegraphics[width=1.0\linewidth,height=1.0\linewidth]{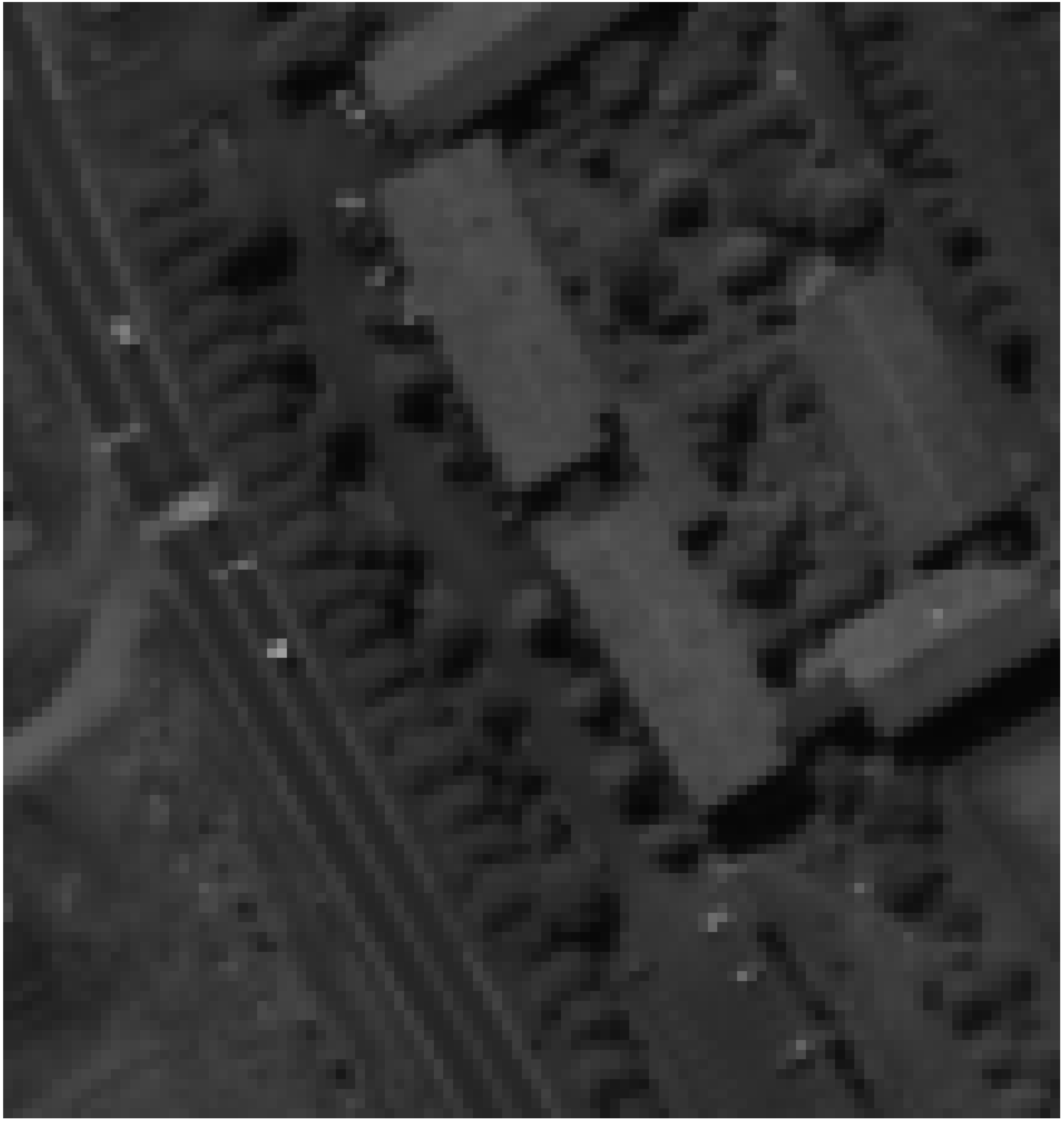}}
  \centerline{(a) }\medskip
\end{minipage}
\begin{minipage}{0.19 \linewidth}  
  \centering
  \centerline{\includegraphics[width=1.0\linewidth,height=1.0\linewidth]{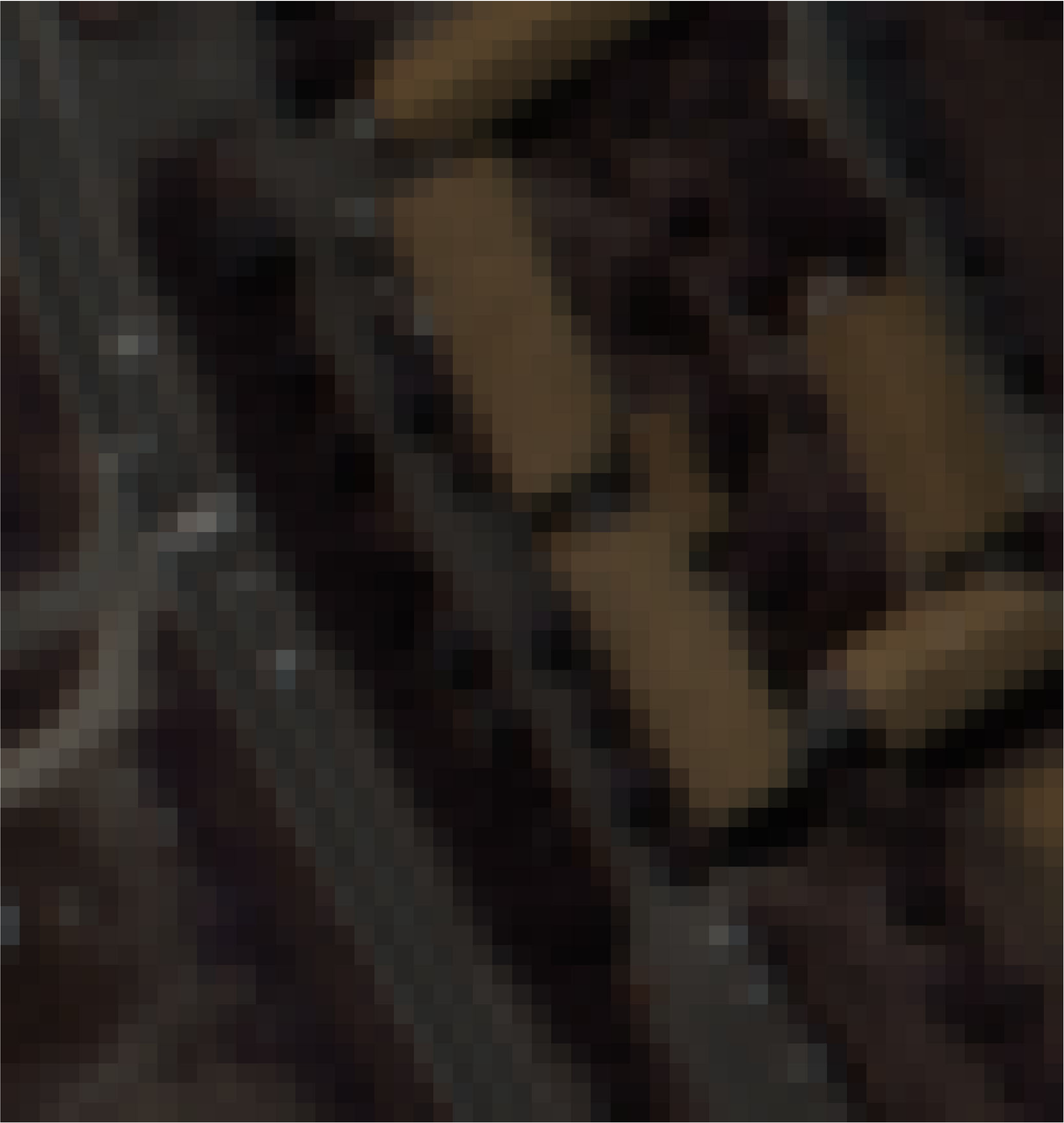}}
  \centerline{(b)} \medskip
  \end{minipage}  
  \begin{minipage}{0.19 \linewidth}
  \centering
  \centerline{\includegraphics[width=1.0\linewidth,height=1.0\linewidth]{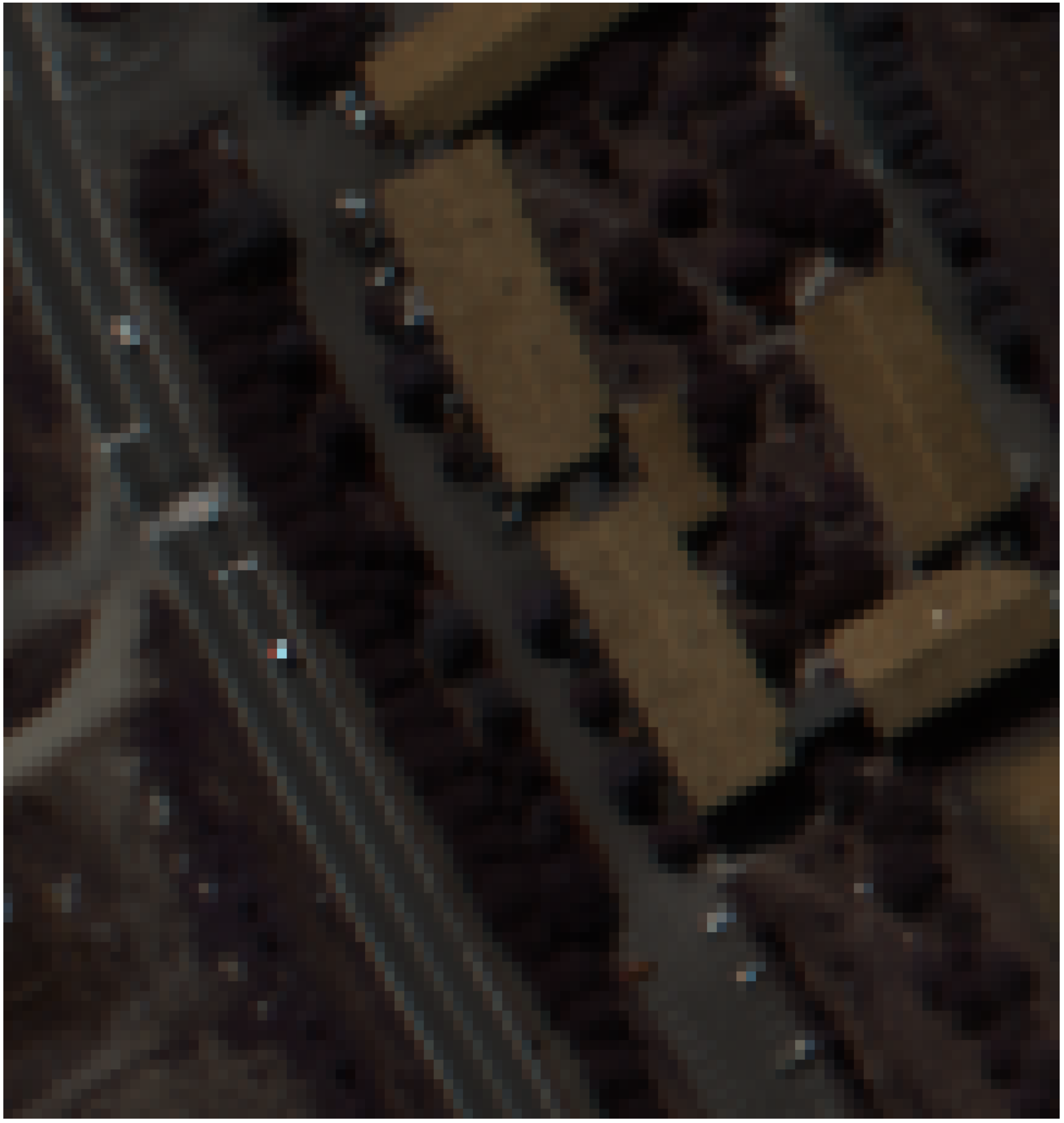}}
  \centerline{(c) }\medskip
\end{minipage}
\caption{The input images and the ground-truth HRMS image corresponding to the blind pansharpening experiment on~\textit{Pavia University}. (a) The PAN image. (b) The RGB channels of the LRMS image. Each pixel is enlarged by a factor of $2$, both horizontally and vertically to fit the space. (c) The RGB channels of the ground-truth HRMS image.}
\label{fig:results3}
\end{figure*}

\begin{figure*}[!thb]
\centering
\begin{minipage}{0.19 \linewidth}
  \centering
  \centerline{\includegraphics[width=1.0\linewidth,height=1.0\linewidth]{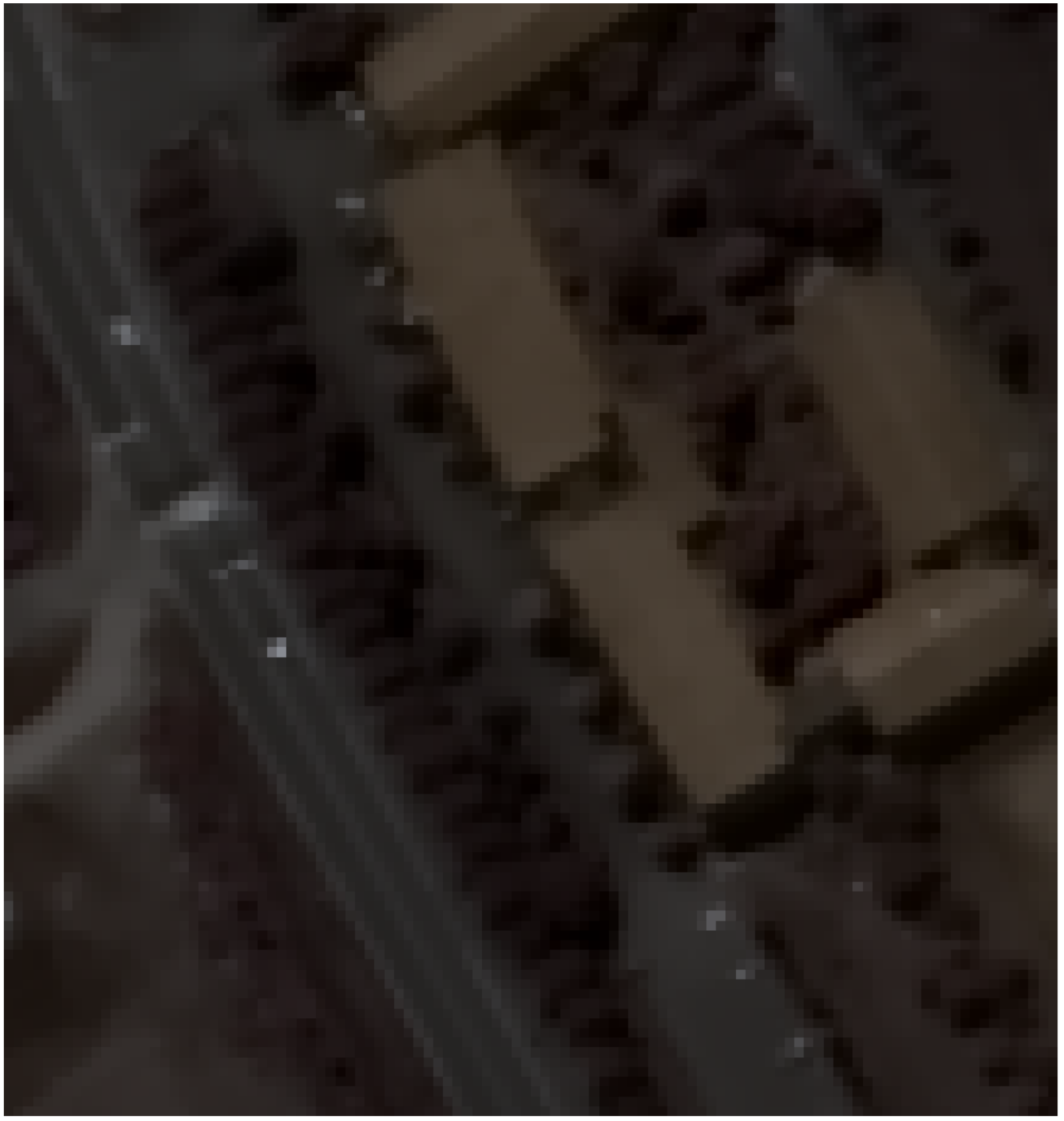}}
  \centerline{(a) }\medskip
\end{minipage}
\begin{minipage}{0.19 \linewidth}
  \centering
  \centerline{\includegraphics[width=1.0\linewidth,height=1.0\linewidth]{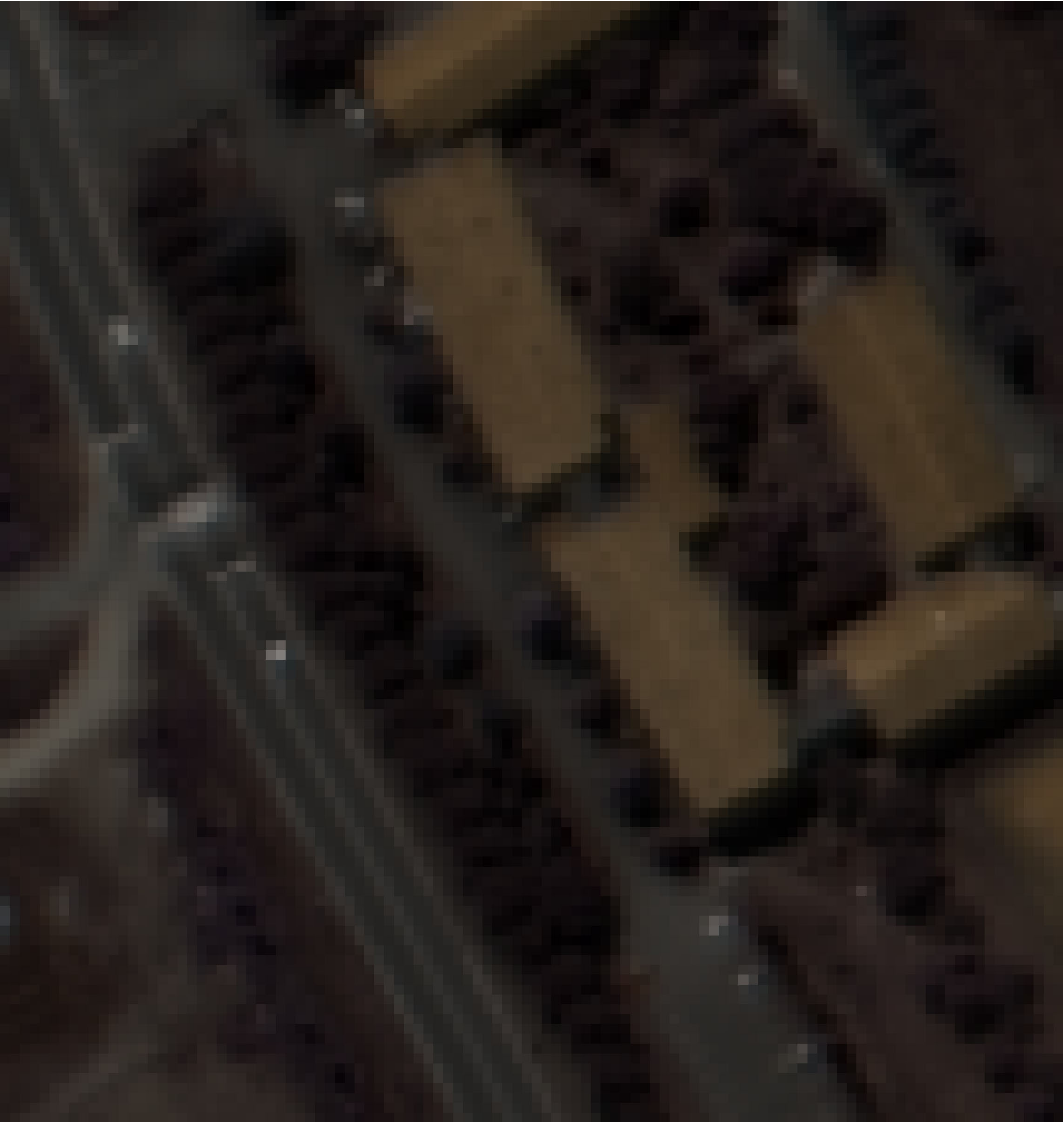}}
\centerline{(b) }\medskip
\end{minipage}
\begin{minipage}{0.19 \linewidth}
  \centering
  \centerline{\includegraphics[width=1.0\linewidth,height=1.0\linewidth]{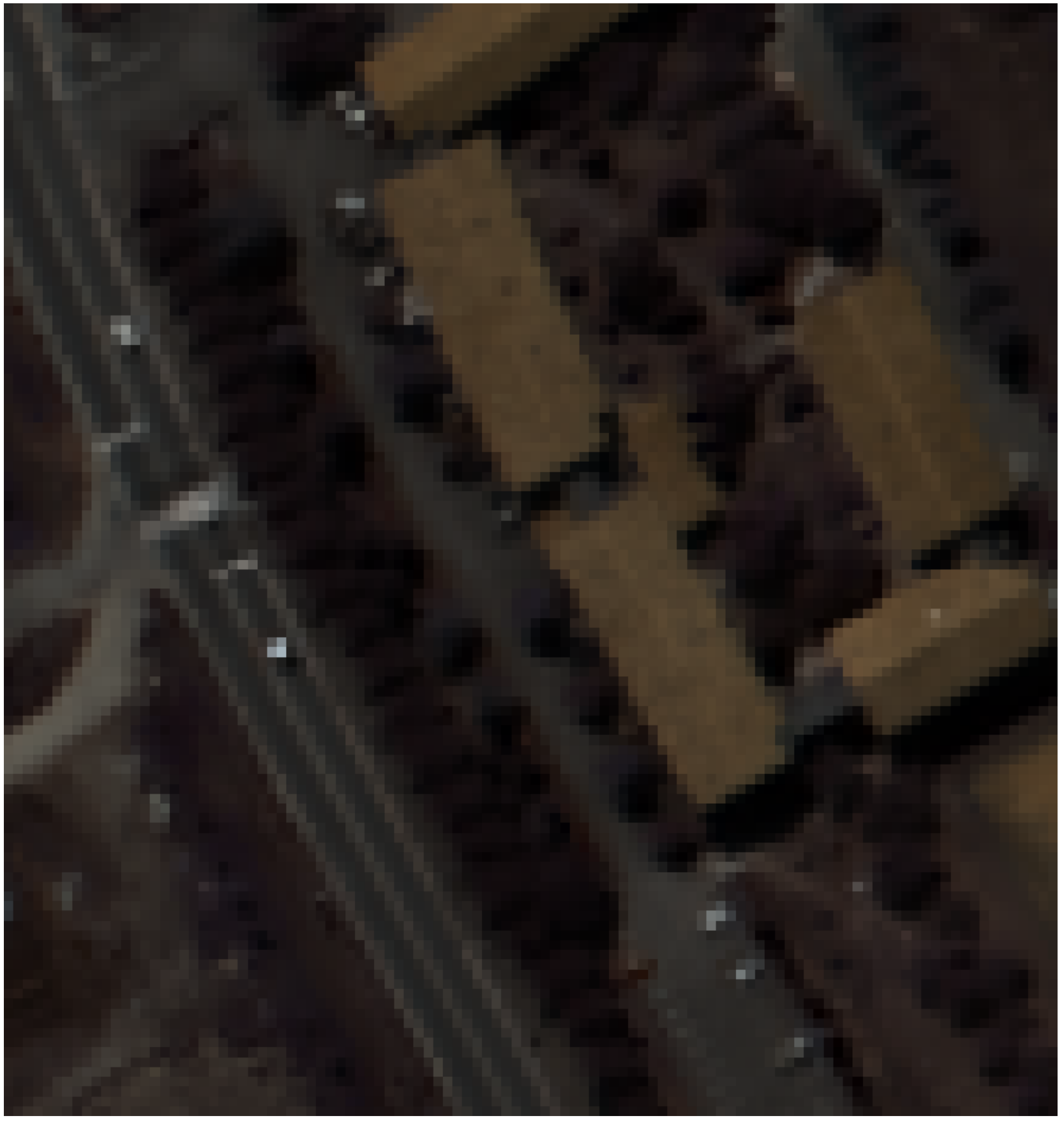}}
  \centerline{(c) }\medskip
\end{minipage} 
\begin{minipage}{0.19 \linewidth}
  \centering
  \centerline{\includegraphics[width=1.0\linewidth,height=1.0\linewidth]{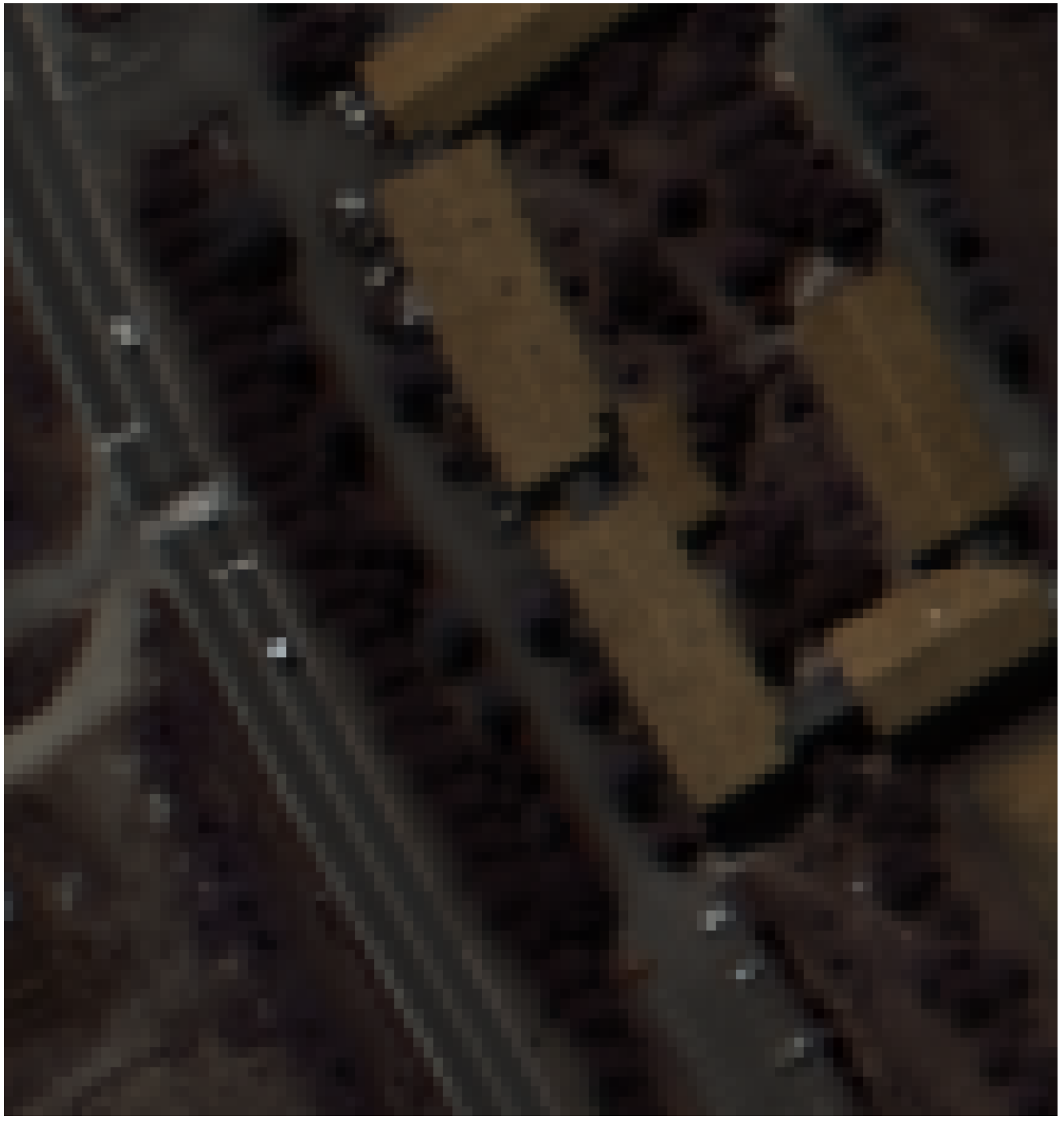}}
  \centerline{(d) }\medskip
\end{minipage}
\begin{minipage}{0.19 \linewidth}
  \centering
  \centerline{\includegraphics[width=1.0\linewidth,height=1.0\linewidth]{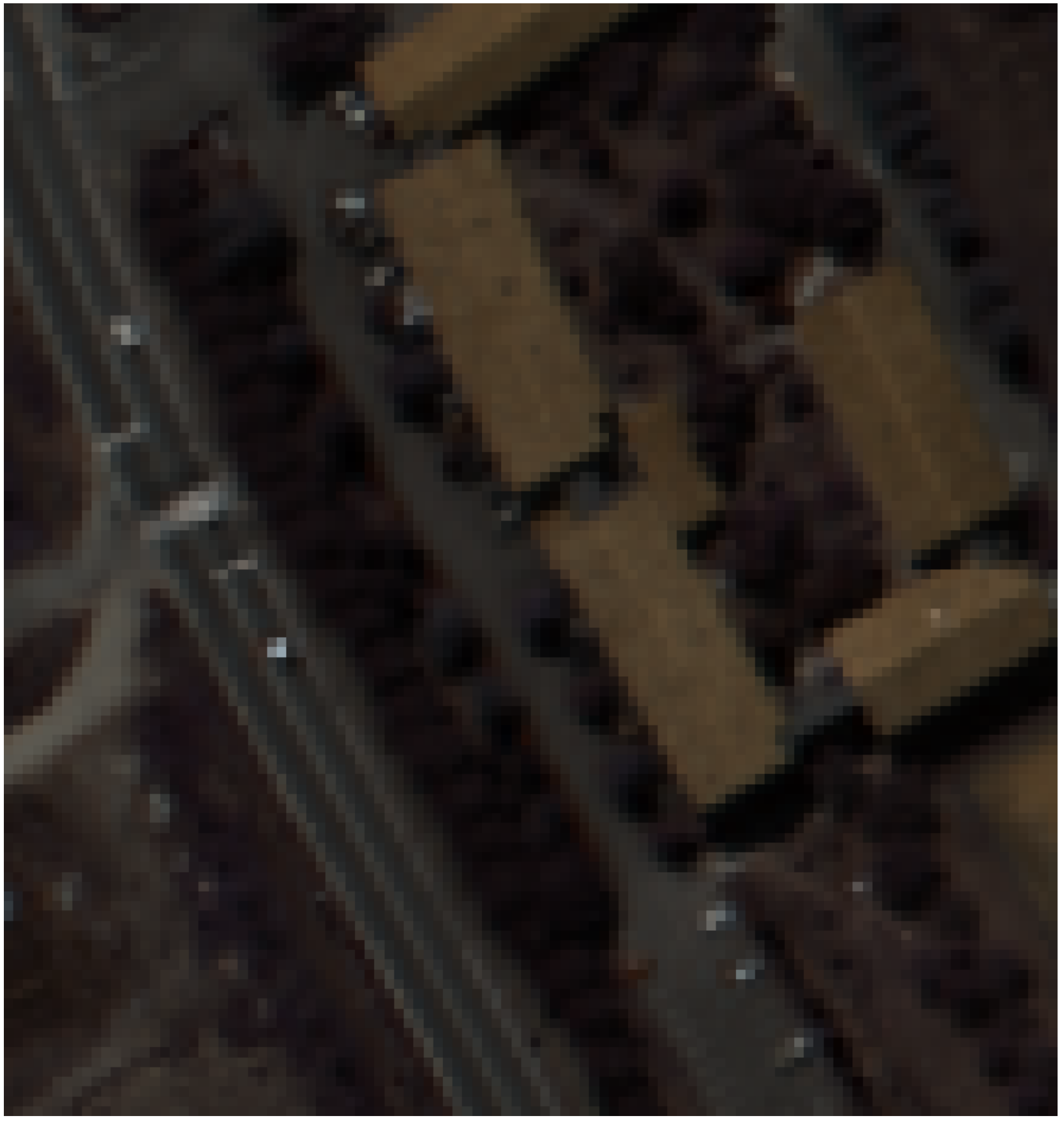}}
  \centerline{(e) }\medskip
\end{minipage}  \\
\centering
\begin{minipage}{0.19 \linewidth}
\centering
\centerline{\includegraphics[width=1.0\linewidth,height=1.0\linewidth]{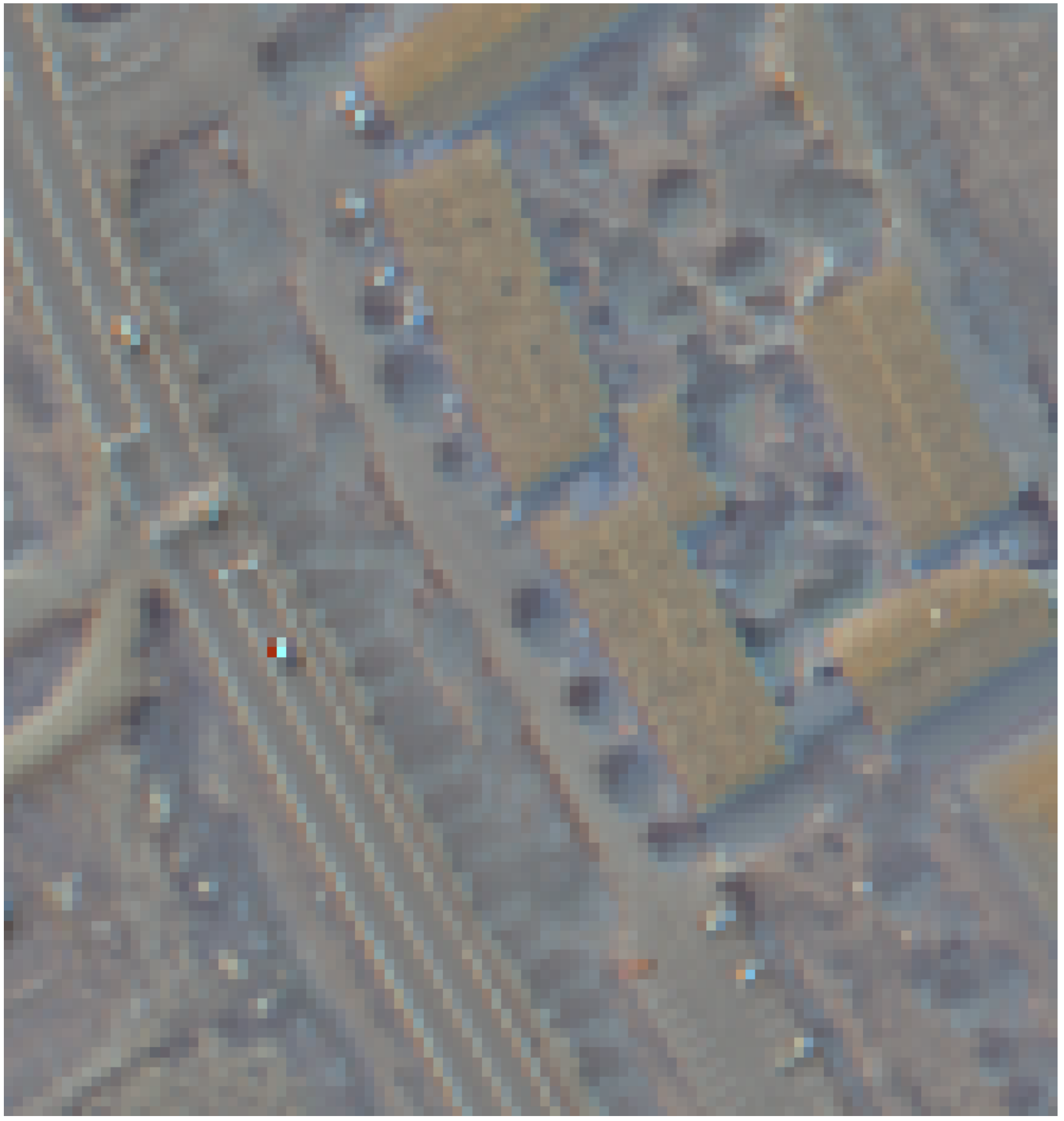}}
\centerline{(f) }\medskip
\end{minipage}
\begin{minipage}{0.19 \linewidth}  
\centering
\centerline{\includegraphics[width=1.0\linewidth,height=1.0\linewidth]{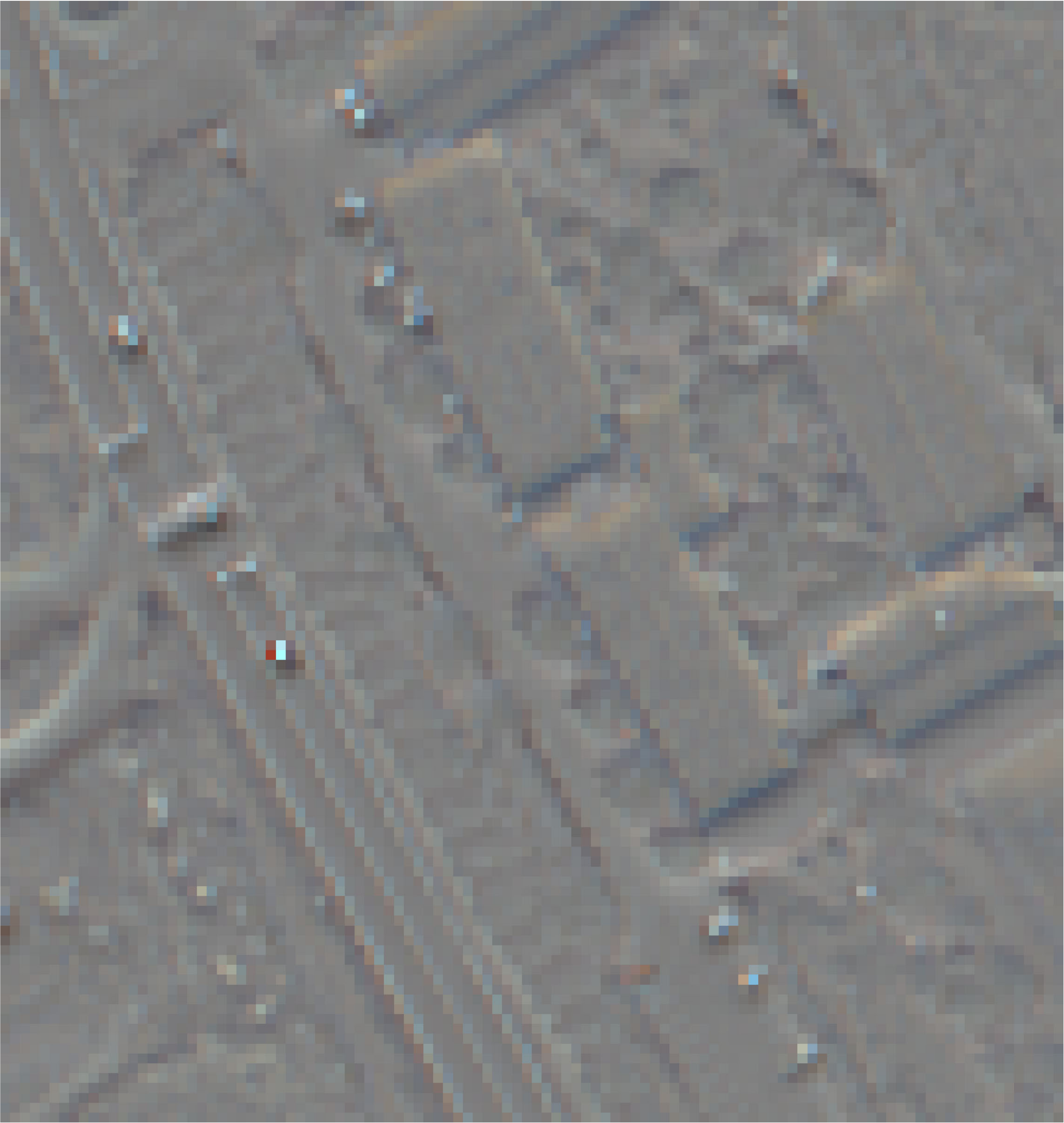}}
\centerline{(g)} \medskip
\end{minipage}  
\begin{minipage}{0.19 \linewidth}
\centering
\centerline{\includegraphics[width=1.0\linewidth,height=1.0\linewidth]{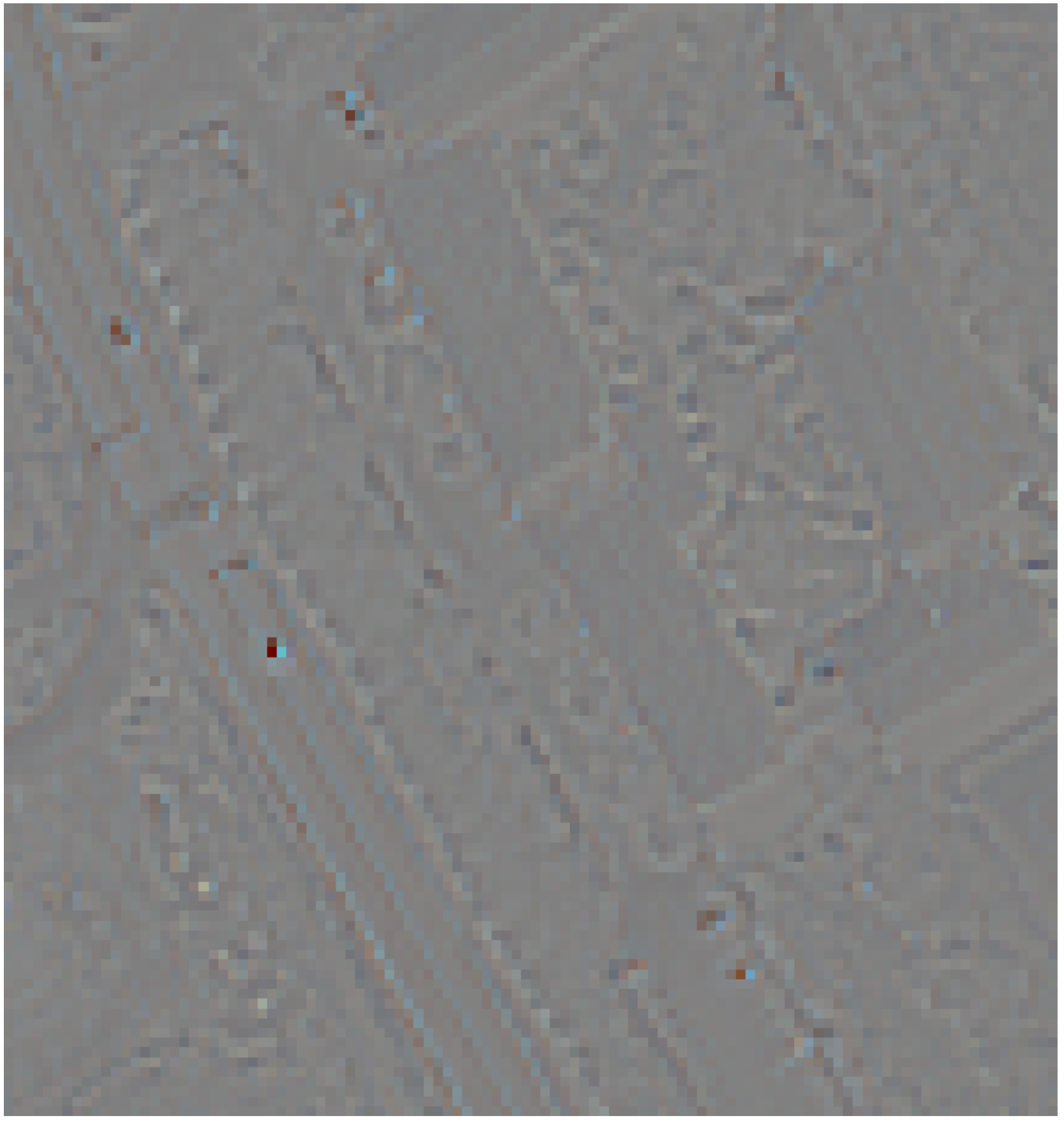}}
\centerline{(h) }\medskip
\end{minipage}
\centering
\begin{minipage}{0.19 \linewidth}
\centering
\centerline{\includegraphics[width=1.0\linewidth,height=1.0\linewidth]{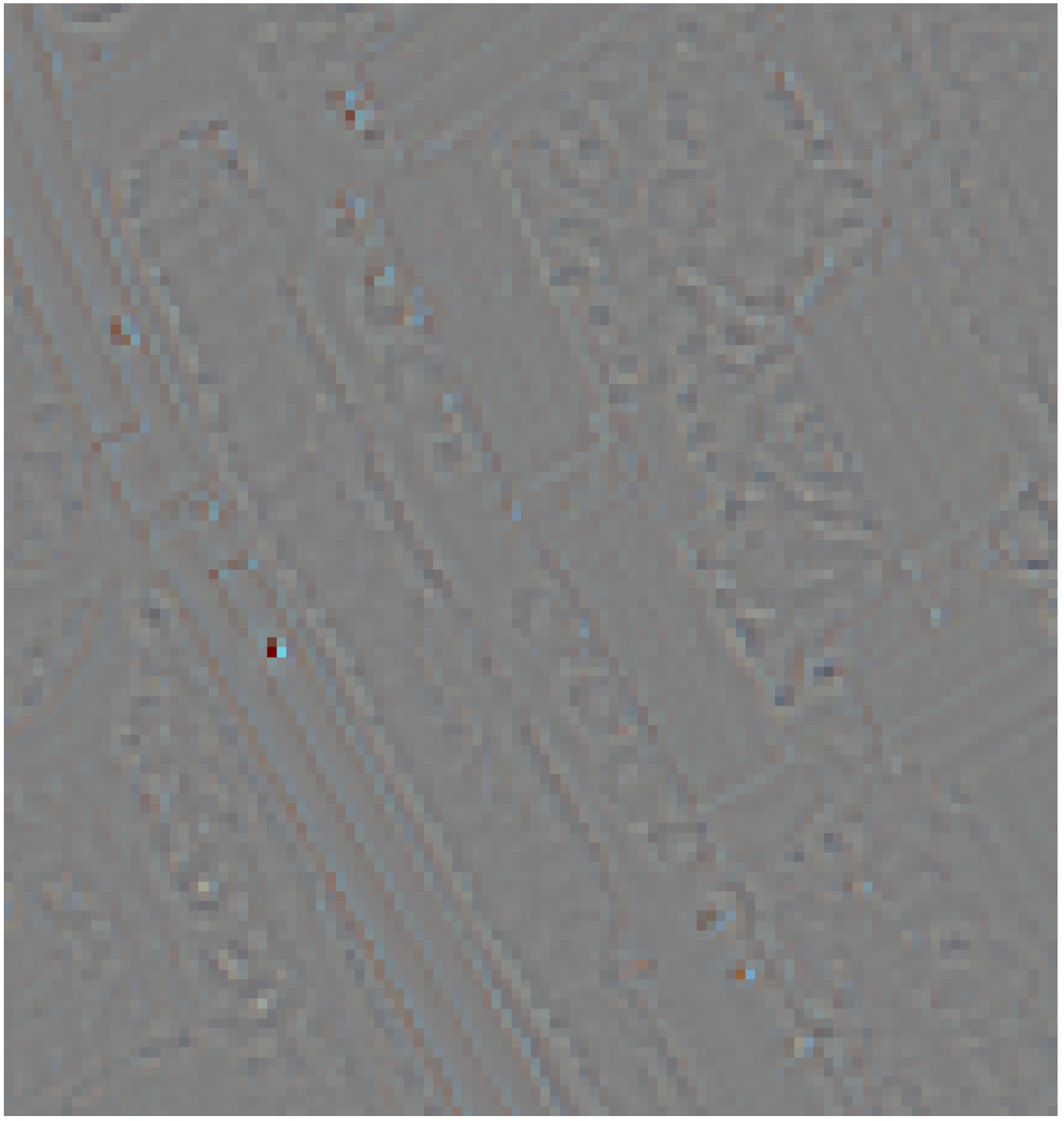}}
\centerline{(i) }\medskip
\end{minipage} 
\begin{minipage}{0.19 \linewidth}
\centering
\centerline{\includegraphics[width=1.0\linewidth,height=1.0\linewidth]{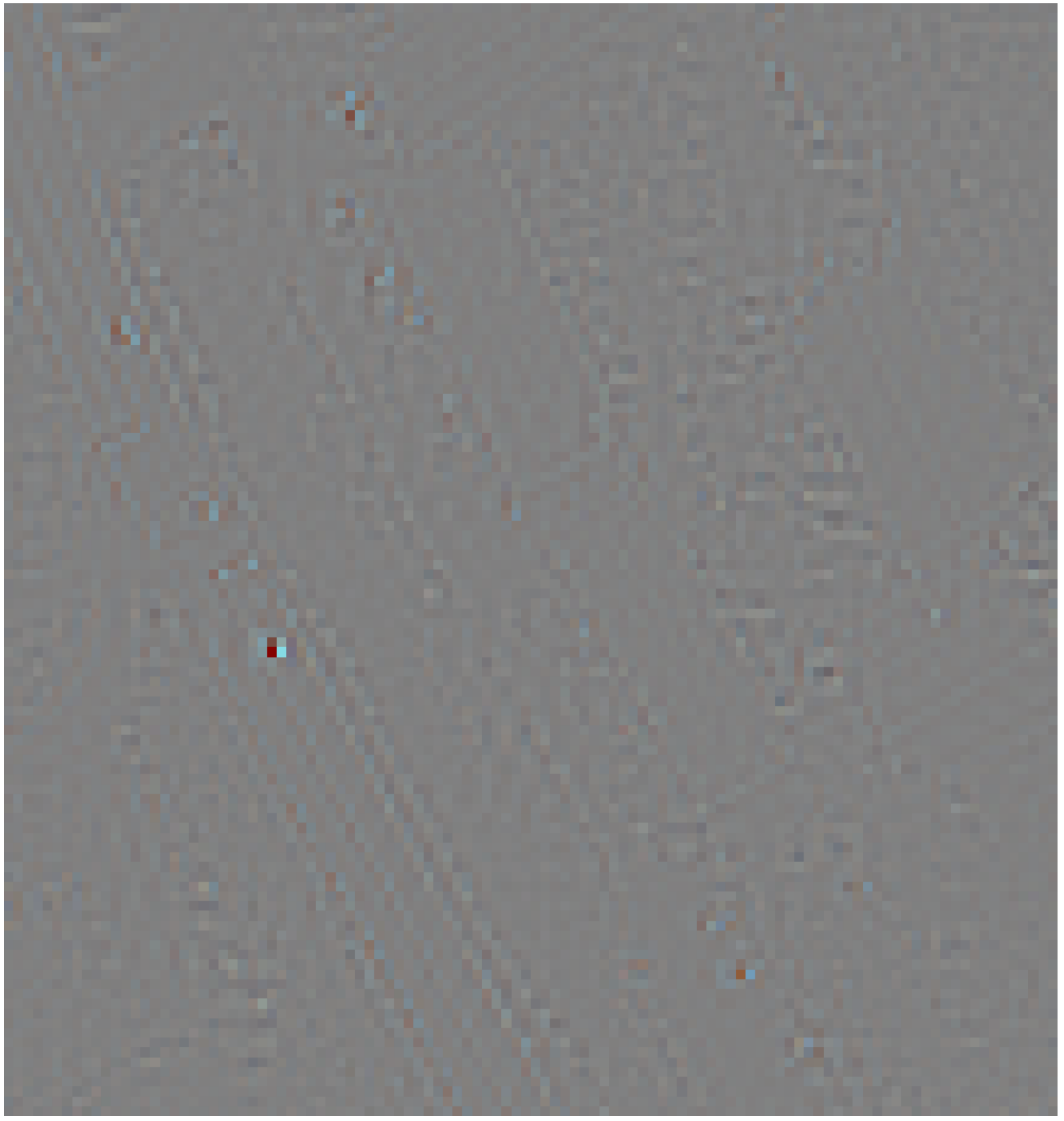}}
\centerline{(j) }\medskip
\end{minipage}
\caption{Pansharpening results from different approaches and their residuals to the ground truth (only RGB channels are shown). (a, f) Results and residuals from HySure. (b, g) Results and residuals from R-FUSE. (c, h) Results and residuals from GLR. (d, i) Results and residuals from BLT. (e, j) Results and residuals from F-BMP. For illustrative convenience, the original residuals are magnified by a factor of $2$ and added with a constant pixel intensity value ($128$) in each channel.}
\label{fig:results3_residual}
\end{figure*}

\begin{figure*}[tb]
\centering
\begin{minipage}{0.49 \linewidth}
\centerline{\includegraphics[width=9 cm]{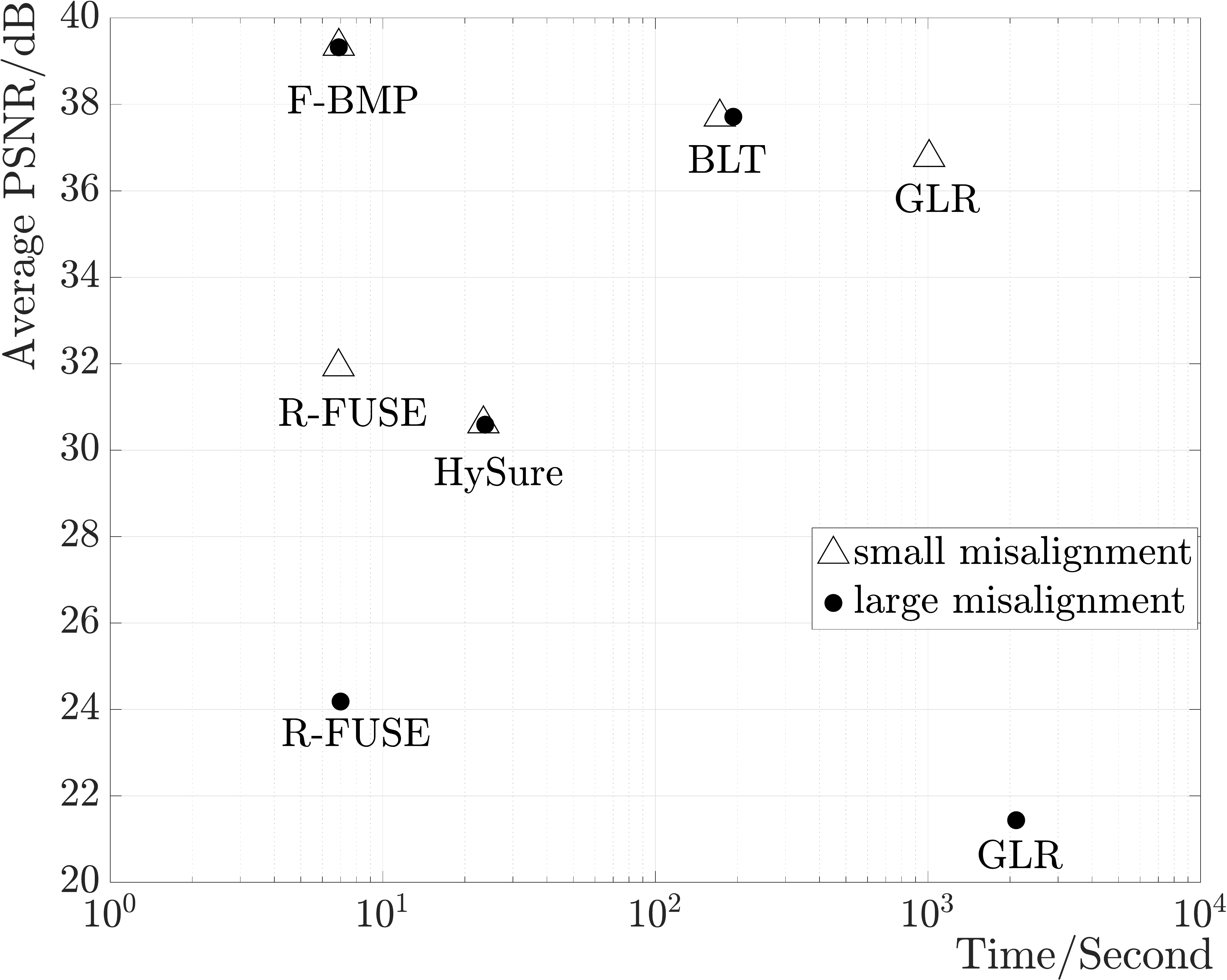}}
\centerline{(a)} \medskip
\end{minipage}
\centering
\begin{minipage}{0.49 \linewidth}
\centerline{\includegraphics[width=9 cm]{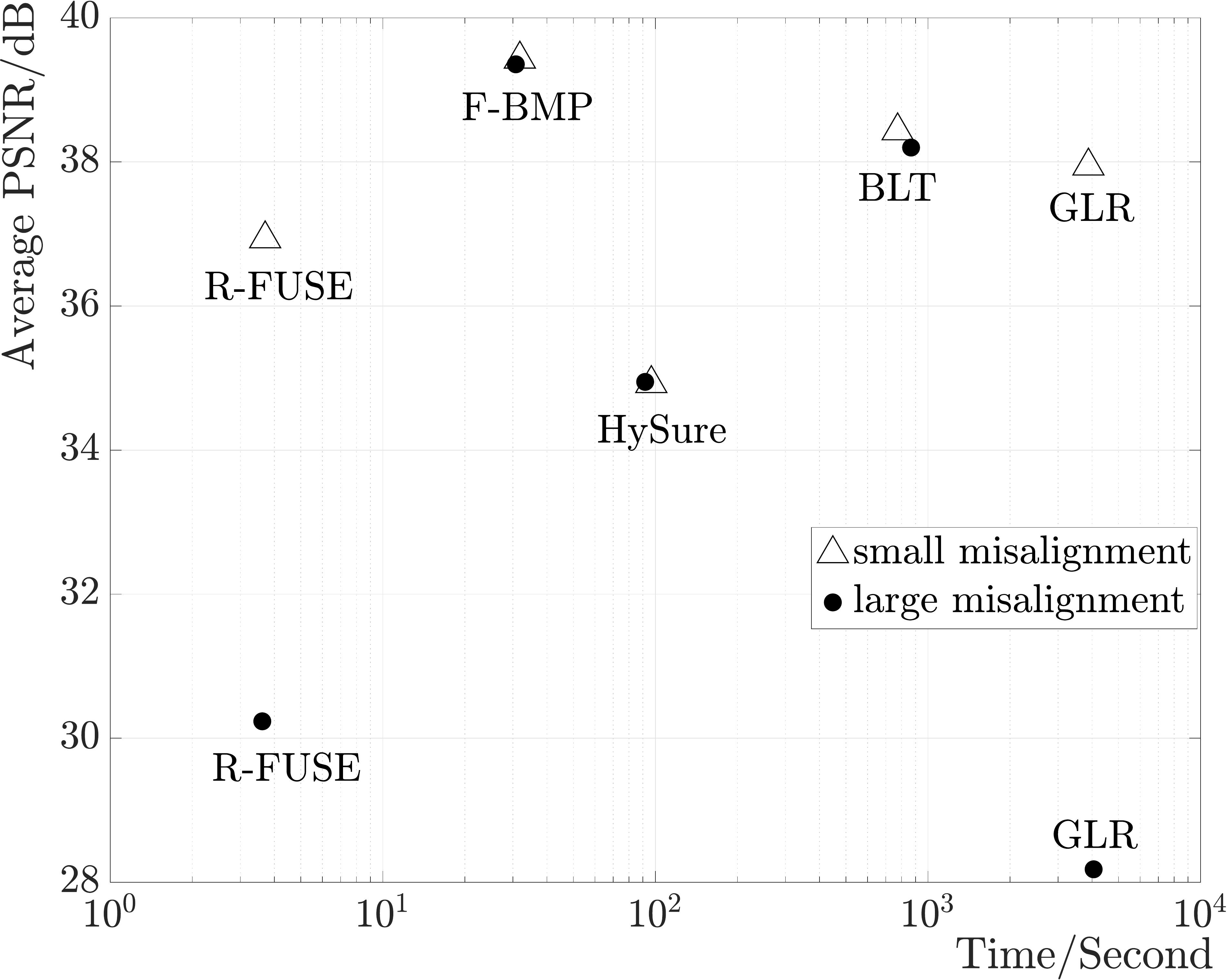}}
\centerline{(b)} \medskip
\end{minipage}
\caption{Comparisons of runtime and PSNR between F-BMP and the baseline algorithms. (a) The runtimes and PSNRs in the task of pansharpening~\textit{Pavia University} by a factor of 2, both with small and large misalignments; (b) The runtimes and PSNRs in the task of pansharpening~\textit{Cambria Fire} by a factor of 4, both with small and large misalignments.}
\label{fig:runtime_vs_psnr}
\end{figure*}

Fig.~\ref{fig:results3_residual} provides visual comparisons of the blindly pansharpened HRMS images from these algorithms when dealing with a small misalignment in a task of pansharpening by a factor of $2$. For illustrative convenience, we only show a typical spatial region with rich edges and textures in Red, Green, Blue (RGB) channels. The input PAN and LRMS images, as well as the ground-truth HRMS image, are shown in Fig.~\ref{fig:results3}(a), (b), (c), respectively. The corresponding results are presented in Fig.~\ref{fig:results3_residual}(a), (b), (c), (d), (e) for each of the aforementioned approaches, respectively. To exaggerate each result's distance to the ground-truth HRMS image, in Fig.~\ref{fig:results3_residual}(f), (g), (h), (i), (j)  we demonstrate the residual images by magnifying the actual residuals in RGB channels by a factor of $2$ and adding the magnified residuals with a constant pixel intensity value ($128$) in each channel.
 
We observe from Fig.~\ref{fig:results3_residual}(a) and Fig.~\ref{fig:results3_residual}(f) that HySure fails to preserve details around edges and textures. Further, the saturation of the pansharpened yellow rectangles in Fig.~\ref{fig:results3_residual}(a) is noticeably smaller than that of the ground-truth HRMS image in Fig.~\ref{fig:results3}(c). From Fig.~\ref{fig:results3_residual}(b) and Fig.~\ref{fig:results3_residual}(g), we find that R-FUSE also has noticeable color distortion. Also, it generates blurry edges of the yellow rectangles and three blurry parallel white lines along the diagonal direction. Fig.~\ref{fig:results3_residual}(c, h) and Fig.~\ref{fig:results3_residual}(d, i) show that GLR and BLT can both roughly preserve the sharpness of the structures, but with noticeable residuals in the vicinity of edges. Fig.~\ref{fig:results3_residual}(e) and Fig.~\ref{fig:results3_residual}(j) show that F-BMP manages to generate the HRMS image with minor residuals to the ground truth. 

The quantitative performance comparisons of the five algorithms are shown in Table~\ref{tab:result_2x} and Table~\ref{tab:result_4x}. The results of F-BMP, BLT and HySure have nearly consistent quantitative image qualities across small and large misalignments in terms of $\overline{\operatorname{PSNR}}$, $\mathrm{EGRAS}$, $\mathrm{SAM}$ and $\mathrm{RASE}$. For GLR and R-FUSE, there exists significant performance drops when dealing with a large misalignment. These drops in GLR are attributed to that the convergence of estimating the blur kernel getting trapped into bad local minima. These drops in R-FUSE is due to that R-FUSE does not factor into spatial misalignment. When the misalignment is small, in $\overline{\operatorname{PSNR}}$ sense, F-BMP outperforms BLT by $2.36$ dB and the approaches other than BLT by at least $3.69$ dB in the task of pansharpening by a factor of $2$. Also, it outperforms BLT by $1.11$ dB and and the approaches other than BLT by at least $2.76$ dB in the task of pansharpening by a factor of $4$. When the misalignment is large, in $\overline{\operatorname{PSNR}}$ sense, F-BMP outperforms BLT by $2.25$ dB and HySure by at least $9.96$ dB in the task of pansharpening by a factor of $2$. Also, it outperforms BLT by $1.15$ dB and HySure by $5.53$ dB in the task of pansharpening by a factor of $4$. The superiority of F-BMP over the related algorithms is also reflected in terms of $\mathrm{EGRAS}$, $\mathrm{SAM}$ and $\mathrm{RASE}$, shown in Table~\ref{tab:result_2x} and Table~\ref{tab:result_4x}.

F-BMP not only demonstrates state-of-the-art pansharpening quality, but also costs short runtime among the aforementioned model-based approaches. To demonstrate both the runtime and $\overline{\operatorname{PSNR}}$, we choose two typical datasets,~\textit{Pavia University} and~\textit{Cambria Fire}, in pansharpening by a factor of $2$ and $4$ tasks, respectively, both with small and large misalignments. The runtimes are measured on a 2.6G Hz 18-Core Intel i9 processor using MATLAB (R2020b) and are demonstrated in Fig.~\ref{fig:runtime_vs_psnr}. F-BMP takes the second shortest runtimes and generates the highest $\overline{\operatorname{PSNR}}$ among all the blind pansharpening approaches. F-BMP's runtimes on both datasets are not the shortest, since R-FUSE is the fastest algorithm solving the HRMS image via a closed-form expression accelerated by FFT. However, R-FUSE cannot handle large misalignment: both Fig.~\ref{fig:runtime_vs_psnr}(a) and Fig.~\ref{fig:runtime_vs_psnr}(b) show significant $\overline{\operatorname{PSNR}}$ drops when the spatial misalignments are large. In addition, the algorithms, GLR and BLT, that iteratively compute the blur kernel and the HRMS image in an alternating fashion are slow. In conclusion, there does not exist a model-based blind pansharpening algorithm addressing spatial misalignments with both high efficiency and high quality in the literature. Among all the model-based blind pansharpening algorithms that address the spatial misalignment issue, F-BMP demonstrates the highest $\overline{\operatorname{PSNR}}$ and the shortest runtime.

\subsection{Comparison with a Deep Learning-based Approach}

\begin{figure}[!t]
\centering 
    \begin{minipage}{.5\linewidth}
    \centering
    \includegraphics[width=.98\linewidth]{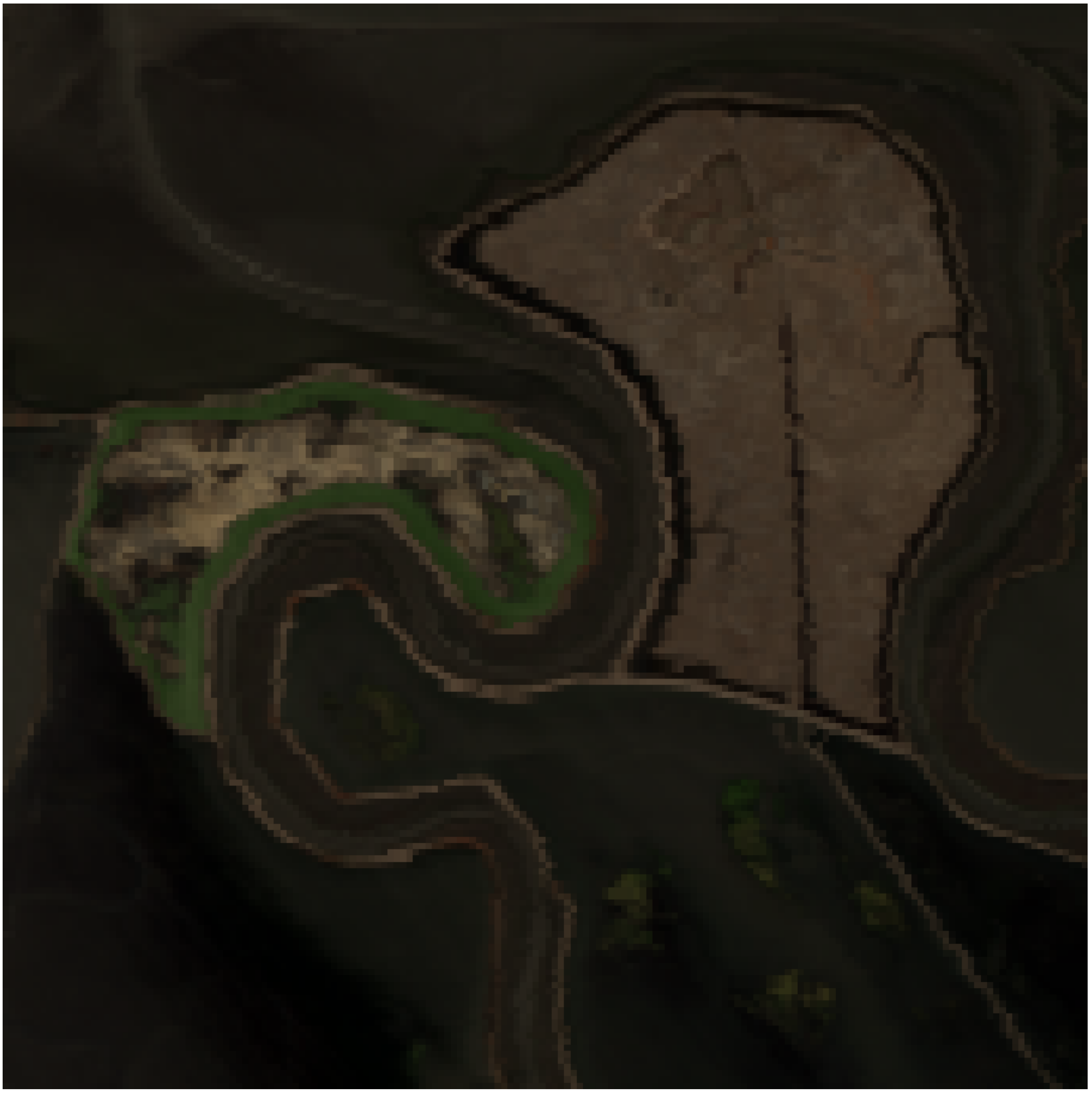}
    \centerline{(a)}      
    \end{minipage}\hfill
    \begin{minipage}{.5\linewidth}
    \centering
    \includegraphics[width=.98\linewidth]{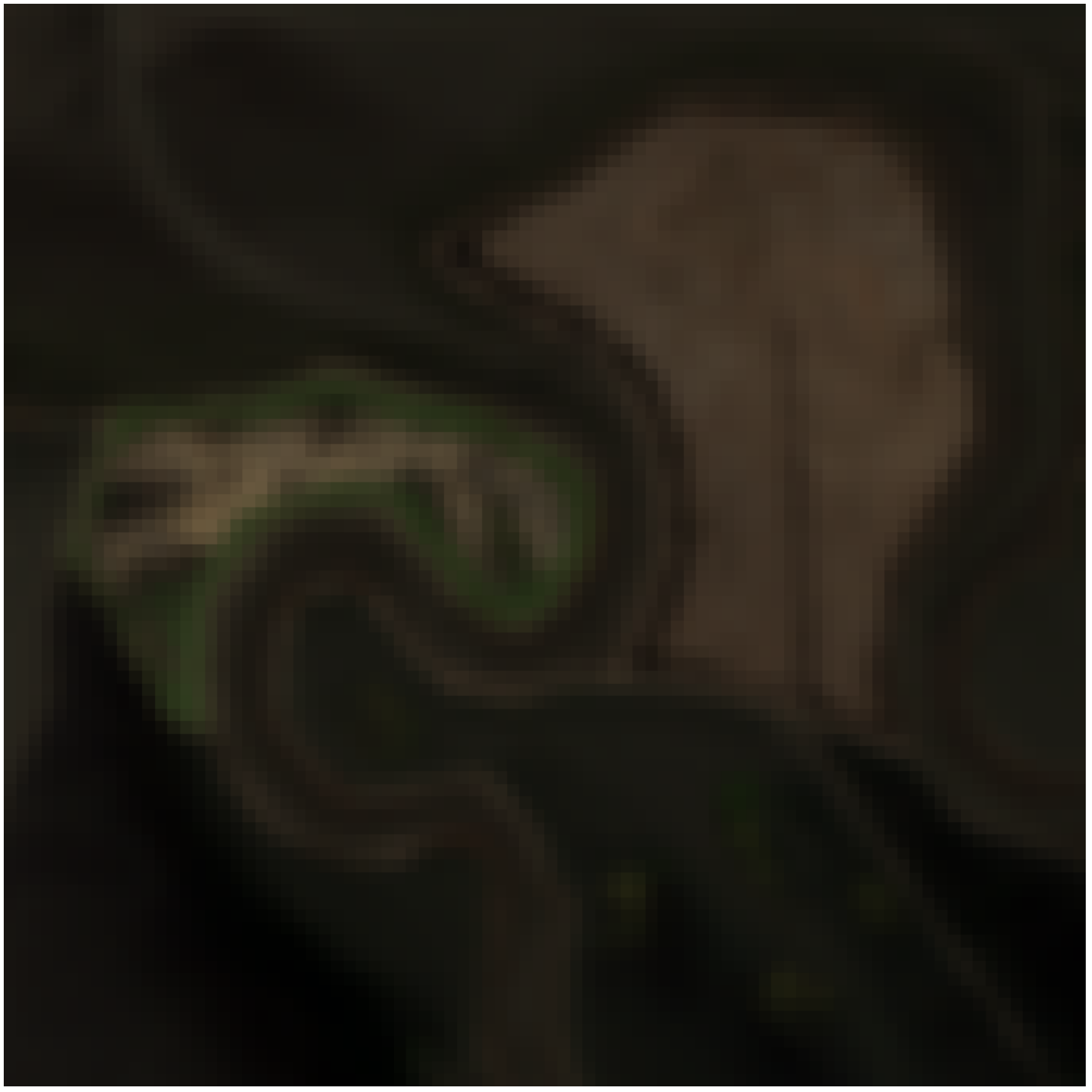}
    \centerline{(b)}        
    \end{minipage}\hfill
    \begin{minipage}{.5\linewidth}
    \centering
    \includegraphics[width=.98\linewidth]{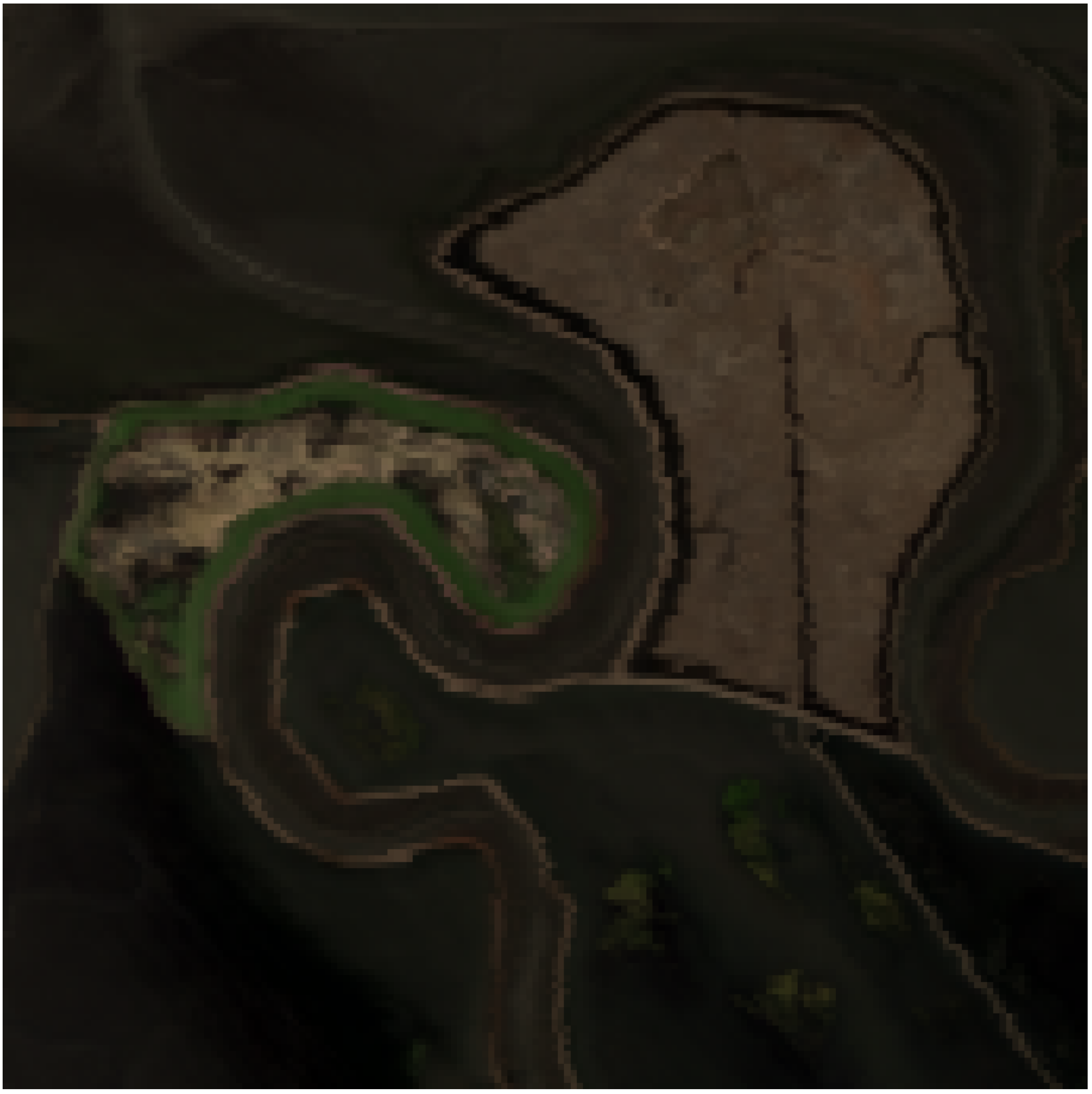}
    \centerline{(c)}        
    \end{minipage}\hfill
    \begin{minipage}{.5\linewidth}
    \centering
    \includegraphics[width=.98\linewidth]{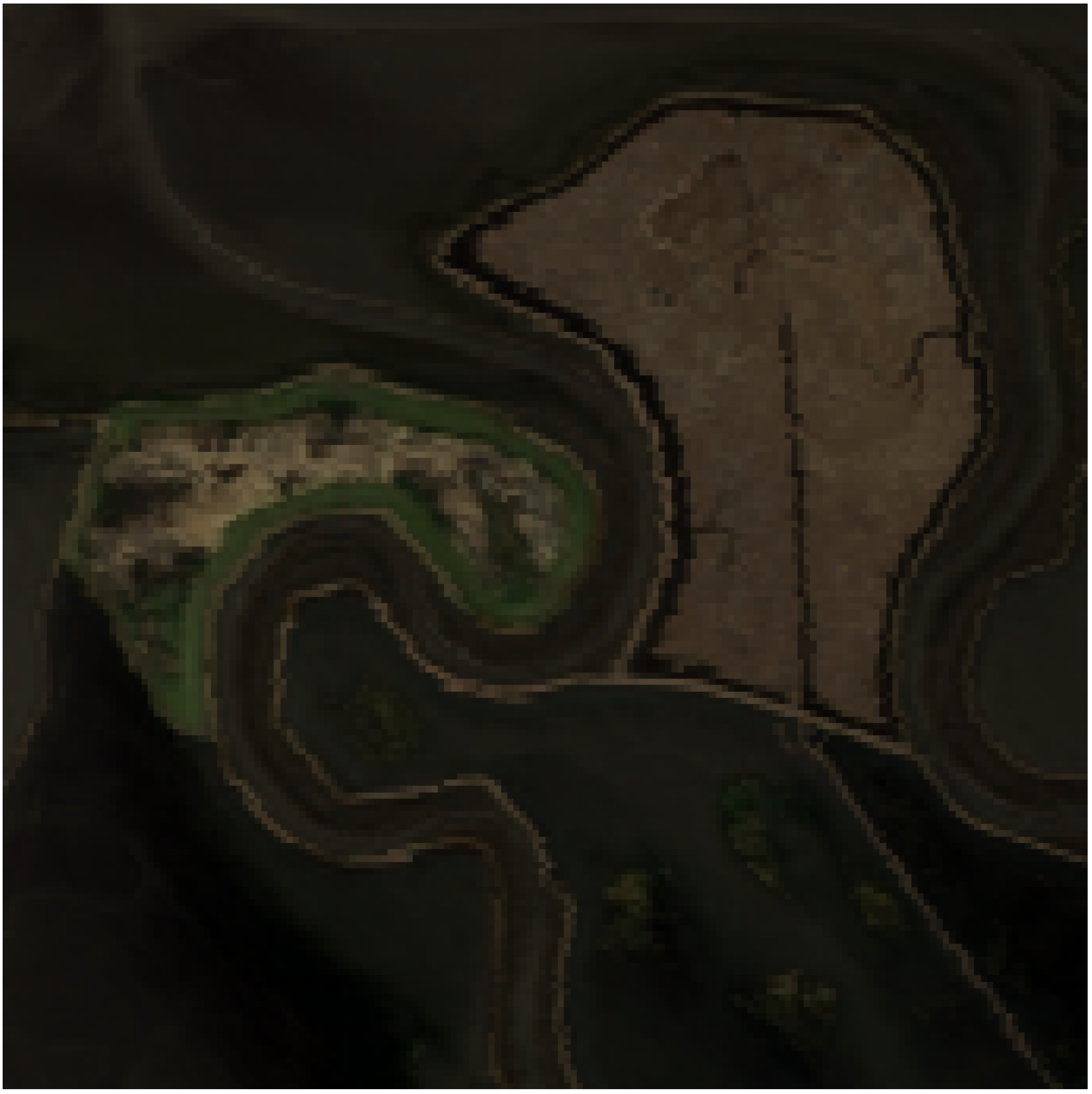}
    \centerline{(d)}        
    \end{minipage}\hfill    
    \begin{minipage}{.5\linewidth}
    \centering
    \includegraphics[width=.98\linewidth]{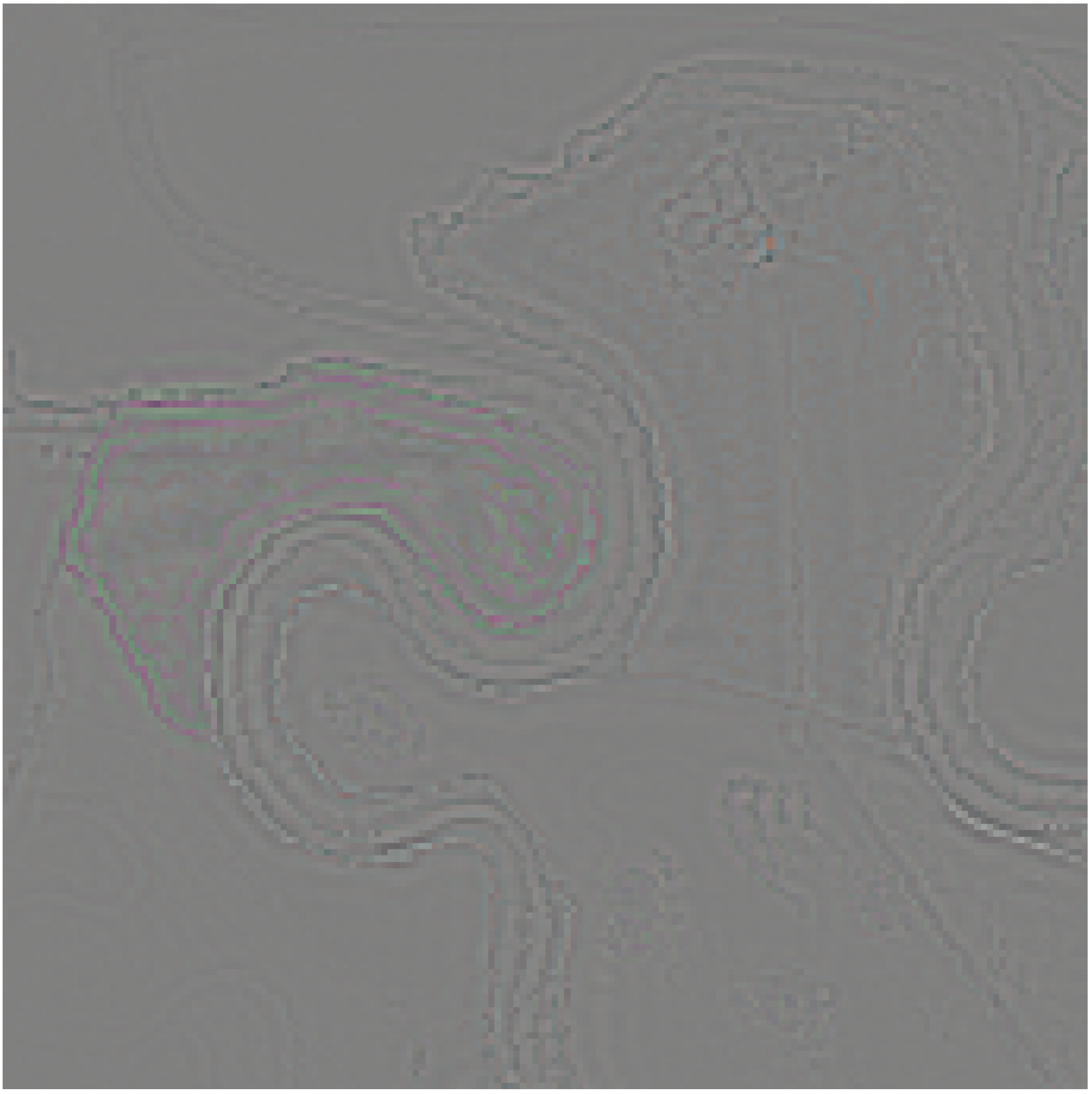}
    \centerline{(e)}                
    \end{minipage}\hfill
    \begin{minipage}{.5\linewidth}
    \centering
    \includegraphics[width=0.98\linewidth]{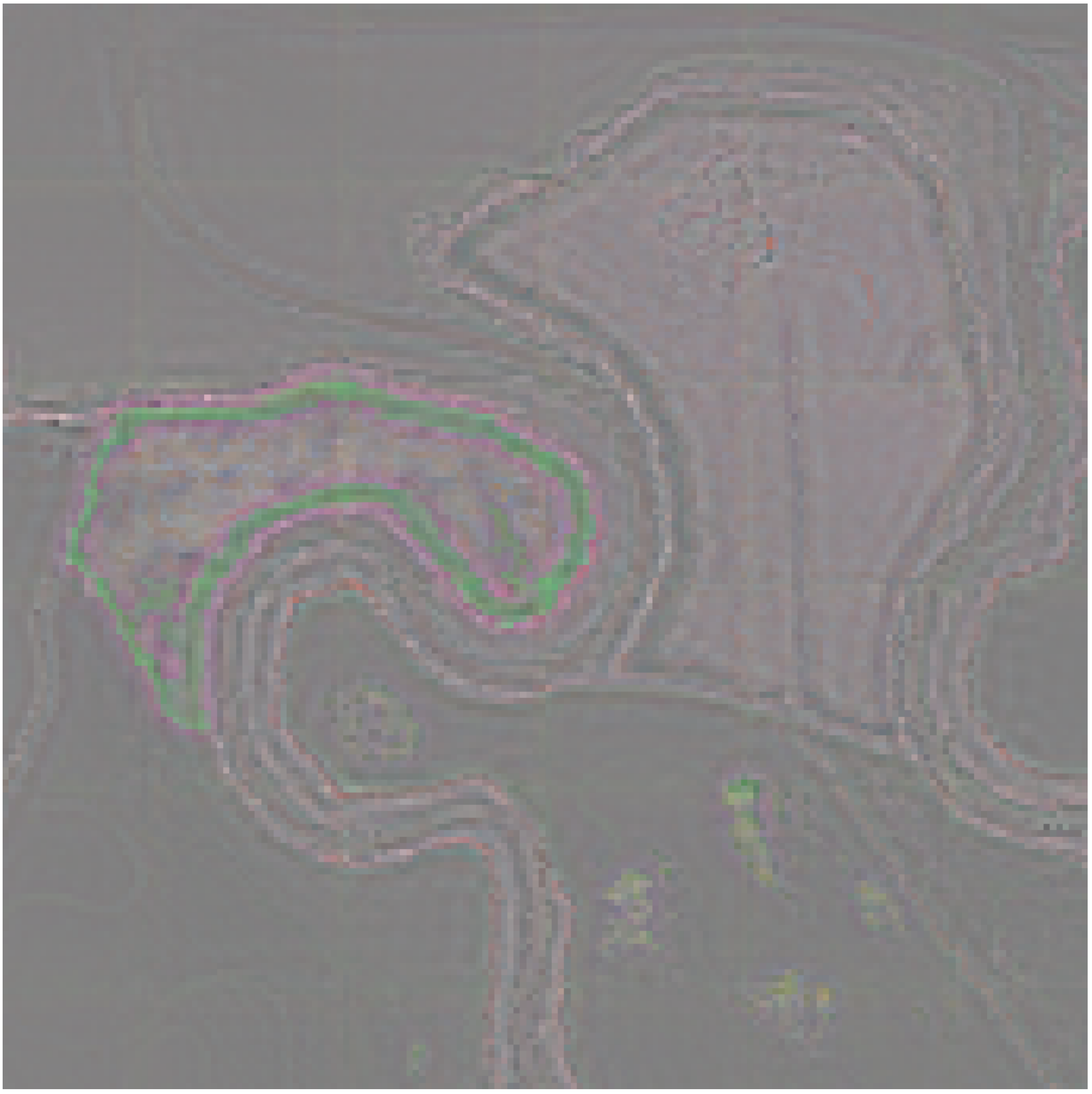}
    \centerline{(f)}                
    \end{minipage}
    \caption{Blind pansharpening results comparison between F-BMP and PGDCNN (data source:~\textit{Moffett}). (a) The RGB channels of the ground-truth HRMS image. (b) The RGB channels of the input LRMS image. Each pixel is purposely enlarged by a factor of $2$, both horizontally and vertically, to fit the space. (c) The RGB channels of the HRMS image from F-BMP. (d) The RGB channels of the HRMS image from PGDCNN. (e) The residual RGB channels between the HRMS image from F-BMP and the ground-truth HRMS image. (f) The residual RGB channels between the HRMS image from PGDCNN and F-BMP. The original residuals are magnified by a factor of $4$ and added by a constant intensity($128$) to generate (e) and (f).}
\label{fig:results_vs_dl}
\end{figure}

We also compare F-BMP with a deep learning-based algorithm in the task of pansharpening by a factor of $2$ with small spatial misalignment. We choose the most recent work, PGDCNN~\cite{lohit2019unrolled}, as the benchmark and use PGDCNN's dataset for comparison. To make a fair comparison, we use the same Gaussian blur kernel to simulate the LRMS images. Table~\ref{tab:results5} demonstrates that compared with PGDCNN, F-BMP yields much higher performance, with a $4.48$ dB $\overline{\operatorname{PSNR}}$ gain and $1.66$ dB $\overline{\operatorname{PSNR}}_{reg}$ gain over PGDCNN in four test images:~\textit{Moffett},~\textit{Cuprite},~\textit{Los Angeles}~(\textit{L.A.}), and \textit{Cambria Fire}~(\textit{C.F.}).

\begin{table}[!t]
\caption{Average PSNR (in decibels, upper two rows) and Regressed PSNR (in decibels, lower two rows) Comparisons of Blind Pansharpening Results Using F-BMP and PGDCNN.}
\centering
	\begin{tabular}{||c|c|c|c|c|c||}
		\hline
		Test Images & Moffett & Cuprite & L.A. & C.F. & Mean \\
		\hline
		 PGDCNN & $38.16$ &  $38.92$ & $37.86$ & $33.70$ & $37.16$ \\
		\hline		 
		 F-BMP & $\mathbf{41.38}$ &  $\mathbf{43.26}$  & $\mathbf{40.54}$  &  $\mathbf{41.36}$ & $\mathbf{41.64}$ \\	
		\hline
		 PGDCNN & $40.13$ & $42.20$ & $38.30$ & $39.57$ & $40.05$ \\
		\hline		 
		 F-BMP & $\mathbf{41.41}$ &  $\mathbf{43.32}$  & $\mathbf{40.55}$  &  $\mathbf{41.57}$ & $\mathbf{41.71}$ \\	
		\hline		 
	\end{tabular}
	\label{tab:results5}	
\end{table}	
	
Fig.~\ref{fig:results_vs_dl} compares the visual qualities of both algorithms in recovering the RGB channels of~\textit{Moffett} and demonstrates the exaggerated residual images. As evident from these residuals, F-BMP outperforms PGDCNN in many smooth areas. The main reason for this discrepancy seems to be that it is more difficult for a learning algorithm to be trained in a large variety of conditions, including blur, misalignment, etc., requiring data to be provisioned for a wide variety of cases to provide robustness. In contrast, F-BMP adaptively determines a kernel for each pair of PAN and LRMS images, providing more robustness, especially to the level of blurring under different blur conditions.

\subsection{Experiments when the Ground Truth is not Available}

\begin{figure*}[!thb]
\centering
\begin{minipage}{0.19 \linewidth}
  \centering
  \centerline{\includegraphics[width=1.0\linewidth]{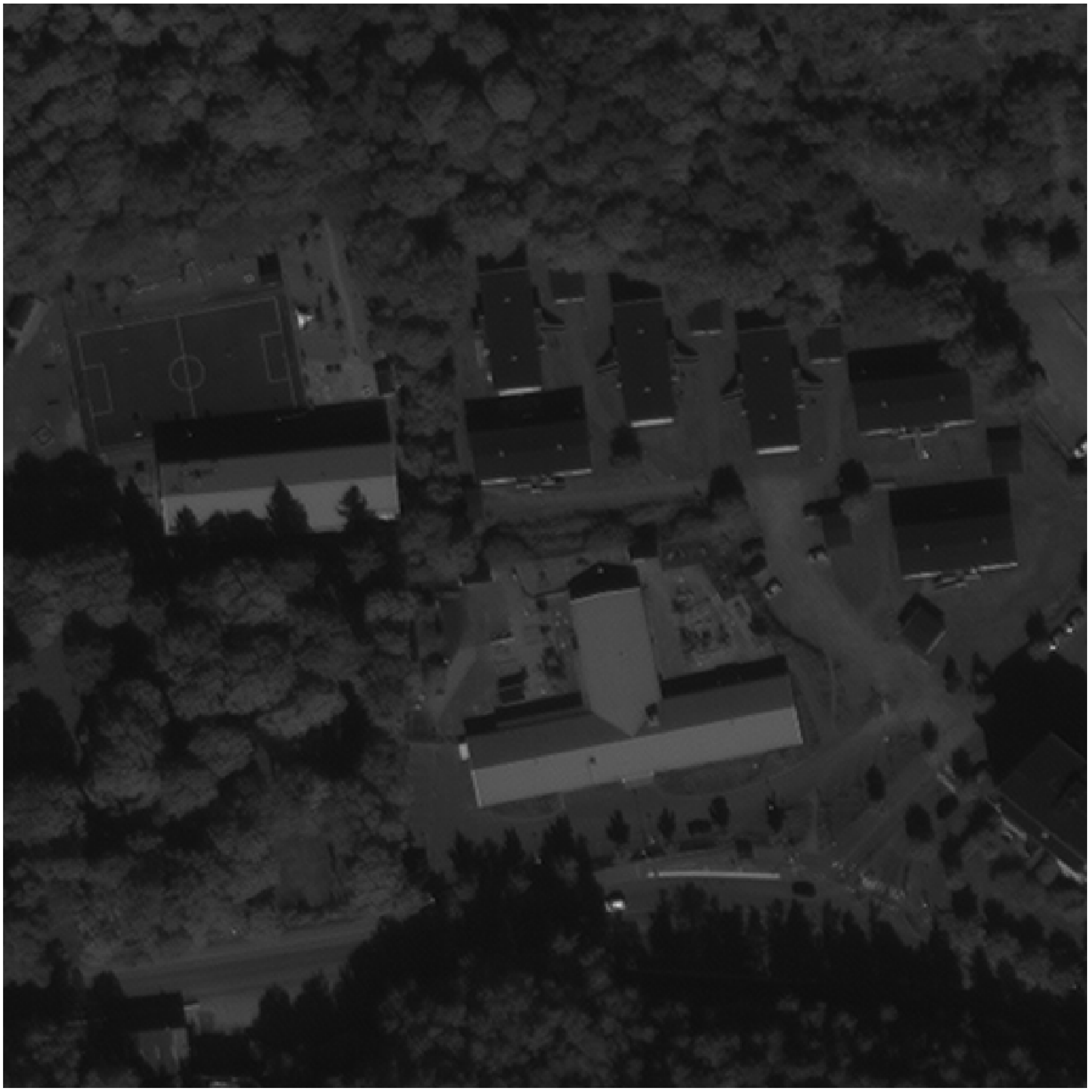}}
  \centerline{(a) PAN}\medskip
\end{minipage}
\begin{minipage}{0.19 \linewidth}
  \centering
  \centerline{\includegraphics[width=1.01\linewidth]{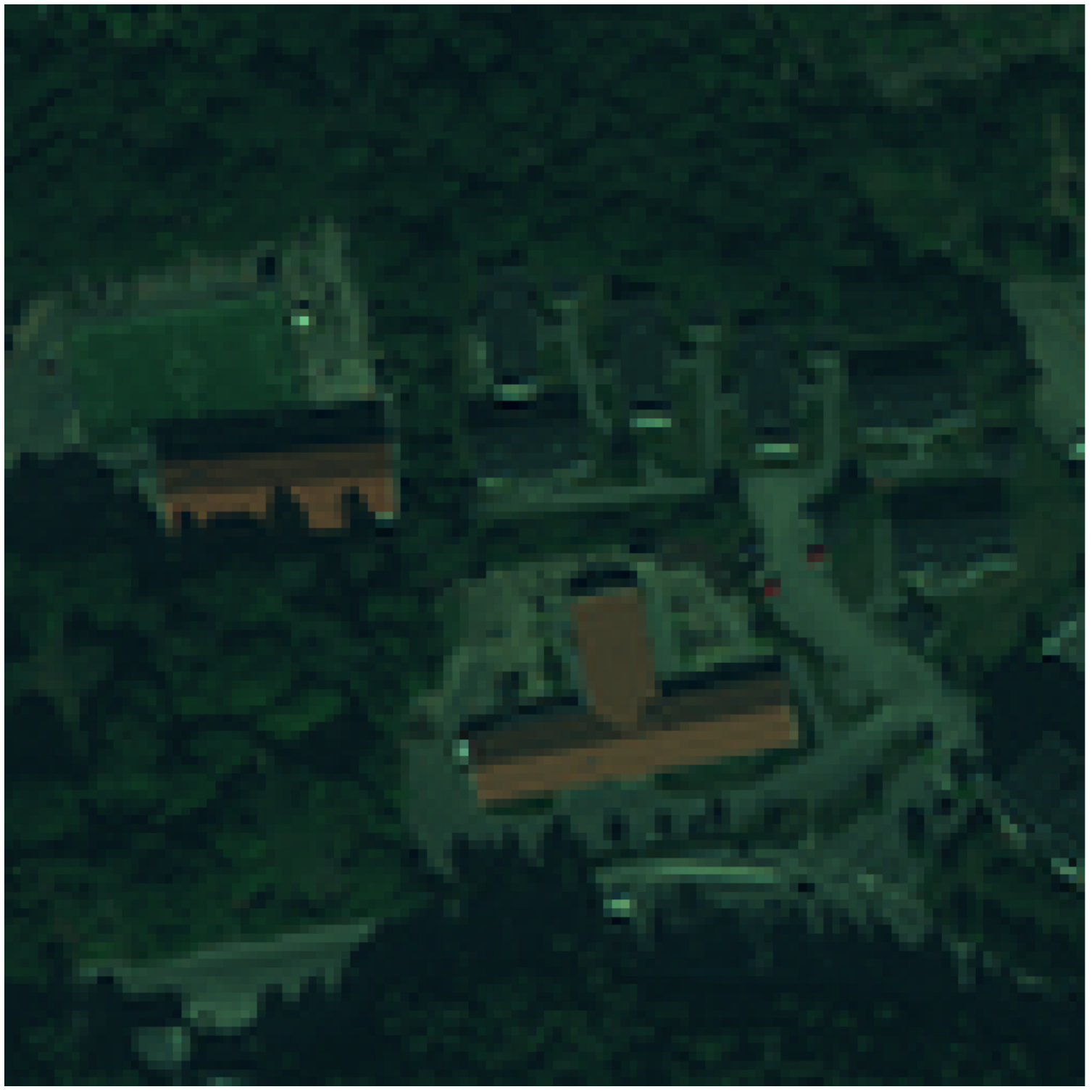}}
  \centerline{(b) LRMS}\medskip
\end{minipage} \\
\begin{minipage}{0.19 \linewidth}
  \centering
  \centerline{\includegraphics[width=1.0\linewidth]{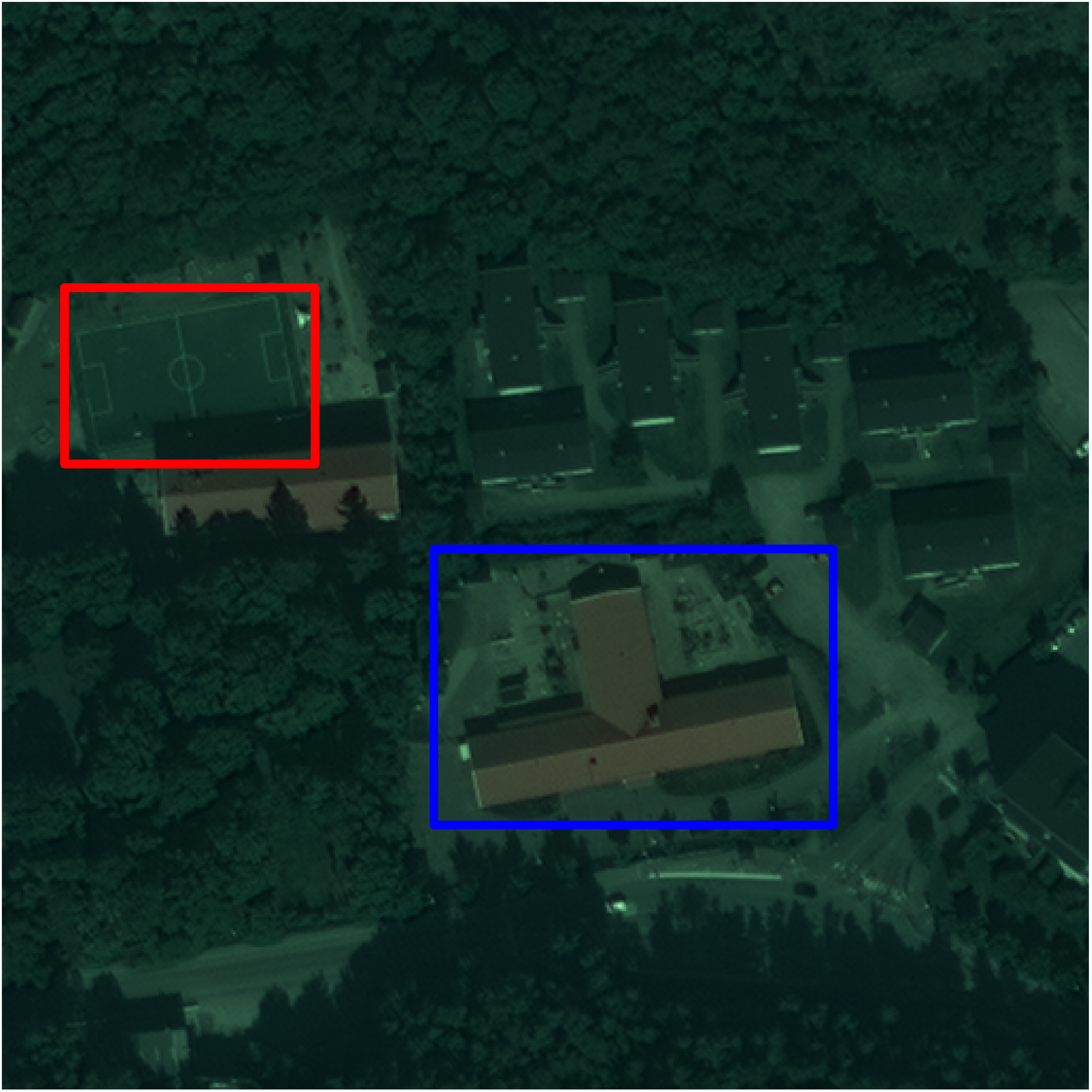}}
  \centerline{(c) HySure }\medskip
\end{minipage} 
\begin{minipage}{0.19 \linewidth}
  \centering
  \centerline{\includegraphics[width=1.0\linewidth]{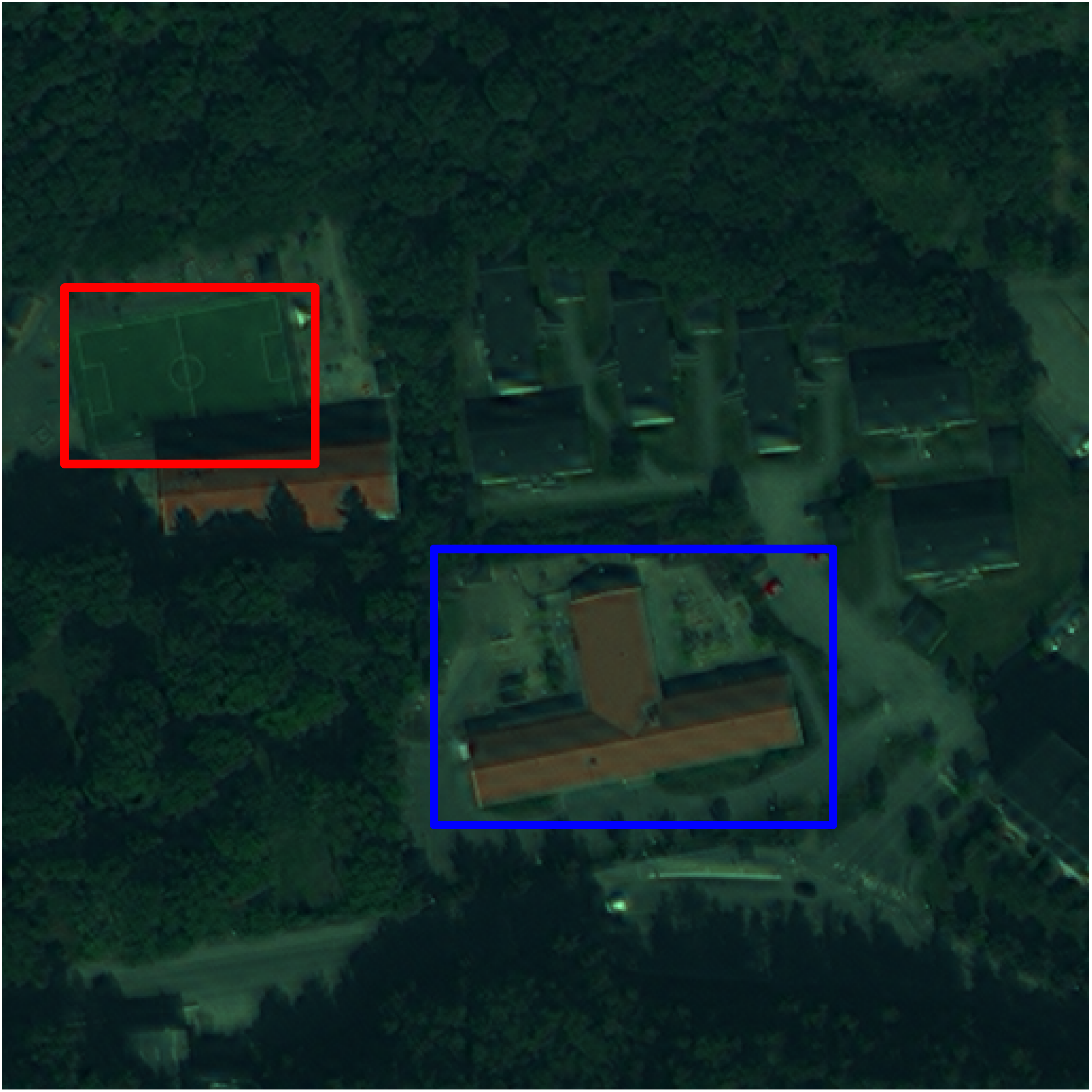}}
  \centerline{(d) R-FUSE}\medskip
\end{minipage} 
\begin{minipage}{0.19 \linewidth}
  \centering
  \centerline{\includegraphics[width=1.0\linewidth]{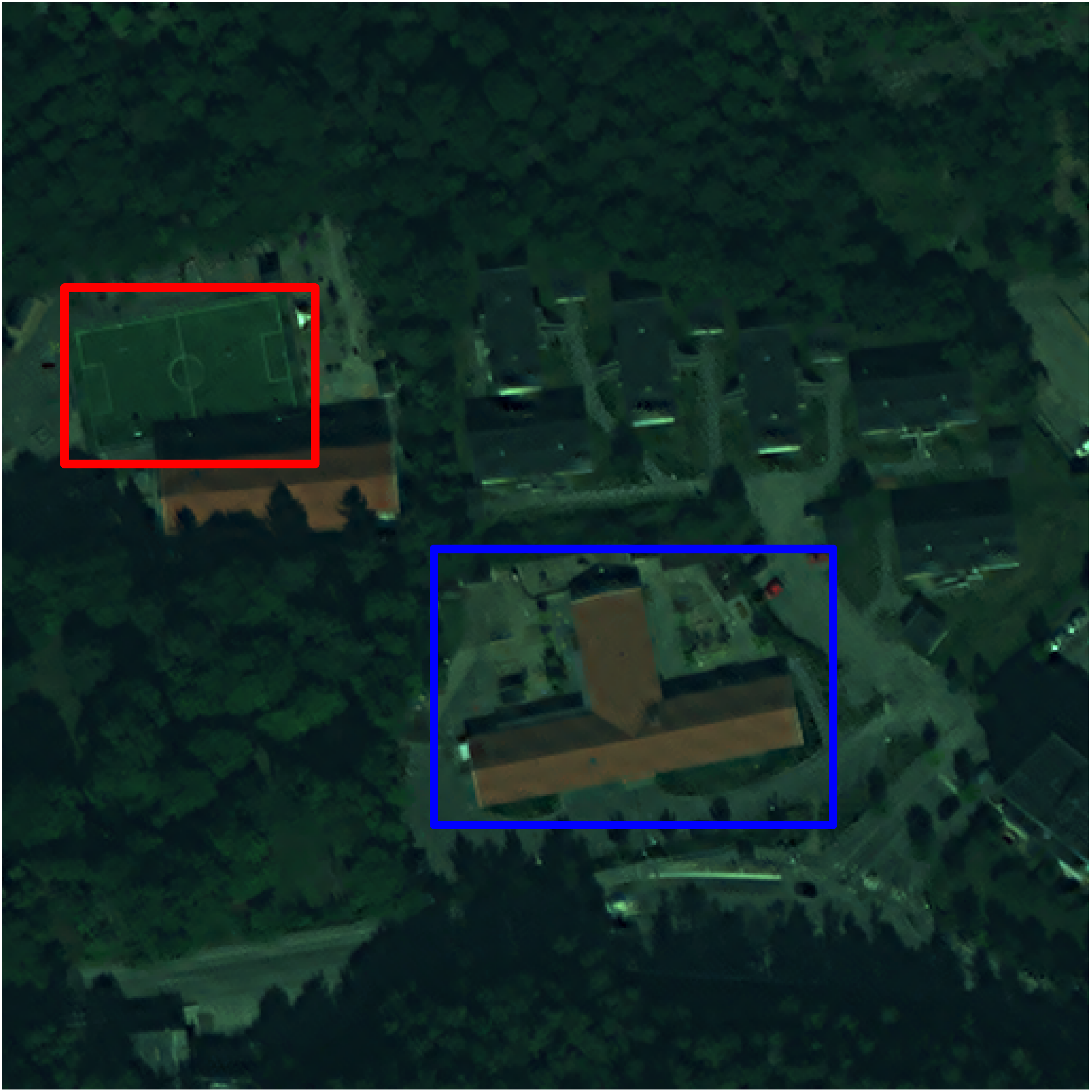}}
  \centerline{(e) GLR}\medskip
\end{minipage} 
\begin{minipage}{0.19 \linewidth}
  \centering
  \centerline{\includegraphics[width=1.0\linewidth]{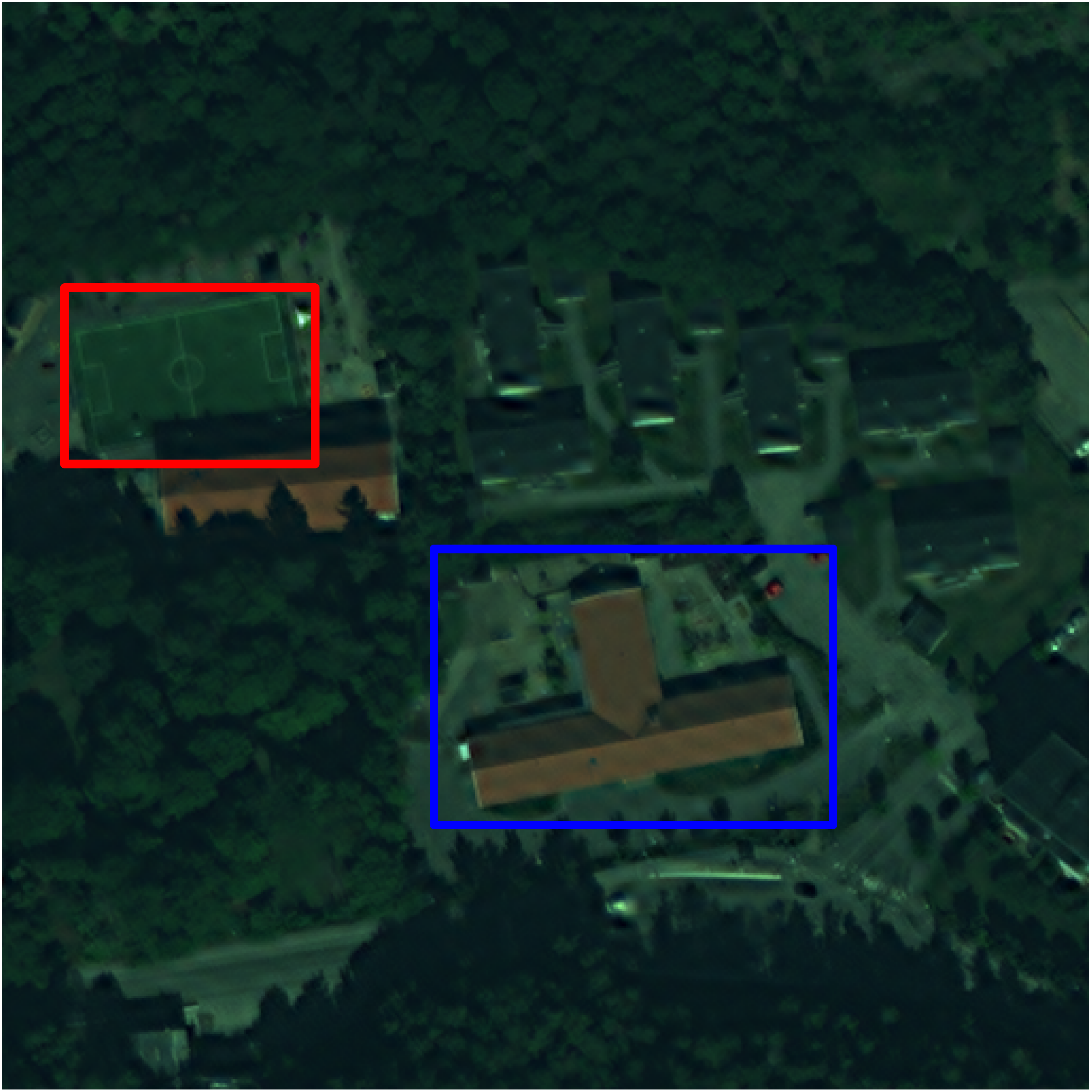}}
  \centerline{(f) BLT }\medskip
\end{minipage} 
\begin{minipage}{0.19 \linewidth}
  \centering
  \centerline{\includegraphics[width=1.0\linewidth]{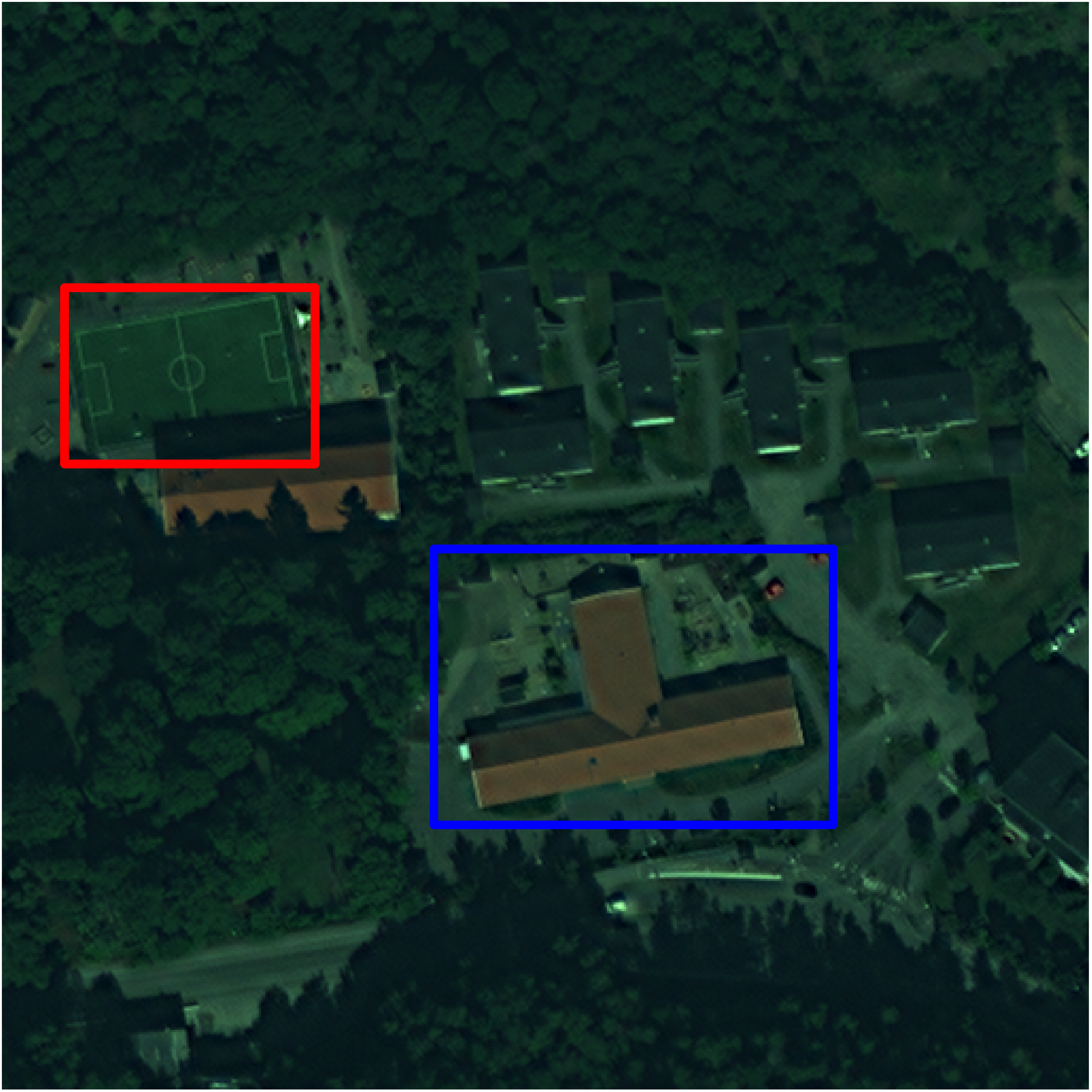}}
  \centerline{(g) F-BMP}\medskip
\end{minipage} \\
\begin{minipage}{0.19 \linewidth}
  \centering
  \centerline{\includegraphics[width=1.0\linewidth]{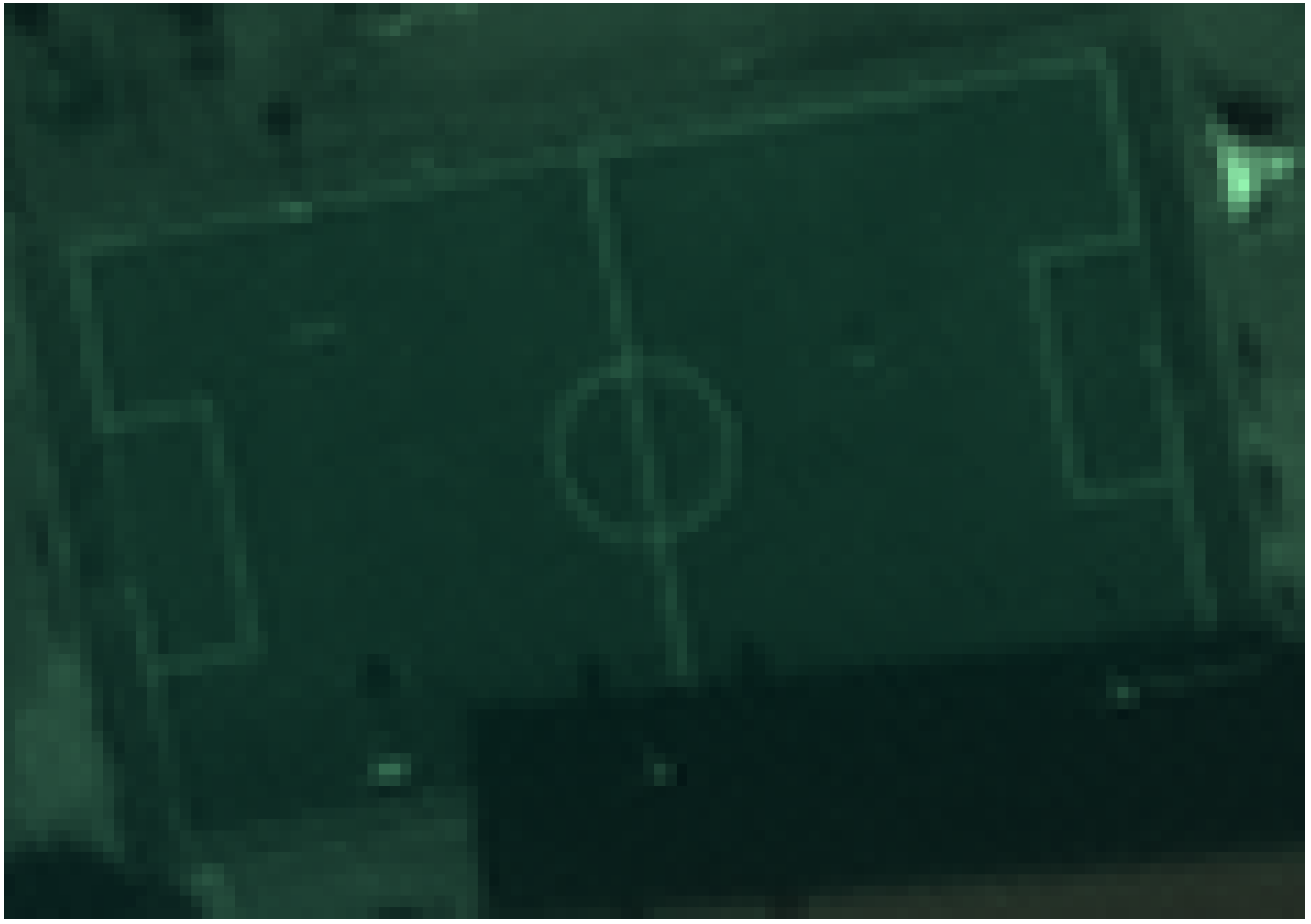}}
  \centerline{(h) HySure~(red)}\medskip
\end{minipage} 
\begin{minipage}{0.19 \linewidth}
  \centering
  \centerline{\includegraphics[width=1.0\linewidth]{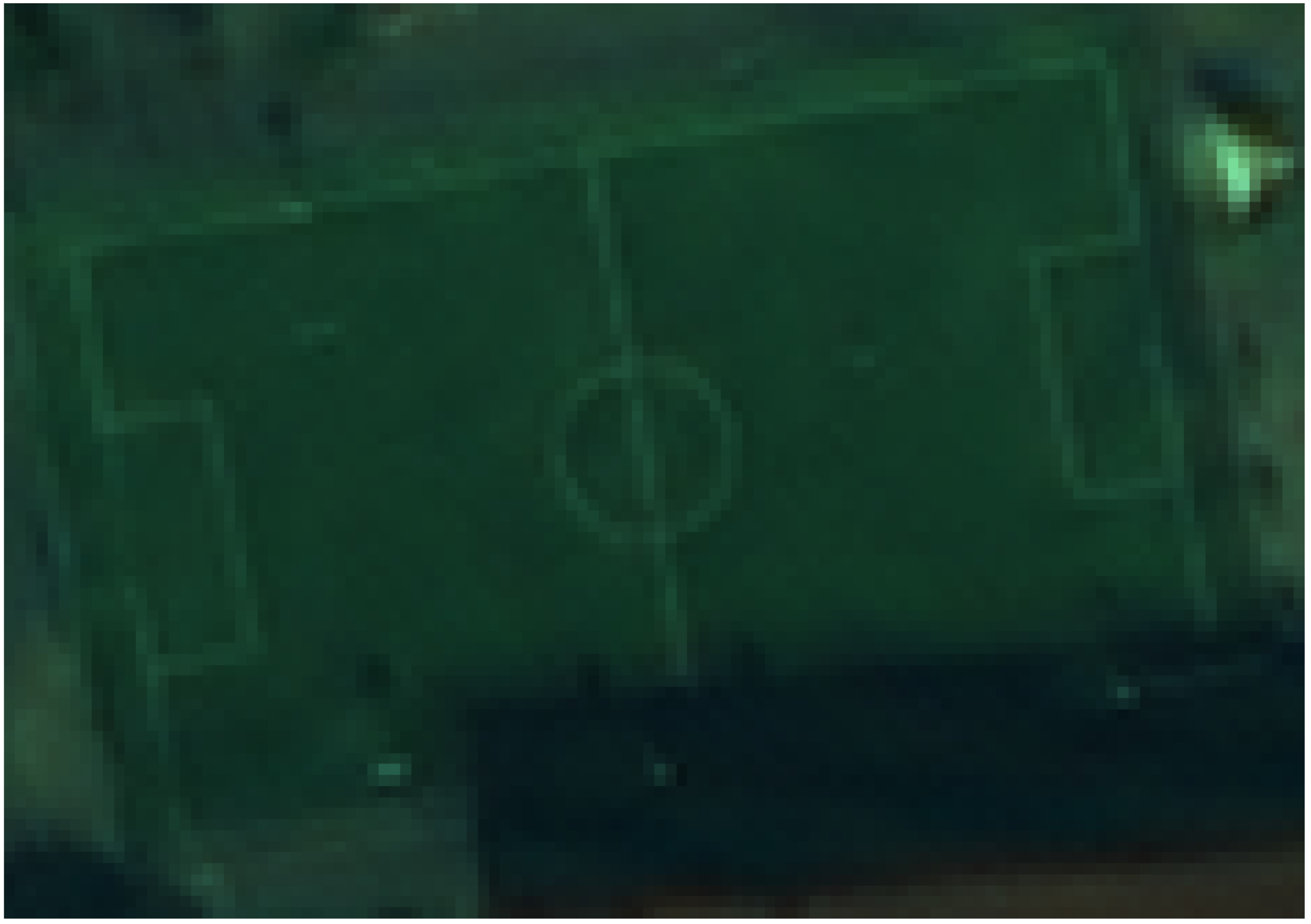}}
  \centerline{(i) R-FUSE~(red)}\medskip
\end{minipage} 
\begin{minipage}{0.19 \linewidth}
  \centering
  \centerline{\includegraphics[width=1.0\linewidth]{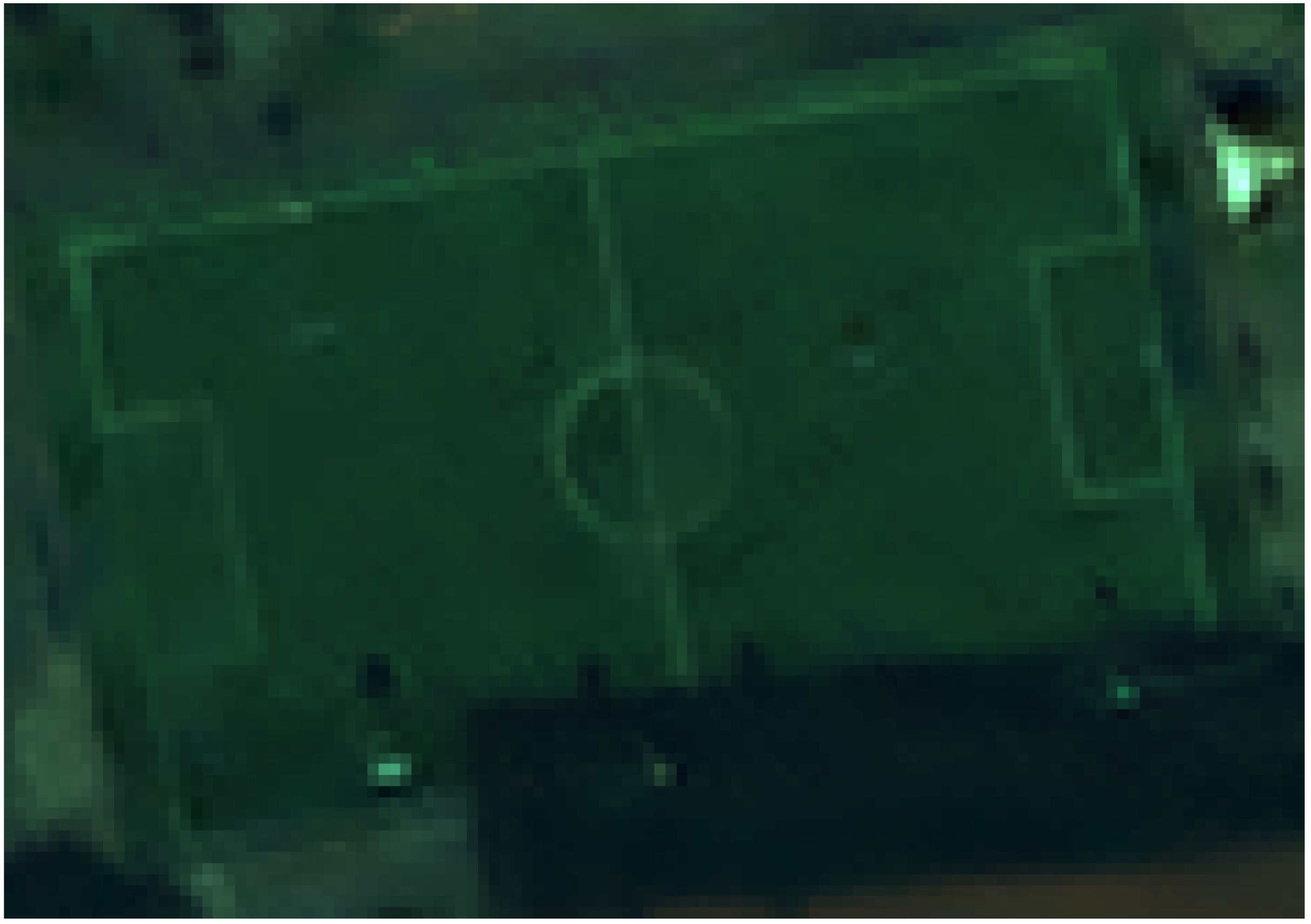}}
  \centerline{(j) GLR~(red)}\medskip
\end{minipage} 
\begin{minipage}{0.19 \linewidth}
  \centering
  \centerline{\includegraphics[width=1.0\linewidth]{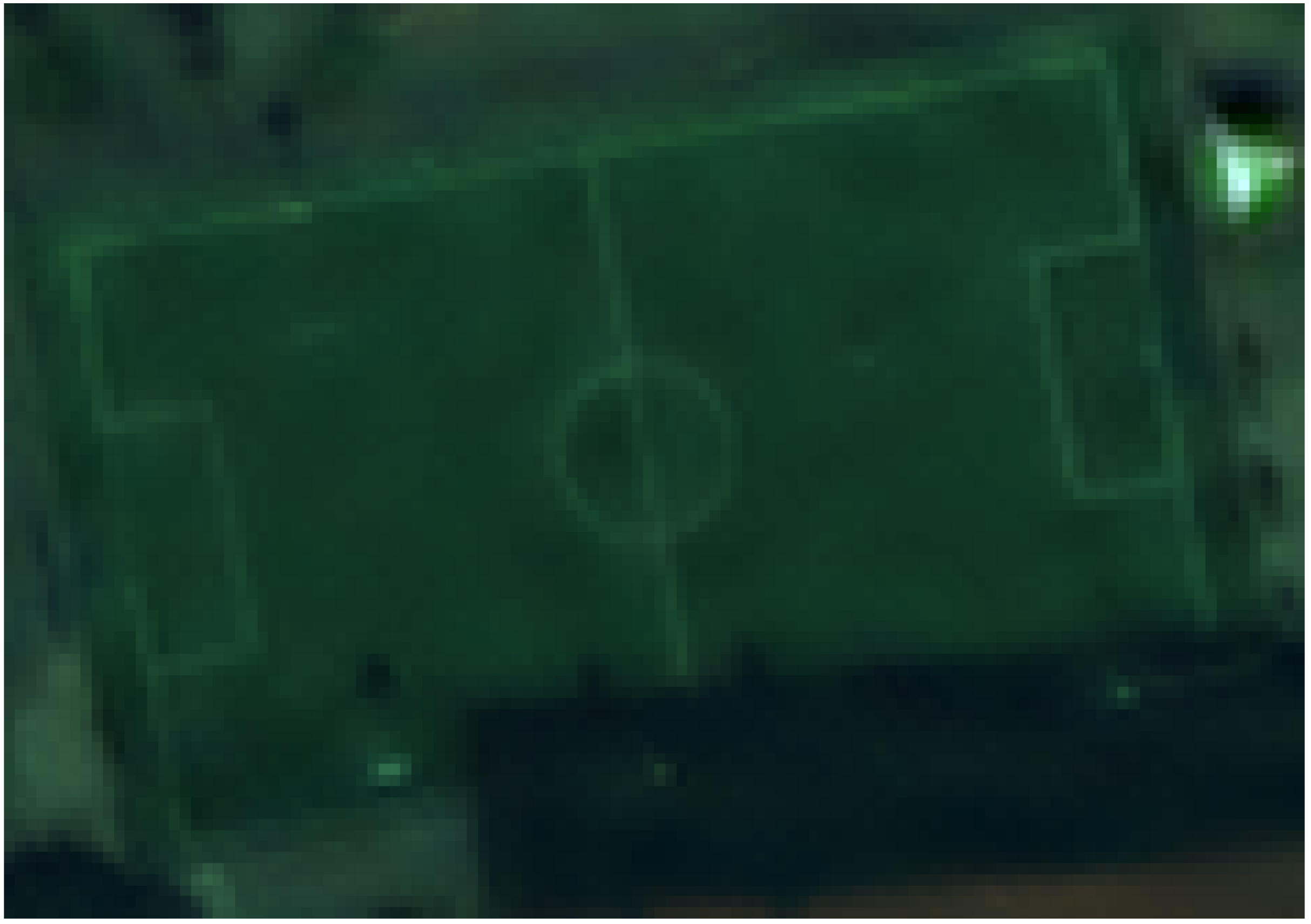}}
  \centerline{(k) BLT~(red)}\medskip
\end{minipage} 
\begin{minipage}{0.19 \linewidth}
  \centering
  \centerline{\includegraphics[width=1.0\linewidth]{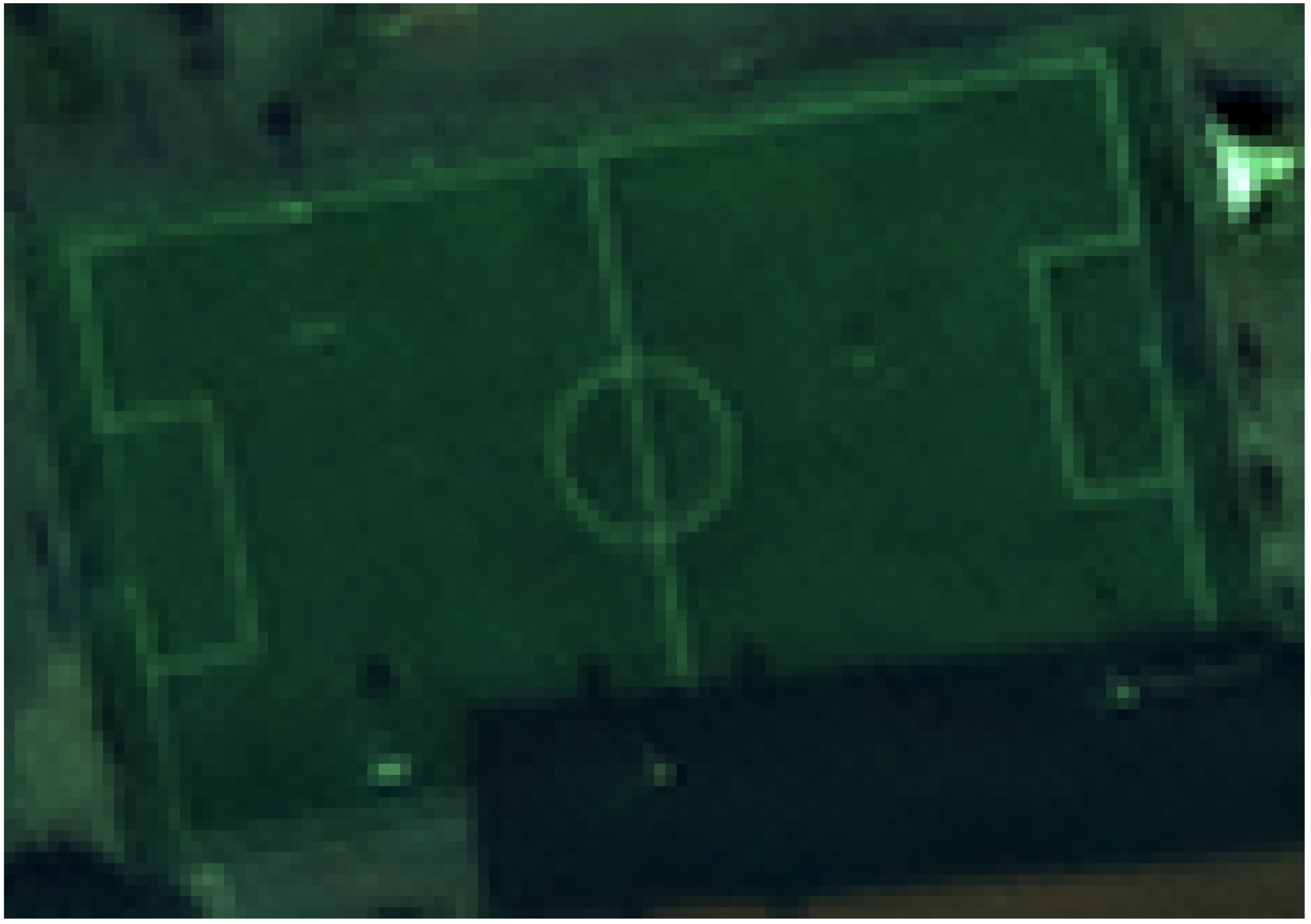}}
  \centerline{(l) F-BMP~(red)}\medskip
\end{minipage} 
\begin{minipage}{0.19 \linewidth}
  \centering
  \centerline{\includegraphics[width=1.0\linewidth]{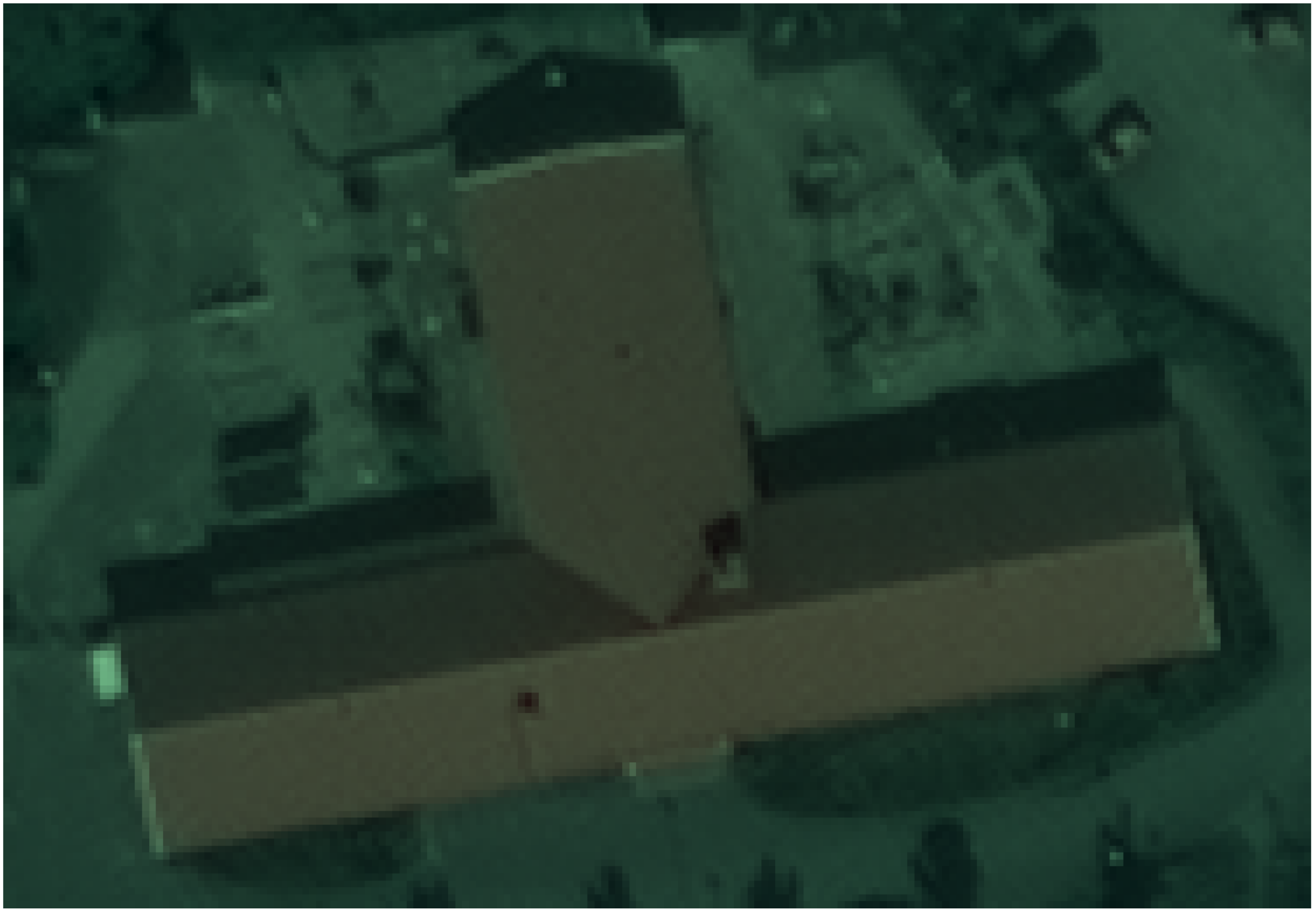}}
\centerline{(m) HySure~(blue)}\medskip
\end{minipage} 
\begin{minipage}{0.19 \linewidth}
  \centering
  \centerline{\includegraphics[width=1.0\linewidth]{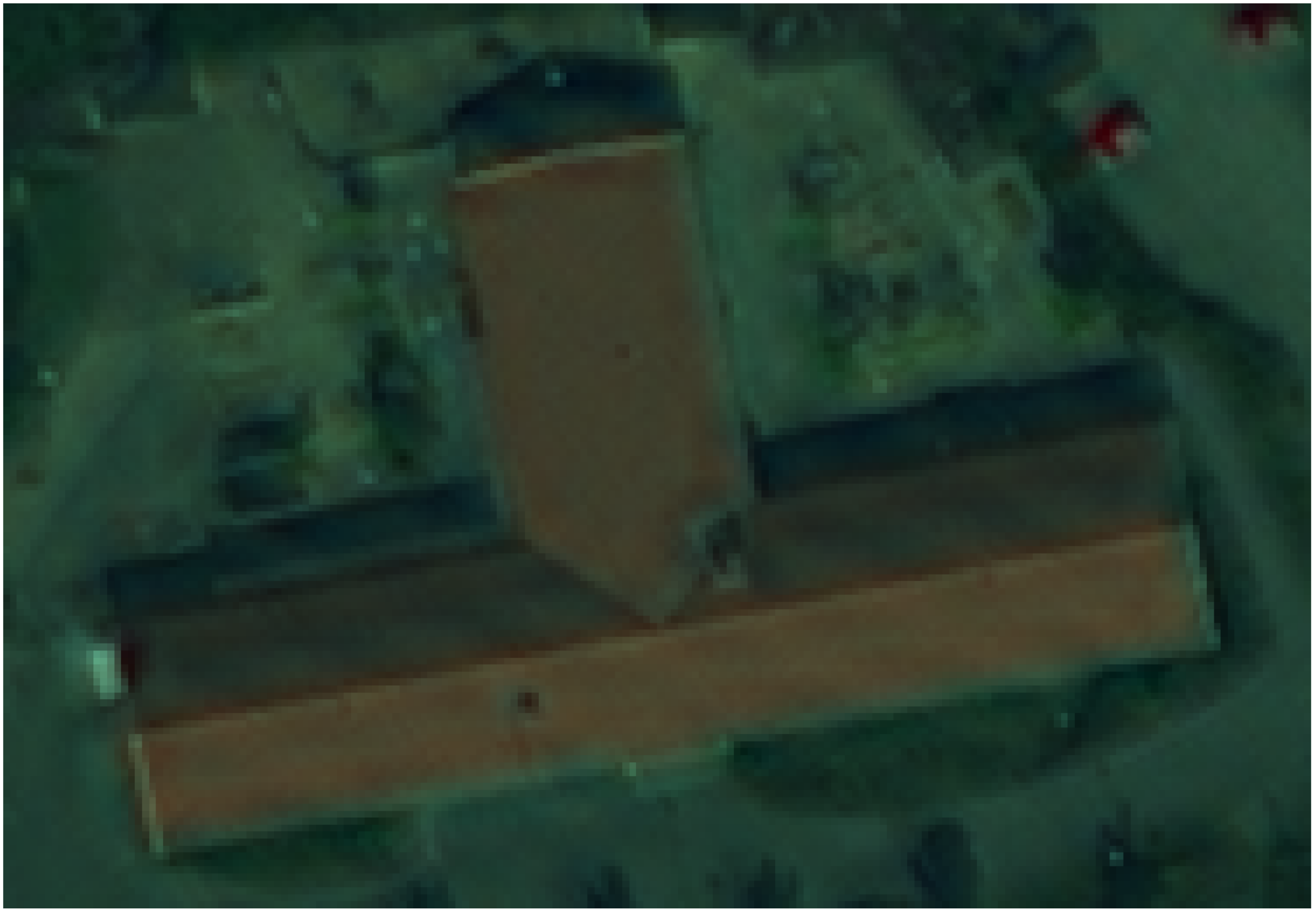}}
\centerline{(n) R-FUSE~(blue)}\medskip
\end{minipage} 
\begin{minipage}{0.19 \linewidth}
  \centering
  \centerline{\includegraphics[width=1.0\linewidth]{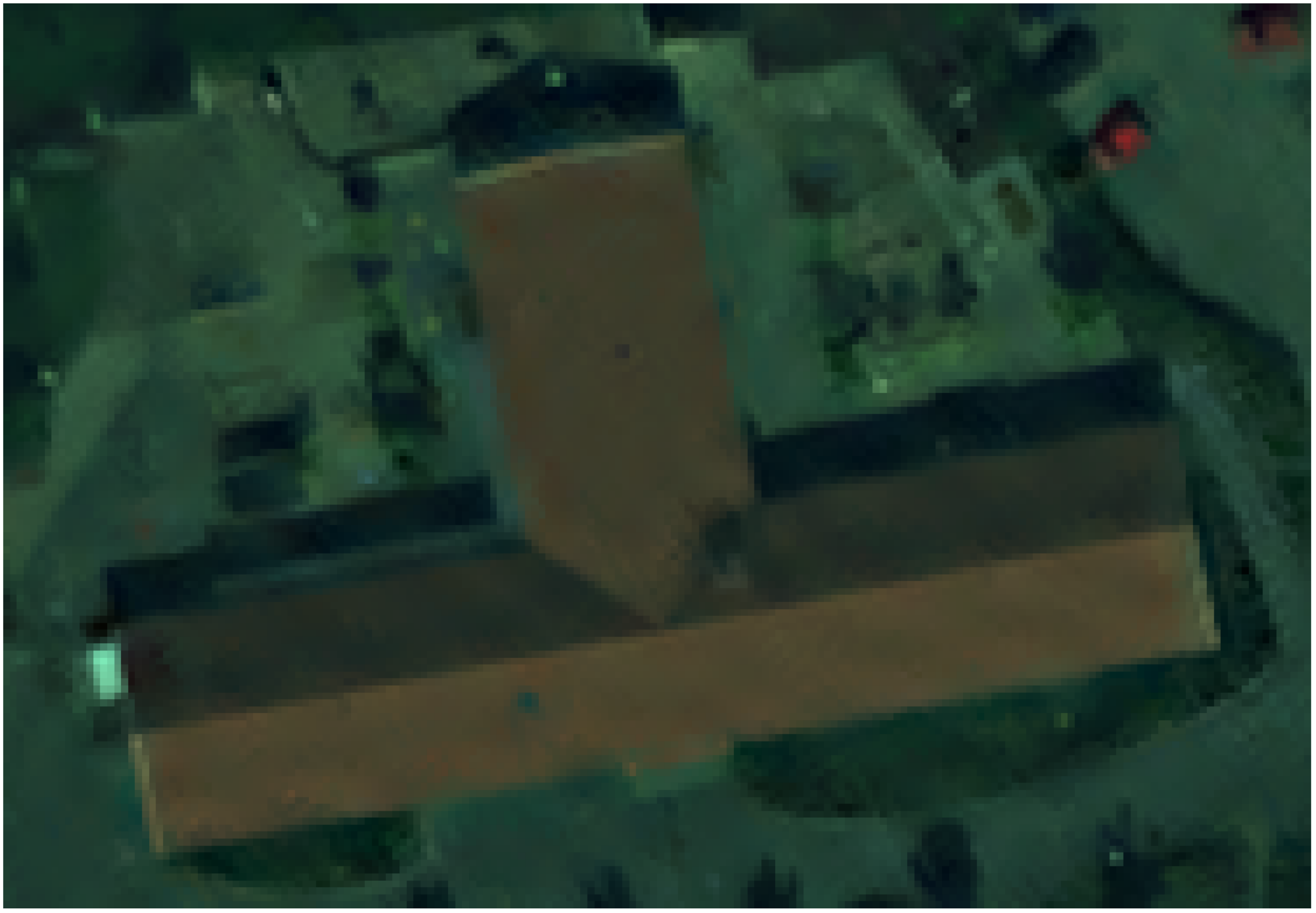}}
  \centerline{(o) GLR~(blue)}\medskip
\end{minipage} 
\begin{minipage}{0.19 \linewidth}
  \centering
  \centerline{\includegraphics[width=1.0\linewidth]{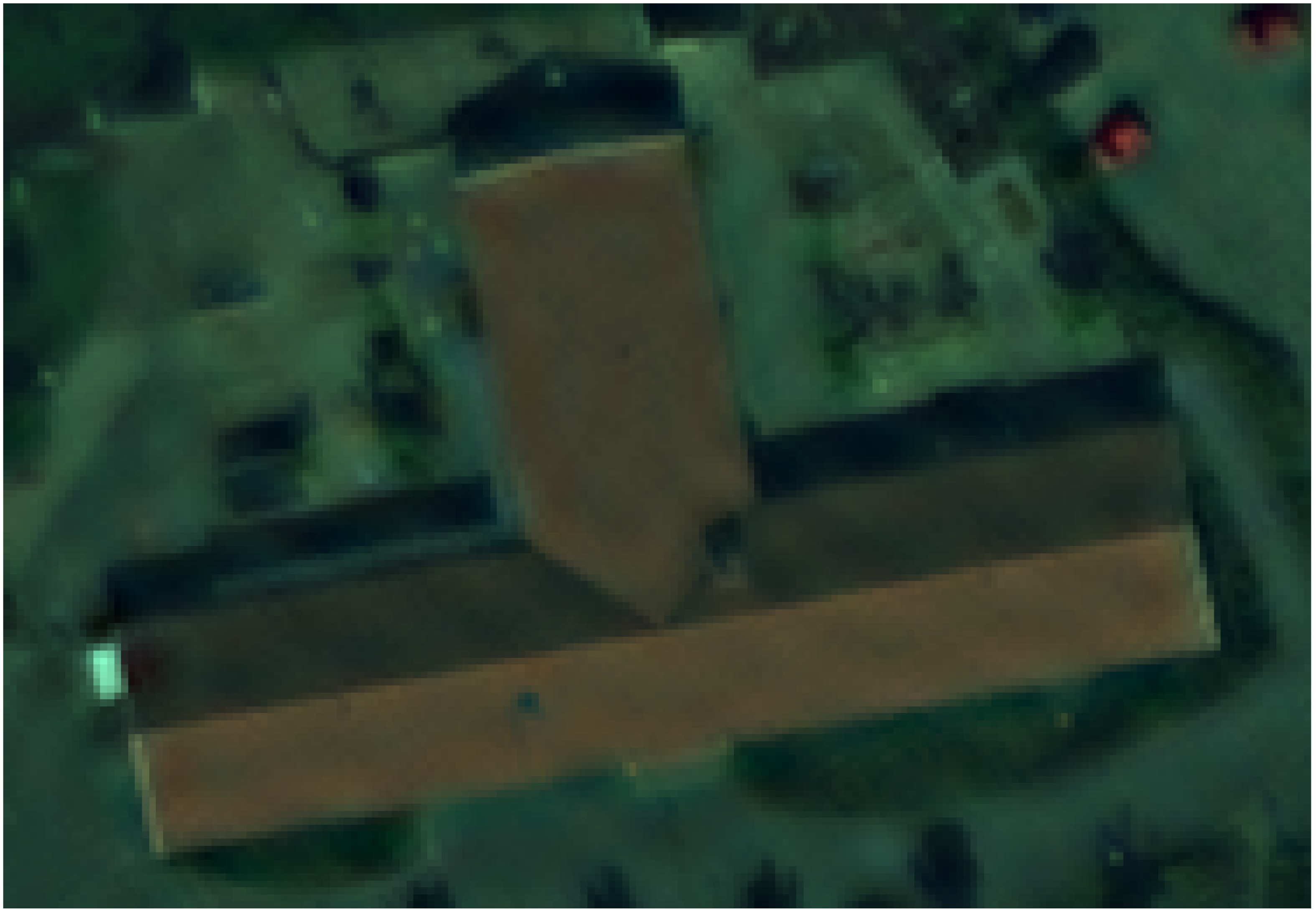}}
  \centerline{(p) BLT~(blue)}\medskip
\end{minipage} 
\begin{minipage}{0.19 \linewidth}
  \centering
  \centerline{\includegraphics[width=1.0\linewidth]{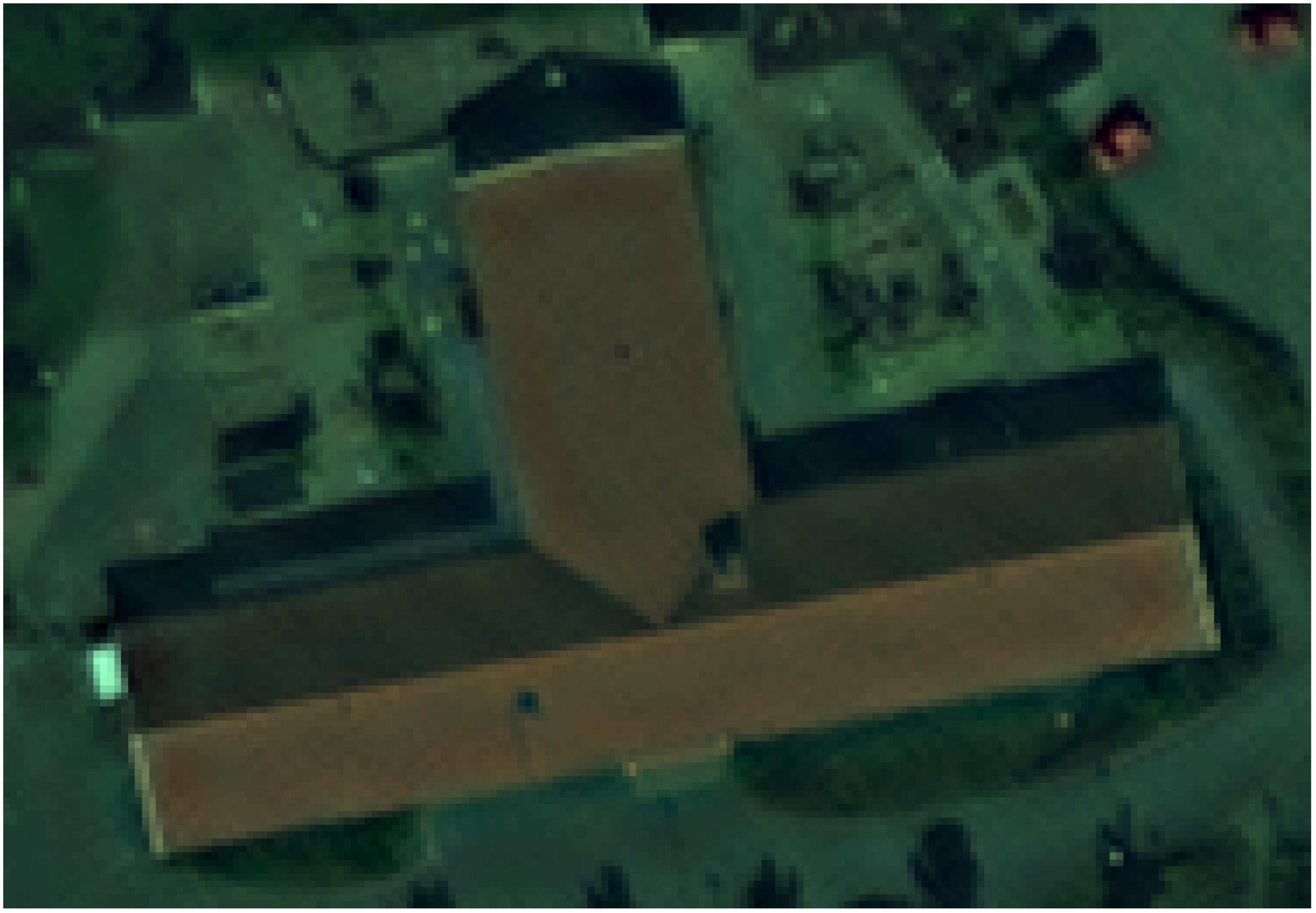}}
  \centerline{(q) F-BMP~(blue) }\medskip
\end{minipage} 

\caption{Blind pansharpening results comparison and the input data (data source:~\textit{Stockholm}). (a) The input PAN image. (b) The RGB channels of the input LRMS image (each pixel are purposely enlarged by a factor of $4$, both horizontally and vertically, to fit the space). (c) The HRMS image from HySure. (d) The HRMS image from R-FUSE. (e) The HRMS image from GLR. (f) The HRMS image from BLT. (g) The HRMS image from F-BMP. (h-l) show the zoom-in version of the region within the red block in (c-g). (m-q) show the zoom-in version of the region within blue block in (c-g).}
\label{fig:results_no_gt}
\end{figure*}

We use~\textit{Stockholm} dataset to test F-BMP by comparing with HySure, R-FUSE, GLR and BLT. In Fig.~\ref{fig:results_no_gt}, we demonstrate their pansharpened HRMS images in RGB channels with the spatial resolution of $512 \times 512$ corresponding to one of the $10$ datasets to evaluate the visual qualities. Meanwhile, we use $\overline{\operatorname{SSIM}}$ in~\eqref{eqn:ssim} as the metric to quantify the blindly pansharpened image qualities from F-BMP and the related algorithms and show the results in Table~\ref{tab:QNR}. It is worth noticing that this dataset typically has a less blurry blur kernel than we use before and the LRMS images exhibit aliasing. Thus, we use a larger $\lambda$ than that in previous experiments to better regularize the high-frequency components of the HRMS image. 

F-BMP yields the second largest $\overline{\operatorname{SSIM}}$ as shown in Table~\ref{tab:QNR} and its pansharpened HRMS images appear to carry the highest visual quality, with consistent color to the LRMS images, as well as with smoothness along the edge contours and sharpness across the edge profiles. Though HySure has the highest $\overline{\operatorname{SSIM}}$, HySure's reconstructed HRMS images have color distortion. This is exemplified in Fig.~\ref{fig:results_no_gt}(c) with noticeable color distortion from the original LRMS images. For example, the original orange roof in the LRMS image from Fig.~\ref{fig:results_no_gt}(b) turns yellow in Fig.~\ref{fig:results_no_gt}(m). The HRMS images from R-FUSE have blur and color artifacts in the roof within Fig.~\ref{fig:results_no_gt}(n). The HRMS images from GLR and BLT have blurry white lines on the soccer field in Fig.~\ref{fig:results_no_gt}(j, k) and blurry roof in Fig.~\ref{fig:results_no_gt}(o, p).
\begin{table}[tb]
\caption{Average Performance Comparison on 10~\textit{Stockholm} Images. The idea value of $\overline{\operatorname{SSIM}}$ is $1$.}
\centering
\begin{tabular}{||c|c|c|c|c|c||}
\hline
Algorithm & HySure & R-FUSE & GLR & BLT & F-BMP \\
\hline		 		
$\overline{\operatorname{SSIM}}$ &  $\mathbf{0.8722}$ & $0.6418$ & $0.7727$ & $0.5629$ & $0.7939$  \\
\hline		 
\end{tabular}
\label{tab:QNR}	
\end{table}	

\section{Conclusion}

In this article, we propose a novel and fast method for misaligned multi-spectral image pan-sharpening based on the local Laplacian prior and the Second-Order Total Generalized Variation and develop an algorithm named F-BMP (Fast and high-quality Blind Multi-spectral Pansharpening). Numerical experiments show that F-BMP significantly outperforms state-of-the-art model-based baselines in terms of both quantitative image qualities (involving known ground-truth image in the metric) and running speed. Further, when applying F-BMP to the imagery from WorldView-2 satellite without the ground-truth HRMS images, the blindly pansharpened results demonstrate the highest visual quality. F-BMP also outperforms a deep learning-based baseline in terms of average PSNR and average regressed PSNR. Moreover, F-BMP has a better generalization ability than a deep learning-based algorithm, without external training data, providing flexibility and adaptability to deal with multi-spectral imagery from a large variety of imaging platforms. 

Overall, the image priors we are exploiting in this article is rooted in the local relationship between image channels. In the future, we will exploit non-local similarity within each image channel and across image channels to pursue a higher pansharpening quality.

\appendices
\section{Solving the $\{\mathbf{u},\mathbf{p}\}$-Subproblem}
To solve the $\{\mathbf{u},\mathbf{p}\}$-subproblem efficiently, we take the approach in~\cite{guo2014new}. First, we split $\mathbf{p}$ into the left and right column as $\mathbf{p}_1$ and $\mathbf{p}_2$. Likewise, we split $\mathbf{\Lambda}^t_1$ into the left and right column as $\mathbf{\Lambda}^t_{1,1}$ and $\mathbf{\Lambda}^t_{1,2}$, $\mathbf{x}^t$ into $\mathbf{x}^t_{1}$ and $\mathbf{x}^t_{2}$. We also denote the first column of $\mathbf{\Lambda}^t_2$ as $\mathbf{\Lambda}^t_{2,1}$, the fourth column of $\mathbf{\Lambda}^t_2$ as $\mathbf{\Lambda}^t_{2,2}$ and the second or the third column of $\mathbf{\Lambda}^t_2$ as $\mathbf{\Lambda}^t_{2,3}$. Likewise, we also denote the first column of $\mathbf{y}^t$ as $\mathbf{y}^t_{1}$, the fourth column of $\mathbf{y}^t$ as $\mathbf{y}^t_{2}$ and the second or the third column of $\mathbf{y}^t$ as $\mathbf{y}^t_{3}$. By enforcing the first-order necessary conditions for optimality, we obtain the following linear equations: 

\begin{equation*}
\left\{
\begin{aligned}
& \alpha_1 \mu_1 \mathbf{D}_h^\top(\mathbf{D}_h\mathbf{u}-\mathbf{p}_1+\mathbf{\Lambda}^t_{1,1}-\mathbf{x}^t_1) + \\ &  ~~~~\alpha_1 \mu_1 \mathbf{D}_v^\top(\mathbf{D}_v\mathbf{u}-\mathbf{p}_2+\mathbf{\Lambda}^t_{1,2}-\mathbf{x}^t_2)  = \mathbf{0} \\
& \alpha_1 \mu_1 (\mathbf{p}_1-\mathbf{D}_h\mathbf{u} + \mathbf{x}^t_1-\mathbf{\Lambda}^t_{1,1}) + \\
& ~~~~\alpha_2 \mu_2 \mathbf{D}_h^{\top} (\mathbf{D}_h\mathbf{p}_1+\mathbf{\Lambda}^t_{2,1}-\mathbf{y}^t_1) + \\ 
&  ~~~~\alpha_2 \mu_2\mathbf{D}^{\top}_v(\frac{1}{2}\mathbf{D}_v\mathbf{p}_1+\frac{1}{2}\mathbf{D}_h\mathbf{p}_2 + \mathbf{\Lambda}^t_{2,3}-\mathbf{y}^t_3)   = \mathbf{0} \\
& \alpha_1 \mu_1 (\mathbf{p}_2-\mathbf{D}_v\mathbf{u}+\mathbf{x}^t_2-\mathbf{\Lambda}^t_{1,2}) + \\
&  ~~~~\alpha_2 \mu_2 \mathbf{D}_v^{\top} (\mathbf{D}_v\mathbf{p}_2+\mathbf{\Lambda}^t_{2,2}-\mathbf{y}^t_2) + \\ &  ~~~~\alpha_2 \mu_2\mathbf{D}^{\top}_h(\frac{1}{2}\mathbf{D}_h\mathbf{p}_2+\frac{1}{2}\mathbf{D}_v\mathbf{p}_1+\mathbf{\Lambda}^t_{2,3}-\mathbf{y}^t_3)   = \mathbf{0}
\end{aligned}
\right.
\end{equation*}
After grouping the terms, we have the following linear system:
\begin{equation}\label{eqn:block_toeplitz}
\begin{bmatrix}
{\mathbf{d}_{1}} & {\mathbf{d}_{4}^{\top}} & {\mathbf{d}_{5}^{\top}} \\
{\mathbf{d}_{4}} & {\mathbf{d}_{2}} & {\mathbf{d}_{6}^{\top}} \\
{\mathbf{d}_{5}} & {\mathbf{d}_{6}} & {\mathbf{d}_{3}}
\end{bmatrix}
\begin{bmatrix}
 {\mathbf{u}} \\
 {\mathbf{p}_{1}} \\
 {\mathbf{p}_{2}}
\end{bmatrix}=
\begin{bmatrix}
{\mathbf{B}_{1}} \\
{\mathbf{B}_{2}} \\
{\mathbf{B}_{3}}
\end{bmatrix}
\end{equation}
where the block matrices are defined as follows:
\begin{equation*}
\left\{
\begin{aligned}
\mathbf{B}_1 & = \alpha_1\mu_1\mathbf{D}_h^{\top}(\mathbf{x}^t_1-\mathbf{\Lambda}^t_{1,1})+\alpha_1\mu_1\mathbf{D}_v^{\top}(\mathbf{x}^t_2-\mathbf{\Lambda}^t_{1,2}) \\
\mathbf{B}_2 & = \alpha_1\mu_1(\mathbf{\Lambda}^t_{1,1}-\mathbf{x}^t_1)+\alpha_2\mu_2\mathbf{D}_h^{\top}(\mathbf{y}^t_1-\mathbf{\Lambda}^t_{2,1})+ \\
& ~~~~\alpha_2\mu_2\mathbf{D}_v^{\top}(\mathbf{y}^t_3-\mathbf{\Lambda}^t_{2,3}) \\
\mathbf{B}_3 & = \alpha_1\mu_1(\mathbf{\Lambda}^t_{1,2}-\mathbf{x}^t_2)+\alpha_2\mu_2\mathbf{D}_v^{\top}(\mathbf{y}^t_2-
    \mathbf{\Lambda}^t_{2,2})+ \\
& ~~~~\alpha_2\mu_2\mathbf{D}_h^{\top}(\mathbf{y}^t_3-\mathbf{\Lambda}^t_{2,3})\end{aligned}
\right.
\end{equation*}
\begin{equation*}
\left\{
\begin{aligned}
\mathbf{d}_1 & = \alpha_1\mu_1(\mathbf{D}_h^{\top}\mathbf{D}_h+\mathbf{D}_v^{\top}\mathbf{D}_v) \\
\mathbf{d}_2 & = \alpha_1\mu_1+\frac{1}{2}\alpha_2 \mu_2\mathbf{D}_v^{\top}\mathbf{D}_v+\alpha_2 \mu_2\mathbf{D}_h^{\top}\mathbf{D}_h \\
\mathbf{d}_3 & = \alpha_1\mu_1+\frac{1}{2}\alpha_2 \mu_2\mathbf{D}_h^{\top}\mathbf{D}_h+\alpha_2 \mu_2\mathbf{D}_v^{\top}\mathbf{D}_v \\
\mathbf{d}_4 & = -\alpha_1\mu_1 \mathbf{D}_h\\
\mathbf{d}_5 & = -\alpha_1\mu_1 \mathbf{D}_v\\
\mathbf{d}_6 & = \frac{1}{2}\alpha_2\mu_2 \mathbf{D}^{\top}_h \mathbf{D}_v
\end{aligned}
\right.
\end{equation*}
\eqref{eqn:block_toeplitz} can be efficiently solved by block-diagonalizing the coefficient matrix:
\begin{equation*}
\begin{aligned}
& \begin{bmatrix}
 \mathbf{F} & \mathbf{0} & \mathbf{0} \\
 \mathbf{0} & \mathbf{F} & \mathbf{0} \\
 \mathbf{0} & \mathbf{0} & \mathbf{F}
 \end{bmatrix}
\begin{bmatrix}
 \mathbf{d}_1 & \mathbf{d}^{\top}_4 & \mathbf{d}^{\top}_5 \\
 \mathbf{d}_4 & \mathbf{d}_2 & \mathbf{d}^{\top}_6 \\
 \mathbf{d}_5 & \mathbf{d}_6 & \mathbf{d}_3
 \end{bmatrix}
\begin{bmatrix}
 \mathbf{F} & \mathbf{0} & \mathbf{0} \\
 \mathbf{0} & \mathbf{F} & \mathbf{0} \\
 \mathbf{0} & \mathbf{0} & \mathbf{F}\end{bmatrix}^{*}  \begin{bmatrix}
  \mathbf{F}\mathbf{u} \\
  \mathbf{F}\mathbf{p}_1 \\
  \mathbf{F}\mathbf{p}_2 \\
  \end{bmatrix} 
 = \\
 & \begin{bmatrix}
  \mathbf{F} & \mathbf{0} & \mathbf{0} \\
  \mathbf{0} & \mathbf{F} & \mathbf{0} \\
  \mathbf{0} & \mathbf{0} & \mathbf{F}
  \end{bmatrix}
\begin{bmatrix}
  \mathbf{B}_1 \\
  \mathbf{B}_2 \\
  \mathbf{B}_3
\end{bmatrix}
\end{aligned}
\end{equation*}
 where $\mathbf{F}$ is the Fourier Transform matrix. By denoting $\widetilde{\mathbf{d}_j}= \diag(\mathbf{F}\mathbf{d}_j\mathbf{F}^*)$ and $\widetilde{\mathbf{d}^{\top}_j}=\conj(\diag(\mathbf{F}\mathbf{d}_j\mathbf{F}^*))$, we have:
\begin{equation*}
\left\{
\begin{aligned}
	{\widetilde{\mathbf{d}_{1}}}.*(\mathbf{F}\mathbf{u})+\widetilde{\mathbf{d}_{1}^{\top}}.*\left(\mathbf{F}\mathbf{p}_{1}\right)+\widetilde{\mathbf{d}_{5}^{\top}}.*\left(\mathbf{F}\mathbf{p}_{2}\right) & =\mathbf{F}\mathbf{B}_1 \\
	{\widetilde{\mathbf{d}_{4}}}.*(\mathbf{F}\mathbf{u})+\widetilde{\mathbf{d}_{2}}.*\left(\mathbf{F}\mathbf{p}_{1}\right)       +\widetilde{\mathbf{d}_{6}^{\top}}.*\left(\mathbf{F}\mathbf{p}_{2}\right) & =\mathbf{F}\mathbf{B}_2 \\
	{\widetilde{\mathbf{d}_{5}}}.*(\mathbf{F}\mathbf{u})+\widetilde{\mathbf{d}_{6}}.*\left(\mathbf{F}\mathbf{p}_{1}\right)       +\widetilde{\mathbf{d}_{3}}.*\left(\mathbf{F}\mathbf{p}_{2}\right) & =\mathbf{F}\mathbf{B}_3
	\end{aligned}\right.
\end{equation*}
By applying Cramer's rule, $\mathbf{u},\mathbf{p}_1,\mathbf{p}_2$ can be solved in the closed form as follows:
\begin{equation}\label{eqn:u_p_solution}
	\left\{
	\begin{aligned}
	\mathbf{u}=\mathbf{F}^{*}\left(
	\begin{vmatrix}
	{\mathbf{F}\mathbf{B}_1} & {\widetilde{\mathbf{d}_4^{\top}}} & {\widetilde{\mathbf{d}_5^{\top}}} \\
	{\mathbf{F}\mathbf{B}_2} & {\widetilde{\mathbf{d}_2}} & {\widetilde{\mathbf{d}_6^{\top}}} \\
	{\mathbf{F}\mathbf{B}_3} & {\widetilde{\mathbf{d}_6}} & {\widetilde{\mathbf{d}_3}}
	\end{vmatrix} . / \operatorname{denom} \right) \\
    \mathbf{p}_1=\mathbf{F}^{*}\left(
    \begin{vmatrix}
    {\widetilde{\mathbf{d}_1}} & {\mathbf{F}\mathbf{B}_1} & {\widetilde{\mathbf{d}_5^{\top}}}  \\
    {\widetilde{\mathbf{d}_4}} & {\mathbf{F}\mathbf{B}_2} & {\widetilde{\mathbf{d}_6^{\top}}}  \\
    {\widetilde{\mathbf{d}_5}} & {\mathbf{F}\mathbf{B}_3} & {\widetilde{\mathbf{d}_3}}
    \end{vmatrix}. / \operatorname{denom} \right)\\
    \mathbf{p}_2=\mathbf{F}^{*}\left(
    \begin{vmatrix}
    {\widetilde{\mathbf{d}_1}} & {\widetilde{\mathbf{d}_4^{\top}}} & {\mathbf{F}\mathbf{B}_1} \\
    {\widetilde{\mathbf{d}_4}} & {\widetilde{\mathbf{d}_2}}        & {\mathbf{F}\mathbf{B}_2}  \\
    {\widetilde{\mathbf{d}_5}} & {\widetilde{\mathbf{d}_6}}        & {\mathbf{F}\mathbf{B}_3}
    \end{vmatrix}. / \operatorname{denom} \right)
	\end{aligned}\right.
\end{equation}

where the division is element-wise and
\begin{equation*}
\operatorname{denom}=\begin{vmatrix}
{\widetilde{\mathbf{d}_1}} & {\widetilde{\mathbf{d}_4^{\top}}} & {\widetilde{\mathbf{d}_{5}^{\top}}} \\
{\widetilde{\mathbf{d}_4}} & {\widetilde{\mathbf{d}_2}}        & {\widetilde{\mathbf{d}_{6}^{\top}}} \\
{\widetilde{\mathbf{d}_5}} & {\widetilde{\mathbf{d}_6}}        & {\widetilde{\mathbf{d}_{3}}}
\end{vmatrix}.
 \end{equation*}
$|\cdot|$ is defined as:$\begin{vmatrix}
{a_{11}} & {a_{12}} & {a_{13}} \\ {a_{21}} & {a_{22}} & {a_{23}} \\ {a_{31}} & {a_{32}} & {a_{33}}
\end{vmatrix}=$
\begin{equation*}
\begin{aligned}
& a_{11} .* a_{22} .* a_{33}+a_{12} .* a_{23} .* a_{31}+ a_{13} .* a_{21} \cdot * a_{32}- \\ 
& a_{13} .* a_{22} .* a_{31}-a_{12} .* a_{21} .* a_{33}-a_{11} .* a_{32} .* a_{23}, 
\end{aligned}
\end{equation*}
where $.*$ is element-wise multiplication.
\section{Derivation of $\mathbf{U}$}
Consider $\mathbf{g}$ and $\mathbf{h}$ defined in continuous domain. Since $\mathbf{g}$ is isotropic, the convolution of $\mathbf{g}$ and $\mathbf{h}$ can be computed by first convolving the shifted by $(-c_x,-c_y)$ version of $\mathbf{g}$ with the rotated by $-\theta$ version of $\mathbf{h}$, followed by shifting the convolution result by $(c_x,c_y)$ and rotating by $\theta$. For notational convenience, we denote the shifted version of $\mathbf{g}$ and the rotated version of $\mathbf{h}$ as $\mathbf{g}^{\prime}$ and $\mathbf{h}^{\prime}$, respectively.
\begin{equation*}
\mathbf{g}^{\prime}(x,y)=\frac{1}{2\pi\sigma^2}e^{-\frac{x^2+y^2}{2\sigma^2}},~\text{and}
\end{equation*}
\begin{equation*}
\mathbf{h}^{\prime}(x,y)=\begin{cases}
\begin{aligned}
&\frac{1}{d} \delta(y),~-\frac{d}{2}\leqslant x \leqslant \frac{d}{2} \\
& 0, \text{~otherwise}.
\end{aligned}
\end{cases}
\end{equation*}
Step I: Convolve $\mathbf{g}^{\prime}$ with $\mathbf{h}^{\prime}$.
\begin{equation}\label{eqn:blur_prime}
\begin{aligned}
\mathbf{U}^{\prime} & = \mathbf{g}^{\prime} \ast \mathbf{h}^{\prime} \\
& = \frac{1}{d}\int_{-\infty}^{+\infty} \int_{-\infty}^{+\infty} g^{\prime}(u, v) \delta(x-u, y-v) \,du \,dv \\
& = \frac{1}{2\pi d \sigma^2}\int_{x-\frac{d}{2}}^{x+\frac{d}{2}}e^{-\frac{u^2}{2\sigma^2}}\,du\int_{-\infty}^{+\infty}e^{-\frac{v^2}{2\sigma^2}}\delta(y-v)\,dv \\
& = \frac{1}{\sqrt{2\pi}d\sigma}[\Phi(x+\frac{d}{2};\sigma)-\Phi(x-\frac{d}{2};\sigma)]e^{-\frac{y^2}{2\sigma^2}},
\end{aligned}
\end{equation}
where $\Phi(x;\sigma)=\frac{1}{\sqrt{2 \pi}\sigma } \int_{-\infty}^{x} e^{-\frac{t^2}{2 \sigma^2}} \,dt $.

Step II: Shift by $(c_x,c_y)$ and rotate by $\theta$.
\begin{equation}\label{eqn:blur_kernel_continuous}
\widetilde{\mathbf{U}}(x,y)   = \mathbf{U}^{\prime}(x^{\prime},y^{\prime}),\text{~where:} 
\end{equation}
\begin{equation*}
\begin{cases}
\begin{aligned}
x^{\prime} & = ~~(x-c_x)\cos{\theta} + (y-c_y)\sin{\theta}, \\
y^{\prime} & = -(x-c_x)\sin{\theta} + (y-c_y)\cos{\theta}.
\end{aligned}
\end{cases}
\end{equation*}

Step III: Discretize $\widetilde{\mathbf{U}}$.
\begin{equation}\label{eqn:blur_kernel_final}
\mathbf{U}(i,j)=\frac{1}{\Sigma_U}\widetilde{\mathbf{U}}(i,j),
\end{equation} 
where $R$ is the radius of $\mathbf{U}$, $\sigma$ is the standard deviation, $(i, j)$ are integer coordinates, where $-R\leqslant i \leqslant R$, $-R \leqslant j \leqslant R$, the denominator $\Sigma_U$ is the sum of all the sampled $\mathbf{U}$ to make sure the blur kernel has unit gain. 

\ifCLASSOPTIONcaptionsoff
  \newpage
\fi
\bibliographystyle{IEEEtran}
\bibliography{fast}
\end{document}